\documentclass[10pt,a4paper,oneside,notitlepage,fleqn,reqno,english,table]{article}

\usepackage[hidelinks]{hyperref} 
\usepackage{cmap} 				 

\usepackage{multicol} 

\usepackage{longtable} 
\usepackage{multirow} 
\usepackage{rotating} 
\usepackage[cp1252]{inputenc} 
\usepackage{lscape} 
\usepackage{url}    
\usepackage[section]{placeins}  
\usepackage{afterpage} 
\usepackage{capt-of}        
\usepackage{dirtytalk} 

\usepackage[T1]{fontenc}
\usepackage{tabularx, booktabs}   
\usepackage{makecell} 
\usepackage{array} 
\usepackage{xcolor} 
\usepackage{xfrac} 

\usepackage{selinput}
\SelectInputMappings{
  aacute={á},
  eacute={é},
  iacute={í},
  oacute={ó},
  uacute={ú},
  ntilde={ñ}
}

\graphicspath{{figs/}} 					
\usepackage[parfill]{parskip} 					

\usepackage{amsmath} 			
\usepackage{amsfonts}			
\usepackage{mathtools}			
\everymath{\displaystyle}		
\usepackage{amssymb}  			
\usepackage{nicefrac}  			
\usepackage{accents}			

\usepackage{tikz}  
\usetikzlibrary{shapes,arrows,calc,shadows,fit,intersections,datavisualization.formats.functions}
\tikzstyle{block}        = [draw, fill=blue!30, rectangle, text centered, minimum height=3em, minimum width=6em] 
\tikzstyle{blockred}     = [draw, fill=red!30, rectangle, text centered, minimum height=3em, minimum width=6em] 
\tikzstyle{blockyellow}  = [draw, fill=yellow!30, rectangle, text centered, minimum height=3em, minimum width=6em] 
\tikzstyle{blockyellow1} = [draw, fill=yellow!15, rectangle, text centered, minimum height=3em, minimum width=6em] 
\tikzstyle{blockyellow2} = [draw, fill=yellow!30, rectangle, text centered, minimum height=3em, minimum width=6em] 
\tikzstyle{blockyellow3} = [draw, fill=yellow!45, rectangle, text centered, minimum height=3em, minimum width=6em] 
\tikzstyle{blockyellow4} = [draw, fill=yellow!60, rectangle, text centered, minimum height=3em, minimum width=6em] 
\tikzstyle{blockgreen}   = [draw, fill=green!30, rectangle, text centered, minimum height=3em, minimum width=6em] 
\tikzstyle{blockgreen1}  = [draw, fill=green!15, rectangle, text centered, minimum height=3em, minimum width=6em] 
\tikzstyle{blockgreen2}  = [draw, fill=green!30, rectangle, text centered, minimum height=3em, minimum width=6em] 
\tikzstyle{blockgreen3}  = [draw, fill=green!45, rectangle, text centered, minimum height=3em, minimum width=6em] 
\tikzstyle{blockgreen4}  = [draw, fill=green!60, rectangle, text centered, minimum height=3em, minimum width=6em] 
\tikzstyle{blockgrey1}   = [draw, fill=black!8,  rectangle, text centered, minimum height=3em, minimum width=6em] 
\tikzstyle{blockgrey2}   = [draw, fill=black!16, rectangle, text centered, minimum height=3em, minimum width=6em] 
\tikzstyle{blockgrey3}   = [draw, fill=black!24, rectangle, text centered, minimum height=3em, minimum width=6em] 
\tikzstyle{blockgrey4}   = [draw, fill=black!32, rectangle, text centered, minimum height=3em, minimum width=6em] 
\tikzstyle{blockgrey5}   = [draw, fill=black!40, rectangle, text centered, minimum height=3em, minimum width=6em] 
\tikzstyle{blocknofill}  = [draw=black!50, line width=1.5pt, rectangle, rounded corners, text centered, minimum height=3em, minimum width=6em]
\tikzstyle{blockhigh}    = [draw, fill=blue!20, rectangle, text centered, minimum height=6em, minimum width=6em]
\tikzstyle{noblock}      = [draw=black!50, line width=1.5pt, rectangle, rounded corners, text centered, minimum height=3em, minimum width=6em]
\tikzstyle{sum}          = [draw, fill=blue!20, circle, node distance=1cm]
\tikzstyle{sumrest}      = [draw, fill=black, circle, radius=0.5cm]
\tikzstyle{pinstyle}     = [pin edge={to-,thin,black}]
\tikzstyle{title}        = [text centered]
\pgfdeclarelayer{background}
\pgfdeclarelayer{foreground}
\pgfsetlayers{background,main,foreground}

\usepackage{pgfplots}
\pgfplotsset{compat=1.13}
\pgfplotsset{colormap/bluered}
\usepackage{pgfplotstable}
\usepgfplotslibrary{polar}
\usepgfplotslibrary{colormaps}
\usepgfplotslibrary{external}

\usepackage{graphicx, color} 	
\usepackage{epstopdf}  			

\allowdisplaybreaks 

\newcommand{\ds}    {\displaystyle} 		
\newcommand{\sss}   {\scriptscriptstyle}    

\newcommand{\nm}   		[1] {\ensuremath{\mathrm{#1}}} 									
\newcommand{\neweq}     [2] {\begin{equation} \mathrm{#1}\label{#2} \end{equation}} 	

\newcommand{\CC}		{{C\nolinebreak[4]\hspace{-.05em}\raisebox{.4ex}{\tiny\bf ++}}}

\renewcommand{\vec}		[1] {\mbox{\boldmath{\ensuremath{\mathrm{#1}}}}}	
\newcommand{\lrp}       [1] {\left(#1\right)}								
\newcommand{\lrsb}      [1] {\left[#1\right]}								

\newcommand{\muEND}      [1] {\mu_{{\sss{END}} {#1}}}
\newcommand{\sigmaEND}   [1] {\sigma_{{\sss{END}} {#1}}}
\newcommand{\maxEND}     [1] {{\zeta}_{{\sss{END}}\lvert {#1} \rvert}}
\newcommand{\mun}        [1] {\mu_{{#1}n}}
\newcommand{\sigman}     [1] {\sigma_{{#1}n}}
\newcommand{\mui}        [1] {\mu_{{#1}i}}
\newcommand{\sigmai}     [1] {\sigma_{{#1}i}}

\newcommand{\DeltarNBBestnorm}  {\|\Delta \hat{\vec r}_{\sss NB}^{\sss B}\|}
\newcommand{\DeltarNBBvisnorm}  {\|\Delta \circled{\vec r}_{\sss NB}^{\sss B}\|}

\newcommand{\XENDjworst}      [1] {{#1}_{{\sss END,WORST}j}}

\newcommand{\DeltarBestnorm}    {\|\Delta\hat{\vec r}^{\sss B}\|}

\newcommand{\DeltarBvisnorm}    {\|\Delta \circled{\vec r}^{\sss B}\|}

\newcommand{\hypertt}	[1] {\hyperlink{#1}{\texttt{#1}}}

\newcommand{\circled}            [1] {\accentset{\circ}{#1}}
\newcommand{\OC}          {O_{\sss C}}      
\newcommand{\iIMGi}         {\vec i_{1}^{\sss IMG}}
\newcommand{\iIMGii}        {\vec i_{2}^{\sss IMG}}
\newcommand{\OIMG}          {O_{\sss IMG}}     

\newcommand{\Sh}            {S_{\sss H}}
\newcommand{\Sv}            {S_{\sss V}}
\newcommand{\cIMG}          {\vec c^{\sss IMG}}
\newcommand{\pIMGi}         {p_{1}^{\sss IMG}}
\newcommand{\pIMGii}        {p_{2}^{\sss IMG}}
\newcommand{\sPX}           {s_{\sss PX}}
\newcommand{\cIMGi}         {c_{1}^{\sss IMG}}
\newcommand{\cIMGii}        {c_{2}^{\sss IMG}}

\newcommand{\DeltatTRUTH}	{\Delta t_{\sss TRUTH}}
\newcommand{\DeltatSENSED}  {\Delta t_{\sss SENSED}}
\newcommand{\DeltatEST}  	{\Delta t_{\sss EST}}
\newcommand{\DeltatCNTR}  	{\Delta t_{\sss CNTR}}

\newcommand{\DeltatIMG}  	{\Delta t_{\sss IMG}}

\newcommand{\deltaCNTR}  	{\vec{\delta}_{\sss CNTR}}		
\newcommand{\deltaTARGET}  	{\vec{\delta}_{\sss TARGET}}	    

\newcommand{\xvec}   			{\vec x}
\newcommand{\xveczero}	 		{\vec x_0}
\newcommand{\xvecestzero}		{\hat{\vec x}_0}

\newcommand{\xvecest}	  		{\hat{\vec x}}
\newcommand{\xvectilde} 		{\widetilde{\vec x}}

\newcommand{\xvecvis}			{\circled{\vec x}}
\newcommand{\xvecviszero}		{\circled{\vec x}_0}

\newcommand{\xTRUTH}			{\vec x_{\sss TRUTH}}
\newcommand{\xSENSED}  			{\vec x_{\sss SENSED}}
\newcommand{\xEST}  			{\vec x_{\sss EST}}
\newcommand{\xREF}	  			{\vec x_{\sss REF}}

\newcommand{\xIMG}  			{\vec x_{\sss IMG}}

\newcommand{\qNBvis}		{\circled{\vec q}_{\sss NB}}

\newcommand{\FE}		 {F_{\sss E}}
\newcommand{\FC}		 {F_{\sss C}}
\newcommand{\FIMG}		 {F_{\sss IMG}}

\newcommand{\iEi}        {\vec i_{\sss1}^{\sss E}}
\newcommand{\iEii}       {\vec i_{\sss2}^{\sss E}}
\newcommand{\iEiii}      {\vec i_{\sss3}^{\sss E}}

\newcommand{\FN} 		  {F_{\sss N}}

\newcommand{\FB}		  {F_{\sss B}}

\newcommand{\iBi}         {\vec i_{\sss1}^{\sss B}}
\newcommand{\iBii}        {\vec i_{\sss2}^{\sss B}}
\newcommand{\iBiii}       {\vec i_{\sss3}^{\sss B}}

\newcommand{\FP}		  {F_{\sss P}}

\newcommand{\TEgdt}   		{\vec T^{\sss E,GDT}}

\newcommand{\zetaEBvis}	{\circled{\vec \zeta}_{\sss EB}}

\newcommand{\qBC}		{\vec q_{\sss BC}}

\newcommand{\zetaECivis}		{\circled{\vec \zeta}_{{\sss EC}i}}
\newcommand{\vECC}				{\vec v_{\sss EC}^{\sss C}}
\newcommand{\vCPC}				{\vec v_{\sss CP}^{\sss C}}

\newcommand{\vB} 	   			{\vec v^{\sss B}}

\newcommand{\vN}	    		{\vec v^{\sss N}}

\newcommand{\vNii}      	  	{v_{\sss 2}^{\sss N}}

\newcommand{\vECE}		{\vec v_{\sss EC}^{\sss E}}
\newcommand{\vEPE}		{\vec v_{\sss EP}^{\sss E}}
\newcommand{\vCPE}		{\vec v_{\sss CP}^{\sss E}}

\newcommand{\vTAS}          {\vec v_{\sss TAS}}

\newcommand{\vtasINI}       {v_{\sss TAS,INI}}
\newcommand{\vtasEND}       {v_{\sss TAS,END}}

\newcommand{\vwindINI}     	{v_{\sss WIND,INI}}
\newcommand{\vwindEND}     	{v_{\sss WIND,END}}

\newcommand{\pC}          {\vec p^{\sss C}}
\newcommand{\pCdot}	      {\vec{\dot p}^{\sss C}}
\newcommand{\xiECC}			{\vec \xi_{\sss EC}^{\sss C}}
\newcommand{\wECC}			{\vec \omega_{\sss EC}^{\sss C}}
\newcommand{\pCbar} 	  {\vec {\bar p}^{\sss C}}
\newcommand{\pCbari}      {{\bar p}_{1}^{\sss C}}
\newcommand{\pCbarii}     {{\bar p}_{2}^{\sss C}}
\newcommand{\pCi}         {p_{1}^{\sss C}}
\newcommand{\pCii}        {p_{2}^{\sss C}}
\newcommand{\pCiii}       {p_{3}^{\sss C}}
\newcommand{\pIMG}          {\vec p^{\sss IMG}}
\newcommand{\pCbardot} 	  {\vec {\dot {\bar p}}^{\sss C}}
\newcommand{\pCiiidot}    {\dot{p}_{3}^{\sss C}}
\newcommand{\pIMGdot}	    {\vec{\dot p}^{\sss IMG}}

\newcommand{\TECE}		{\vec T_{\sss EC}^{\sss E}}
\newcommand{\TEPE}		{\vec T_{\sss EP}^{\sss E}}
\newcommand{\TCPE}		{\vec T_{\sss CP}^{\sss E}}
\newcommand{\TEPEdot}	{\vec {\dot T}_{\sss EP}^{\sss E}}
\newcommand{\TCPCdot}		{\vec {\dot T}_{\sss CP}^{\sss C}}
\newcommand{\TECEdot}	{\vec {\dot T}_{\sss EC}^{\sss E}}

\newcommand{\wECCskew}		{\widehat{\vec \omega}_{\sss EC}^{\sss C}}
\newcommand{\TCPC}			{\vec T_{\sss CP}^{\sss C}}

\newcommand{\wNBB}			{\vec \omega_{\sss NB}^{\sss B}}

\newcommand{\wECEskew}		{\widehat{\vec \omega}_{\sss EC}^{\sss E}}

\newcommand{\vEPC}			{\vec v_{\sss EP}^{\sss C}}

\newcommand{\REC}		{\vec R_{\sss EC}}
\newcommand{\RCE}		{\vec R_{\sss CE}}

\newcommand{\RECdot}	{\dot{\vec R}_{\sss EC}}

\newcommand{\iCi}         {\vec i_{1}^{\sss C}}
\newcommand{\iCii}        {\vec i_{2}^{\sss C}}
\newcommand{\iCiii}       {\vec i_{3}^{\sss C}}

\newcommand{\qNB}			{\vec q_{\sss NB}}

\newcommand{\qNBest}		{\hat{\vec q}_{\sss NB}}

\newcommand{\gammaTAS}  	  	{\gamma_{\sss TAS}}

\newcommand{\chiWINDINI}       	{\chi_{\sss WIND,INI}}
\newcommand{\chiWINDEND}       	{\chi_{\sss WIND,END}}

\newcommand{\psiest}			{\hat \psi}
\newcommand{\psivis}			{\circled \psi}

\newcommand{\thetaest}			{\hat \theta}
\newcommand{\thetavis}			{\circled \theta}

\newcommand{\xiest}				{\hat \xi}
\newcommand{\xivis}				{\circled \xi}

\newcommand{\chiINI}            {\chi_{\sss INI}}
\newcommand{\chiEND}            {\chi_{\sss END}}

\newcommand{\xiTURN}            {\xi_{\sss TURN}}

\newcommand{\gammaTASCLIMB}     {\gamma_{\sss TAS,CLIMB}}

\newcommand{\tTURN}				{t_{\sss TURN}}

\newcommand{\tGNSS}				{t_{\sss GNSS}}
\newcommand{\tEND}				{t_{\sss END}}

\newcommand{\Hp}            {H_{\sss P}}
\newcommand{\HpINI}         {H_{\sss P,INI}}
\newcommand{\HpEND}         {H_{\sss P,END}}

\newcommand{\DeltaT}        {\Delta T}
\newcommand{\Deltap}        {\Delta p}

\newcommand{\hest}			{\hat{h}}
\newcommand{\hvis}			{\circled{h}}

\newcommand{\DeltaTINI}     {\Delta T_{\sss INI}}
\newcommand{\DeltaTEND}     {\Delta T_{\sss END}}

\newcommand{\DeltapINI}     {\Delta p_{\sss INI}}
\newcommand{\DeltapEND}     {\Delta p_{\sss END}}

\newcommand{\Deltaxhorest}				{\Delta \hat{x}_{\sss HOR}}
\newcommand{\Deltaxhorvis}				{\Delta \circled{x}_{\sss HOR}}

\begin{document}

\title{GNSS-Denied Semi Direct Visual Navigation for Autonomous UAVs Aided by PI-Inspired Inertial Priors}

\author{Eduardo Gallo \footnote{Contact: e.gallo@alumnos.upm.es, edugallo@yahoo.com, \url{https://orcid.org/0000-0002-7397-0425}} \footnote{Affiliation: Centro de Automática y Robótica, Universidad Politécnica de Madrid - Consejo Superior de Investigaciones Científicas. Centre for Automation and Robotics, Polytechnic University of Madrid.} and Antonio Barrientos \footnote{Contact: antonio.barrientos@upm.es, \url{https://orcid.org/0000-0003-1691-3907}} \footnote{Affiliation: Centro de Automática y Robótica, Universidad Politécnica de Madrid - Consejo Superior de Investigaciones Científicas. Centre for Automation and Robotics, Polytechnic University of Madrid.}}  
\date{December 2022}
\maketitle


\section*{Abstract}

This article proposes a method to diminish the horizontal position drift experienced by the \hypertt{VNS} (Visual Navigation System) installed onboard a \hypertt{UAV} (Unmanned Air Vehicle) by supplementing its pose estimation non linear optimizations with priors based on the outputs of the \hypertt{GNSS} (Global Navigation Satellite System) Denied \hypertt{INS} (Inertial Navigation System). The method is inspired by a \hypertt{PI} (Proportional Integral) control loop, in which the attitude and altitude inertial outputs act as targets to ensure that the visual estimations do not deviate past certain thresholds from their inertial counterparts. The resulting \hypertt{IA-VNS} (Inertially Assisted Visual Navigation System) achieves major reductions in the horizontal position drift inherent to the \hypertt{GNSS}-Denied navigation of autonomous \hypertt{UAV}s. Stochastic high fidelity Monte Carlo simulations of two representative scenarios involving the loss of \hypertt{GNSS} signals are employed to evaluate the results and to analyze their sensitivity to the terrain type overflown by the aircraft. The authors release the \nm{\CC} implementation of both the navigation algorithms and the high fidelity simulation as open-source software \cite{Gallo2020_simulation}.


\textbf{\emph{Keywords}}: GNSS-Denied, visual inertial navigation, autonomous navigation, UAV, optimization, prior


\section*{Mathematical Notation}

Any variable with a hat accent \nm{< \hat{\cdot} >} refers to its (inertial) estimated value, and with a circular accent \nm{< \circled{\cdot} >} to its (visual) estimated value. In the case of vectors, which are displayed in bold (e.g., \nm{\vec x}), other employed symbols include the wide hat \nm{< \widehat{\cdot} >}, which refers to the skew-symmetric form, the bar \nm{<\bar \cdot>}, which represents the vector homogeneous coordinates, and the double vertical bars \nm{< \| \cdot \| >}, which refer to the norm. In the case of scalars, the vertical bars \nm{< | \cdot | >} refer to the absolute value. When employing quaternions and rigid body poses, which are also displayed in bold (e.g., \nm{\vec q} and \nm{\vec \zeta}), the asterisk superindex \nm{< \cdot^{\ast} >} refers to the conjugate, their concatenation and multiplication are represented by \nm{\circ} and \nm{\otimes} respectively, and \nm{\oplus} and \nm{\ominus} refer to the plus and minus operators.

This article includes various non linear optimizations solved in the spaces of both rigid body rotations and full motions, instead of Euclidean spaces. Hence, it relies on the Lie algebra of the special orthogonal group of \nm{\mathbb{R}^3}, known as \nm{\mathbb{SO}(3)}, and that of the special euclidean group of \nm{\mathbb{R}^3}, represented by \nm{\mathbb{SE}(3)}, in particular what refers to the groups actions, concatenations, perturbations, and Jacobians, as well as with their tangent spaces (the rotation vector \nm{\vec r} and angular velocity \nm{\vec \omega} for rotations, the transform vector \nm{\vec \tau} and twist \nm{\vec \xi} for motions). \cite{LIE, Sola2017, Sola2018} are recommended as references.
\begin{table}[ht]
\begin{tabular}{lp{6.4cm}p{0.1cm}lp{6.4cm}}
\nm{\gammaTAS}				& Aerodynamic path angle					          & & \nm{h}							        & Geometric altitude					\\ 
\nm{\delta}						& Error threshold	 						              & & \nm{\Hp}						        & Pressure altitude						\\   
\nm{\deltaCNTR}				& Position of throttle and control surfaces & &	\nm{\vec I}						      & Camera image	 						\\
\nm{\deltaTARGET}			& Control targets							              & & \nm{\vec J}						      & Jacobian	 							\\
\nm{\Delta}						& Estimation error, increment 			        & & \nm{\mathcal M}					    & \nm{\mathbb{SE}(3)} Lie group element	\\
\nm{\Deltap}					& Atmospheric pressure offset				        & & \nm{\vec p}						      & Point, feature	 					\\
\nm{\DeltaT}					& Atmospheric temperature offset		        & & \nm{\vec q}						      & Attitude, unit quaternion				\\
\nm{\theta}						& Body pitch angle					  		          & & \nm{\vec r}						      & Attitude, rotation vector				\\
\nm{\vec \zeta}				& Pose, unit dual quaternion	 			        & & \nm{\vec R}						      & Attitude, rotation matrix				\\
\nm{\mu}			  			& Mean or expected value	  				        & & \nm{\mathcal R}			        & \nm{\mathbb{SO}(3)} Lie group element	\\
\nm{\xi}				  		& Body bank angle 							            & & \nm{\sPX}                   & Pixel size \\
\nm{\vec \xi}					& Motion (\nm{\mathbb{SE}(3)}) velocity	or twist & &  \nm{S}							  & Sensor dimension		 				\\  
\nm{\vec \Pi}					& Camera projection 						            & & \nm{t}							        & Time									\\
\nm{\varrho_{TUK}}		& Tukey error function	 			  		        & & \nm{\vec T}						      & Displacement							\\
\nm{\sigma}						& Standard deviation 			    			        & & \nm{\TEgdt}						      & Geodetic coordinates					\\
\nm{\vec \tau}				& Pose, transform vector 					          & & \nm{v}							        & Speed									\\
\nm{\vec \phi}				& Attitude, Euler angles 					          & & \nm{\vec v}						      & Velocity								\\
\nm{\chi}					  	& Bearing									                  & & \nm{w_{TUK}}					      & Tukey weight function					\\
\nm{\psi}						  & Heading or body yaw angle					        & &	\nm{x}							        & Horizontal distance					\\
\nm{\vec \omega}			& Angular (\nm{\mathbb{SO}(3)}) velocity 	  & & \nm{\vec x}						      & Position								\\	
\nm{E_{PO}}						& Pose optimization error 					        & & \nm{\xvecest = \xEST}			  & Inertial estimated trajectory			\\
\nm{E_{q}}						& Attitude adjustment error 				        & & \nm{\xvecvis = \xIMG}			  & Visual estimated trajectory			\\
\nm{E_{RP}}						& Reprojection error 								        & & \nm{\xREF}						      & Reference objectives					\\
\nm{f}						  	& Focal length	                            & & \nm{\xvectilde = \xSENSED}	& Sensed trajectory						\\
\nm{\vec g}			      & Lie group action (transformation)         & & \nm{\xvec = \xTRUTH} 		  	& Real trajectory						\\
\end{tabular}
\end{table}

Five different reference frames are employed in this article: the \hypertt{ECEF} frame \nm{\FE}\footnote{The \hypertt{ECEF} frame \nm{\FE} is centered at the Earth center of mass, with \nm{\iEiii} pointing towards the geodetic North along the Earth rotation axis, \nm{\iEi} contained in both the Equator and zero longitude planes, and \nm{\iEii} orthogonal to \nm{\iEi} and \nm{\iEiii} forming a right handed system.}, the \hypertt{NED} frame \nm{\FN}\footnote{The \hypertt{NED} frame \nm{\FN} is centered at the aircraft center of mass, with axes aligned with the geodetic North, East, and Down directions.}, the body frame \nm{\FB}\footnote{The body frame \nm{\FB} is centered at the aircraft center of mass, with \nm{\iBi} contained in the plane of symmetry of the aircraft pointing forward along a fixed direction, \nm{\iBiii} contained in the plane of symmetry of the aircraft, normal to \nm{\iBi} and pointing downward, and \nm{\iBii} orthogonal to both in such a way that they form a right hand system.}, the camera frame \nm{\FC}\footnote{The camera frame \nm{\FC} is centered at the optical center (appendix \ref{sec:OpticalFlow}), with \nm{\iCiii} located in the camera principal axis pointing forward, and \nm{\iCi, \, \iCii} parallel to the focal plane.}, and the image frame \nm{\FIMG}\footnote{The two-dimensional image frame \nm{\FIMG} is centered at the sensor corner with axes parallel to the sensor borders (appendix \ref{sec:OpticalFlow}).}. Superindexes are employed over vectors to specify the reference frame in which they are viewed (e.g., \nm{\vN} refers to ground velocity viewed in \nm{\FN}, while \nm{\vB} is the same vector but viewed in \nm{\FB}). Subindexes may be employed to clarify the meaning of the variable or vector, such as in \nm{\vTAS} for air velocity instead of the ground velocity \nm{\vec v}, in which case the subindex is either an acronym or its  meaning is clearly explained when first introduced. Subindexes may also refer to a given component of a vector, e.g. \nm{\vNii} refers to the second component of \nm{\vN}. In addition, where two reference frames appear as subindexes to a vector, it means that the vector goes from the first frame to the second. For example, \nm{\wNBB} refers to the angular velocity from the \nm{\FN} frame to the \nm{\FB} frame viewed in \nm{\FB}. 

In addition, there exist various indexes that appear as subindexes: \emph{j} identifies an specific run within each Monte Carlo simulation, \emph{n} identifies a discrete time instant (\nm{t_n}) for the inertial estimations, \emph{s} (\nm{t_s}) refers to the sensor outputs, \emph{i} identifies an image or frame (\nm{t_i}), and \emph{k} is employed for the keyframes used to generate the map or terrain structure. Other employed subindexes are \emph{l} for the steps of the various iteration processes that take place, and \emph{j} for the features and associated 3D points. With respect to superindexes, two stars \nm{< \cdot^{\star\star} >} represent the reprojection only solution, while two circles \nm{< \cdot^{\circ\circ} >} identify a target.


\section*{Acronyms}

\begin{table}[ht]
\begin{tabular}{lp{6.4cm}p{0.1cm}lp{6.4cm}}
  \hypertarget{BRIEF}{\texttt{BRIEF}}     & Binary Robust Independent Elementary    & & \hypertarget{MX}{\texttt{MX}}		    & MiX	terrain type						 \\   
	                                        & Features	                  						& &	\hypertarget{NED}{\texttt{NED}}		  & North East Down								\\ 
  \hypertarget{DS}{\texttt{DS}}		        & DeSert terrain type	       	            & & \hypertarget{NSE}{\texttt{NSE}}		  & Navigation System Error						\\	
  \hypertarget{DSO}{\texttt{DSO}}		      & Direct Sparse Odometry                  & &	\hypertarget{OKVIS}{\texttt{OKVIS}} & Open Keyframe Visual Inertial \hypertt{SLAM}  \\ 
  \hypertarget{ECEF}{\texttt{ECEF}}	  		&	Earth Centered Earth Fixed							& & \hypertarget{ORB}{\texttt{ORB}}		  & Oriented \hypertt{FAST} and rotated \hypertt{BRIEF} \\ 
  \hypertarget{EKF}{\texttt{EKF}}		      &	Extended Kalman Filter						      & & \hypertarget{PI}{\texttt{PI}}		    & Proportional Integral							\\
  \hypertarget{FAST}{\texttt{FAST}}       & Features from Accelerated Segment Test  & & \hypertarget{PR}{\texttt{PR}}		    & PRaire terrain type										\\ 
  \hypertarget{FM}{\texttt{FM}}		        & FarM terrain type		                    & & \hypertarget{RANSAC}{\texttt{RANSAC}} & RANdom SAmple Consensus						\\
  \hypertarget{FR}{\texttt{FR}}		        & FoRest terrain type		                	& & \hypertarget{ROC}{\texttt{ROC}}		  & Rate Of Climb									\\
  \hypertarget{GNSS}{\texttt{GNSS}}	      & Global Navigation Satellite System	    & & \hypertarget{ROVIO}{\texttt{ROVIO}} & Robust Visual Inertial Odometry				\\
  \hypertarget{IA-VNS}{\texttt{IA-VNS}}   & Inertially Assisted \hypertt{VNS}   	  & & \hypertarget{SLAM}{\texttt{SLAM}}	  & Simultaneous Localization And Mapping			\\
  \hypertarget{IA-VNSE}{\texttt{IA-VNSE}} & Inertially Assisted Visual Navigation		& & \hypertarget{SLERP}{\texttt{SLERP}}	& Spherical linear interpolation				\\
                                          & System Error                 	        	& & \hypertarget{SVO}{\texttt{SVO}}		  & Semi direct Visual Odometry					\\
  \hypertarget{IMU}{\texttt{IMU}}		     	& Inertial Measurement Unit   						& & \hypertarget{SWaP}{\texttt{SWaP}}	  & Size, Weight, and Power						\\	
  \hypertarget{INS}{\texttt{INS}}         & Inertial Navigation System          		& & \hypertarget{TAS}{\texttt{TAS}}		  & True Air Speed								\\
  \hypertarget{INSE}{\texttt{INSE}}       & Inertial Navigation System Error       	& &	\hypertarget{UAV}{\texttt{UAV}}		  & Unmanned Aerial Vehicle						\\
  \hypertarget{iSAM}{\texttt{iSAM}}       & Incremental Smoothing And Mapping	 	 		& & \hypertarget{UR}{\texttt{UR}}		    & URban terrain type						\\
  \hypertarget{ISO}{\texttt{ISO}}	        & International Organization for	    		& & \hypertarget{USA}{\texttt{USA}}	    & United States of America	\\
			                                    & Standardization					         				& & \hypertarget{VINS}{\texttt{VINS}}	  & Visual Inertial Navigation System	\\
  \hypertarget{LSD}{\texttt{LSD}} 	      & Large scale Semi Direct           			& & \hypertarget{VIO}{\texttt{VIO}}			& Visual Inertial Odometry      \\                                         
	\hypertarget{MAV}{\texttt{MAV}}			    & Micro Air Vehicle		          	  			& & \hypertarget{VNS}{\texttt{VNS}}	  	& Visual Navigation System			\\	                                     
	\hypertarget{MSCKF}{\texttt{MSCKF}}     & Multi State Constraint Kalman Filter    & & \hypertarget{VNSE}{\texttt{VNSE}}	  & Visual Navigation System Error \\                                     
	\hypertarget{MSF}{\texttt{MSF}}	        & Multi Sensor Fusion		   				    	  & & \hypertarget{VO}{\texttt{VO}}	    	& Visual Odometry								 \\                                         
  \hypertarget{MSL}{\texttt{MSL}}		      & Mean Sea Level	                        & & \hypertarget{WGS84}{\texttt{WGS84}}	& World Geodetic System 1984		 \\                                          
\end{tabular}
\end{table}


\section{Introduction and Outline}\label{sec:Outline}

The extreme dependency of an autonomous \hypertt{UAV} (Unmanned Air Vehicle) navigation to the presence of \hypertt{GNSS} (Global Navigation Satellite System) signals, without which it incurs in a slow but unavoidable position drift that may ultimately lead to the loss of the platform, is discussed in section \ref{sec:GNSS-Denied}. This section also reviews the current approaches to \hypertt{GNSS}-Denied navigation, with emphasis on those based on the images taken by onboard cameras. 

\begin{figure}[h]
\centering
\begin{tikzpicture}[auto, node distance=2cm,>=latex']
	\node [coordinate](midinput) {};
	\node [coordinate, above of=midinput, node distance=0.4cm] (xSENSEDinput){};
	\node [coordinate, below of=midinput, node distance=0.4cm] (xESTzeroinput){};
	\node [block, right of=midinput, minimum width=3.0cm, node distance=8.0cm, align=center, minimum height=1.5cm] (NAVIGATION) {\texttt{INERTIAL} \\ \texttt{NAVIGATION}};
	\node [coordinate, right of=NAVIGATION, node distance=4.5cm] (output){};
		
	\draw [->] (xSENSEDinput) -- node[pos=0.47] {\nm{\xvectilde\lrp{t_s} \setminus \vec I\lrp{t_i} = \xSENSED\lrp{t_s} \setminus \vec I\lrp{t_i}}} ($(NAVIGATION.west)+(0cm,0.4cm)$);
	\draw [->] (xESTzeroinput) -- node[pos=0.1] {\nm{\xvecestzero}} ($(NAVIGATION.west)-(0cm,0.4cm)$);
	\draw [->] (NAVIGATION.east) -- node[pos=0.5] {\nm{\xvecest\lrp{t_n} = \xEST\lrp{t_n}}} (output);
\end{tikzpicture}
\caption{\texttt{INS} flow diagram}
\label{fig:flow_diagram_ins}
\end{figure}
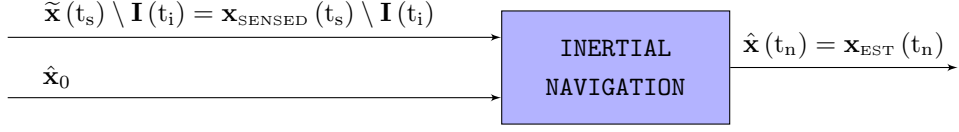

This article focuses on the need to develop navigation systems capable of diminishing the position drift inherent to the flight in \hypertt{GNSS}-Denied conditions of an autonomous fixed wing low \hypertt{SWaP} (Size, Weight, and Power) aircraft so it has a higher probability of reaching the vicinity of a recovery point, from where it can be landed by remote control. A previous article by the same authors, \cite{INSE}, showed that it is possible to take advantage of sensors already present onboard fixed wing aircraft\footnote{These include accelerometers, gyroscopes, magnetometers, Pitot tube, air vanes, thermometer, and barometer.}, the particularities of fixed wing flight, and the atmospheric and wind estimations that can be obtained before the \hypertt{GNSS} signals are lost, to develop an \hypertt{EKF} (Extended Kalman Filter) based \hypertt{INS} (Inertial Navigation System) that results in bounded (no drift) estimations for attitude\footnote{A bounded attitude estimation ensures that the aircraft can remain aloft in \hypertt{GNSS}-Denied conditions for as long as there is fuel available.}, altitude\footnote{The altitude estimation error depends on the change in atmospheric pressure offset \nm{\Deltap} \cite{INSA} from its value at the time the \hypertt{GNSS} signals are lost, which is bounded by atmospheric physics.}, and ground velocity\footnote{The ground velocity estimation error depends on the change in wind velocity from its value at the time the \hypertt{GNSS} signals are lost, which is bounded by atmospheric physics.}, as well as an unavoidable drift in horizontal position caused by integrating the ground velocity without absolute observations. Figure \ref{fig:flow_diagram_ins} graphically depicts that the \hypertt{INS} inputs include all sensor measurements with the exception of the camera images \nm{\vec I}. 

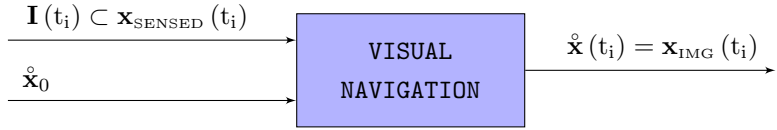
\begin{figure}[h]
\centering
\begin{tikzpicture}[auto, node distance=2cm,>=latex']
	\node [coordinate](midinput) {};
	\node [coordinate, above of=midinput, node distance=0.4cm] (IMAGESinput){};
	\node [coordinate, below of=midinput, node distance=0.4cm] (xVISSTARzeroinput){};
			
	\node [block, right of=midinput, minimum width=3.0cm, node distance=5.3cm, align=center, minimum height=1.5cm] (NAVIGATION) {\texttt{VISUAL} \\ \texttt{NAVIGATION}};
	\node [coordinate, right of=NAVIGATION, node distance=4.8cm] (output){};
		
	\draw [->] (IMAGESinput) -- node[pos=0.45] {\nm{\vec I\lrp{t_i} \subset \xSENSED\lrp{t_i}}} ($(NAVIGATION.west)+(0cm,0.4cm)$);
	\draw [->] (xVISSTARzeroinput) -- node[pos=0.10] {\nm{\xvecviszero}} ($(NAVIGATION.west)-(0cm,0.4cm)$);
		
	\draw [->] (NAVIGATION.east) -- node[pos=0.55] {\nm{\xvecvis\lrp{t_i} = \xIMG\lrp{t_i}}} (output);
\end{tikzpicture}
\caption{\texttt{VNS} flow diagram}
\label{fig:flow_diagram_vns}
\end{figure}

This article focuses on visual navigation, this is, that based on the images periodically taken by a down facing camera rigidly attached to the aircraft structure. The objectives, novelty, and applications of the proposed approach are discussed in section \ref{sec:Objectives}. A \hypertt{VNS} (Visual Navigation System) based on an advanced publicly available \hypertt{VO} (Visual Odometry) algorithm known as \hypertt{SVO} (Semi-Direct Visual Odometry) is introduced in section \ref{sec:SVO} and employed as a baseline. Visual navigation does not rely on absolute references and hence slowly accumulates error (drifts) not only in horizontal position, but also in attitude and altitude. Figure \ref{fig:flow_diagram_vns} shows that the \hypertt{VNS} relies exclusively on the images \nm{\vec I\lrp{t_i}} as well as the initial visual estimations \nm{\xvecviszero}, as explained in section \ref{sec:SVO}.

Section \ref{sec:IA-SVO} describes the proposed \hypertt{IA-VNS} (Inertially Assisted \hypertt{VNS}), which employs the \hypertt{INS} attitude and altitude estimations to introduce priors into the pose estimation non linear optimizations present in \hypertt{SVO}, resulting not only in significant improvements in the visual attitude and altitude estimations, but specially in a major reduction in horizontal position drift, which is a full order of magnitude lower than the drifts obtained by either the \hypertt{INS} or the \hypertt{VNS} by themselves. The \hypertt{IA-VNS} is graphically depicted in figure \ref{fig:flow_diagram_iavns}.
\begin{figure}[h]
\centering
\begin{tikzpicture}[auto, node distance=2cm,>=latex']
	\node [coordinate](midinput) {};
	\node [coordinate, above of=midinput, node distance=0.0cm] (xESTzeroinput0) {};
	\node [coordinate, above of=midinput, node distance=2.2cm] (xVISSTARzeroinput) {};
	\node [coordinate, above of=midinput, node distance=0.4cm] (xSENSEDinput){};
	
	\node [block, right of=midinput, minimum width=3.0cm, node distance=5.6cm, align=center, minimum height=1.5cm] (NAVIGATION) {\texttt{INERTIAL} \\ \texttt{NAVIGATION}};
	\node [block, above of=NAVIGATION, minimum width=4.5cm, node distance=2.5cm, align=center, minimum height=1.5cm] (VISNAVIGATION) {\texttt{INERTIALLY ASSISTED} \\ \texttt{VISUAL NAVIGATION}};
	
	\node [coordinate, right of=NAVIGATION, node distance=2.5cm] (midpoint){};
	\filldraw [black] (midpoint) circle [radius=1pt];
	\node [coordinate, above of=midpoint, node distance=1.2cm] (highpoint){};
		
	\draw [->]  (xESTzeroinput0) -- node[pos=0.08] {\nm{\xvecestzero}} (NAVIGATION.west);
	\draw [->] (xVISSTARzeroinput) -- node[pos=0.10] {\nm{\xvecviszero}} ($(VISNAVIGATION.west)+(0cm,-0.3cm)$);
	
	\draw [->] (NAVIGATION.east) -- node[pos=0.6] {\nm{\xvecest\lrp{t_n} = \xEST\lrp{t_n}}} ($(NAVIGATION.east)+(4.8cm,0.0cm)$);
	\draw [->] (VISNAVIGATION.east) -- node[pos=0.6] {\nm{\xvecvis\lrp{t_i} = \xIMG\lrp{t_i}}} ($(VISNAVIGATION.east)+(4.0cm,0.0cm)$);
	\draw [->] (midpoint) -- (highpoint) -| (VISNAVIGATION.south);
	
	\node [coordinate, below of=midinput, node distance=1.6cm] (xSENSEDINERinput){};
	\node [coordinate, below of=NAVIGATION, node distance=1.6cm] (xSENSEDINERmiddle){};
	\draw [->] (xSENSEDINERinput) -- node[pos=0.45] {\nm{\xvectilde\lrp{t_s} \setminus \vec I\lrp{t_i} = \xSENSED\lrp{t_s} \setminus \vec I\lrp{t_i}}} (xSENSEDINERmiddle) -- (NAVIGATION.south);
	
	\node [coordinate, above of=midinput, node distance=3.5cm] (xSENSEDVISinput){};
	\node [coordinate, above of=VISNAVIGATION, node distance=1.0cm] (xSENSEDVISmiddle){};
	\draw [->] (xSENSEDVISinput) -- (xSENSEDVISmiddle) -- (VISNAVIGATION.north);
	\node at ($(midinput) + (+1.6cm,+3.1cm)$) {\nm{\vec I\lrp{t_i} \subset \xSENSED\lrp{t_i}}};
	
\end{tikzpicture}
\caption{\texttt{IA-VNS} flow diagram}
\label{fig:flow_diagram_iavns}
\end{figure}

Section \ref{sec:Simulation} introduces the stochastic high fidelity simulation employed to evaluate the navigation results by means of Monte Carlo executions of two scenarios\footnote{These are the same scenarios employed to evaluate the \hypertt{INS} in \cite{INSE}.} representative of the challenges of \hypertt{GNSS}-Denied navigation, which are described in detail in \cite{SIMULATION}. The results obtained when applying the proposed algorithms to these two \hypertt{GNSS}-Denied scenarios are described in section \ref{sec:Results}. Section section \ref{sec:Influence_terrain} discusses the sensitivity of the estimations to the type of terrain overflown by the aircraft, as the terrain texture (or lack of) and its elevation relief are key factors on the ability of the visual algorithms to detect and track terrain features. Last, the results are summarized for convenience in section \ref{sec:Summary}, while section \ref{sec:Conclusion} provides a short conclusion.


\section{GNSS-Denied Navigation}\label{sec:GNSS-Denied}

The number, variety, and applications of \hypertt{UAV}s have grown exponentially in the last few years, and the trend is expected to continue in the future \cite{Hassanalian2017,Shakhatreh2019}. This is particularly true in the case of low \hypertt{SWaP} vehicles because their reduced cost makes them suitable for a wide range of applications, both civil and military. \cite{Bijjahalli2020} presents a comprehensive review of low \hypertt{SWaP} \hypertt{UAV} navigation systems and the problems they face, including the degradation or absence of \hypertt{GNSS} signals.

Aircraft navigation has traditionally relied on the measurements provided by accelerometers, gyroscopes, and magnetometers, incurring in an slow but unbounded position drift that could only be stopped by triangulation with the use of external navigation (radio) aids. More recently, the introduction of satellite navigation (\hypertt{GNSS}) has completely removed the position drift and enabled autonomous inertial navigation in low \hypertt{SWaP} platforms \cite{Farrell2008, Groves2008, Chatfield1997}. On the negative side, low \hypertt{SWaP} inertial navigation exhibits an extreme dependency on the availability of \hypertt{GNSS} signals. If the signals are not present or can not be employed, inertial systems rely on dead reckoning, which results in position drift, with the aircraft slowly but steadily deviating from its intended route \cite{Elbanhawi2017}. The availability of \hypertt{GNSS} signals cannot be guaranteed; a throughout analysis of \hypertt{GNSS} threats and reasons for signal degradation is presented in \cite{Sabatini2017}. In \hypertt{GNSS}-Denied conditions, the vehicle is unable to fly its intended route or even return to a safe recovery location, which leads to the uncontrolled loss of the airframe if the \hypertt{GNSS} signals are not recovered before the aircraft runs out of fuel (or battery in case of electric vehicles). 

The extreme dependency on \hypertt{GNSS} availability is not only one of the main impediments for the introduction of small \hypertt{UAV}s in civil airspace, where it is not acceptable to have uncontrolled vehicles causing personal or material damage, but it also presents a significant drawback for military applications, as a single hull loss may compromise the onboard technology. At this time there are no comprehensive solutions to the operation of low \hypertt{SWaP} autonomous \hypertt{UAV}s in \hypertt{GNSS}-Denied scenarios, although the use of onboard cameras seems to be one of the most promising routes. Bigger and more expensive \hypertt{UAV}s, this is, with less stringent \hypertt{SWaP} requirements, can rely to some degree on more accurate Inertial Measurement Units or \hypertt{IMU}s (at the expense of \hypertt{SWaP}) and additional communications equipment to overcome this problem, but for most autonomous \hypertt{UAV}s, the permanent loss of the \hypertt{GNSS} signals is equivalent to losing the airframe in an uncontrolled way.


\subsection{Possible Approaches to GNSS-Denied Navigation}\label{subsec:GNSS-Denied_approach}

\emph{Inertial navigation} employs the periodic readings provided by the \hypertt{IMU} to estimate the pose of a moving object by means of dead reckoning or integration. On aircraft, inertial sensors are complemented by magnetometers and a barometer to add robustness to the inertial solution. Fixed wing aircraft are also equipped with a Pitot tube and air vanes required by their control system, although their measurements are usually not employed for navigation. Absolute references such as those provided by navigation radio aids or \hypertt{GNSS} receivers are required to remove the position drift inherent to inertial navigation.

Low \hypertt{SWaP} autonomous aircraft are too small to incorporate \emph{navigation aid receivers}, which in any case are not available over vast regions of the Earth, exhibiting an extreme dependency on the availability of \hypertt{GNSS} signals. A summary of the challenges of \hypertt{GNSS}-Denied navigation and the research efforts intended to improve its performance is provided by \cite{Tippitt2020}. There exist various approaches to mitigate this problem, with detailed reviews provided by \cite{INSE, Gyagenda2022}. Two promising techniques for completely eliminating the position drift are the use of \emph{signals of opportunity}\footnote{Existing signals originally intended for other purposes, such as those of television and cellular networks, can be employed to triangulate the aircraft position.} \cite{Kapoor2017, Coluccia2014, Goh2013}, and \emph{georegistration}\footnote{The position drift can be eliminated by matching landmarks or terrain features as viewed from the aircraft to preloaded data.} \cite{Couturier2020, Goforth2019, Ziaei2019, Wang2016}, also known as \emph{image registration}.


\subsection{Visual Navigation}\label{subsec:GNSS-Denied_visual}

\emph{Visual Odometry} (\hypertt{VO}) consists on employing the ground images generated by one or more onboard cameras without the use of prerecorded image databases or any other sensors, incrementally estimating the vehicle pose based on the changes that its motion induces on the images \cite{Scaramuzza2011, Fraundorfer2012, Scaramuzza2012}. It  requires sufficient illumination, dominance of static scene, enough texture, and scene overlap between consecutive images or frames. It can rely on a single camera\footnote{This article relies on monocular vision exclusively.} (monocular vision), in which case the motion can only be recovered up to a scale factor, or on various cameras (stereo vision), where the differences among the simultaneous images taken with the different cameras are employed to determine the scale. It has been employed for navigation of ground robots, road vehicles, and multirotors flying both indoors and outdoors.

The incremental concatenation of relative poses results in a slow but unbounded pose drift, which can only be eliminated if aided by \emph{Simultaneous Localization and Mapping} (\hypertt{SLAM}) \cite{Scaramuzza2017, Cadena2016}, a particular case of \hypertt{VO} in which the map of the already viewed terrain is stored and employed for loop closure in case it is revisited by the vehicle during its motion. In this sense, \hypertt{VO} only uses the map to improve the local consistency of the solution, while \hypertt{SLAM} is more concerned with its global consistency \cite{Scaramuzza2011}. The result is that \hypertt{SLAM} is potentially more accurate, but also slower, computationally more expensive, and less robust.

Modern stand-alone algorithms such as Semi Direct Visual Odometry (\hypertt{SVO}) \cite{Forster2014, Forster2016}, Direct Sparse Odometry (\hypertt{DSO}) \cite{Engel2018}, Large scale Semi Dense \hypertt{SLAM} (\hypertt{LSD}-\hypertt{SLAM}) \cite{Engel2014}, and large scale feature based \hypertt{SLAM} (\hypertt{ORB}-\hypertt{SLAM})\footnote{\hypertt{ORB} stands for Oriented \hypertt{FAST} and rotated \hypertt{BRIEF}, a type of blob feature.} \cite{Mur2015, Mur2017, Mur2017bis} are robust and exhibit a limited drift.

A typical \hypertt{VO} algorithm includes steps to obtain the images, detect and extract its features, either match or track those features\footnote{\hypertt{VO} algorithms can be divided into feature-based methods (also known as matching methods) and direct or tracking methods \cite{Scaramuzza2012}.}, estimate the relative motion between consecutive frames, concatenate them to obtain the full camera pose trajectory, and finally perform some local optimization (bundle adjustment) \cite{Scaramuzza2011}.


\subsection{Visual Inertial Navigation}\label{subsec:GNSS-Denied_visualinertial}

Estimating the aircraft pose based on both \hypertt{IMU}s and cameras represents the most promising solution to \hypertt{GNSS}-Denied navigation, in what is known as \emph{Visual Inertial Odometry} (\hypertt{VIO}) \cite{Scaramuzza2019, Huang2019}, which can also be combined with image registration to fully eliminate the remaining pose drift. Current \hypertt{VIO} implementations are also primarily intended for ground robots, multirotors, and road vehicles, and hence rely exclusively on the vehicle \hypertt{IMU} readings and the images taken by the onboard cameras, but do not use other sensors commonly found onboard fixed wing aircraft. \hypertt{VIO} has matured significantly in the last few years, with detailed reviews available in \cite{Scaramuzza2019, Huang2019, Stumberg2019, Feng2019, Alkaff2017}.

\hypertt{VIO} currently appears to represent the state of the art in \hypertt{GNSS}-Denied navigation for low \hypertt{SWaP} \hypertt{UAV}s \cite{Bijjahalli2020}. There exist several open source \hypertt{VIO} packages, such as the Multi State Constraint Kalman Filter (\hypertt{MSCKF}) \cite{Mourikis2007}, the Open Keyframe Visual Inertial \hypertt{SLAM} (\hypertt{OKVIS}) \cite{Leutenegger2013,Leutenegger2015}, the Robust Visual Inertial Odometry (\hypertt{ROVIO}) \cite{Bloesch2015}, the monocular Visual Inertial Navigation System (\hypertt{VINS-Mono}) \cite{Qin2017}, \hypertt{SVO} combined with Multi Sensor Fusion (\hypertt{MSF}) \cite{Forster2014, Forster2016, Lynen2013, Faessler2016}, and \hypertt{SVO} combined with Incremental Smoothing and Mapping (\hypertt{iSAM}) \cite{Forster2014, Forster2016, Forster2016c, Kaess2012}. All these open source pipelines are compared in \cite{Scaramuzza2019}, and their results when applied to the EuRoC \hypertt{MAV} data sets \cite{Burri2016} are discussed in \cite{Delmerico2018}. There also exist various other published \hypertt{VIO} pipelines with implementations that are not publicly available \cite{Mur2017c, Clark2017, Paul2017, Song2017, Solin2017, Houben2016, Eckenhoff2017}, and there are also others that remain fully proprietary.

The existing \hypertt{VIO} schemes can be broadly grouped into two paradigms: \emph{loosely coupled} pipelines process the measurements separately, resulting in independent visual and inertial pose estimations, which are then fused to get the final estimate; on the other hand, \emph{tightly coupled} methods compute the final pose estimation directly from the tracked image features and the \hypertt{IMU} outputs \cite{Scaramuzza2019, Huang2019}. Tightly coupled approaches usually result in higher accuracy, as they use all the information available and take advantage of the \hypertt{IMU} integration to predict the feature locations in the next frame. Loosely coupled methods, although less complex and more computationally efficient, lose information by decoupling the visual and inertial constraints, and are incapable of correcting the drift present in the visual estimator.

A different classification involves the number of images involved in each estimation \cite{Scaramuzza2019, Huang2019, Strasdat2010}, which is directly related with the resulting accuracy and computing demands. \emph{Batch algorithms}, also known as \emph{smoothers}, estimate multiple states simultaneously by solving a large nonlinear optimization problem or bundle adjustment, resulting in the highest possible accuracy. Valid techniques to limit the required computing resources include the reliance on a subset of the available frames (known as \emph{keyframes})\footnote{Common criteria to identify keyframes involve thresholds on the time, position, or pose increments from the previous keyframe.}, the separation of tracking and mapping into different threads, and the development of incremental smoothing techniques based on factor graphs \cite{Kaess2012}. Although employing all available states (\emph{full smoothing}) is sometimes feasible for very short trajectories, most pipelines rely on \emph{sliding window} or \emph{fixed lag smoothing}, in which the optimization relies exclusively on the measurements associated to the last few keyframes, discarding both the old keyframes as well as all other frames that have not been cataloged as keyframes. On the other hand, \emph{filtering algorithms} restrict the estimation process to the latest state; they require less resources but suffer from permanently dropping all previous information and a much harder identification and removal of outliers, both of which lead to error accumulation or drift. 

The success of any \hypertt{VIO} approach relies on an accurate calibration of the pose and time offsets between the \hypertt{IMU} and the camera \cite{Scaramuzza2019, Huang2019}. Additional challenges applicable to all pipelines include the different working frequencies of \hypertt{IMU}s and cameras, as well the initialization requirements to bootstrap the algorithms.


\section{Objective, Novelty, and Application}\label{sec:Objectives}


\subsection{Objective}\label{subsec:Objectives_obj}

The main objective of this article is to improve the \hypertt{GNSS}-Denied navigation capabilities of autonomous aircraft, so in case \hypertt{GNSS} signals become unavailable, they can continue their mission or safely fly to a predetermined recovery location. To do so, the proposed approach combines two different navigation algorithms, employing the outputs of an \hypertt{INS} specifically designed for the flight without \hypertt{GNSS} signals of an autonomous fixed wing low \hypertt{SWaP} aircraft \cite{INSE} to diminish the horizontal position drift generated by a \hypertt{VNS} that relies on an advanced visual odometry pipeline such as \hypertt{SVO} \cite{Forster2014, Forster2016}. Note that the \hypertt{INS} makes use of all onboard sensors except the camera, while the \hypertt{VNS} relies exclusively on the images provided by the camera. 

Each of the two systems by itself incurs in unrestricted and excessive horizontal position drift that renders them inappropriate for long term \hypertt{GNSS}-Denied navigation, but for different reasons: while in the \hypertt{INS} the drift is the result of integrating the bounded ground velocity estimations without absolute position observations, that of the \hypertt{VNS} originates on the slow but continuous accumulation of estimation errors between consecutive frames. The two systems however differ in their estimations of the aircraft attitude and altitude, as they are bounded for the \hypertt{INS} but also drift in the case of the \hypertt{VNS}. The proposed approach modifies the \hypertt{VNS} so in addition to the images it can also accept as inputs the \hypertt{INS} bounded attitude and altitude outputs, converting it into an Inertially Assisted \hypertt{VNS} or \hypertt{IA-VNS} with vastly improved horizontal position estimation capabilities. 


\subsection{Novelty}\label{subsec:Objectives_nov}

The \hypertt{VIO} solutions listed in section \ref{subsec:GNSS-Denied_visualinertial} are quite generic with respect to the platforms on which they are mounted, with most applications focused on ground vehicles, indoor robots, and multirotors, as well as with respect to the employed sensors, which are usually restricted to the \hypertt{IMU} (gyroscopes and accelerometers) and one or more cameras. This article focuses on an specific case (long distance \hypertt{GNSS}-Denied turbulent flight of fixed wing low \hypertt{SWaP} aircraft), and as such is simultaneously more restrictive but also takes advantage of the sensors already onboard these platforms, such as magnetometers, Pitot tube, and air vanes. In addition, and unlike the existing \hypertt{VIO} packages, the proposed solution assumes that \hypertt{GNSS} signals are present at the beginning of the flight. As described in detail in \cite{INSE}, these are key to the obtainment of the bounded attitude and altitude \hypertt{INS} outputs on which the proposed \hypertt{IA-VNS} relies. 

The proposed \hypertt{IA-VNS} method represents a novel approach to diminish the pose drift of a \hypertt{VO} pipeline by supplementing its pose estimation non linear optimizations with priors based on the bounded attitude and altitude outputs of a \hypertt{GNSS}-Denied inertial filter. The method is inspired in a \hypertt{PI} (Proportional Integral) control loop, in which the inertial attitude and altitude outputs act as targets to ensure that the visual estimations do not deviate in excess from their inertial counterparts, resulting in major reductions to not only the visual attitude and altitude estimation errors, but also to the drift in horizontal position.


\subsection{Application}\label{subsec:Objectives_app}

This articles proves that inertial and visual navigation systems can be combined in such a way that the resulting \hypertt{IA-VNS} incurs in a \hypertt{GNSS}-Denied horizontal position drift that is significantly smaller than what can be obtained by either system individually. In the case that \hypertt{GNSS} signals become unavailable in mid flight, \hypertt{GNSS}-Denied navigation is required for the platform to complete its mission or return to base without the absolute position and ground velocity observations provided by \hypertt{GNSS} receivers. As shown in the following sections, the proposed system can significantly increase the possibilities of the aircraft safely reaching the vicinity of the intended recovery location, from where it can be landed by remote control. 


\section{Semi-Direct Visual Odometry}\label{sec:SVO}

\emph{Semi-Direct Visual Odometry} (\hypertt{SVO}) \cite{Forster2014, Forster2016} is a publicly available advanced combination of feature-based and direct \hypertt{VO} techniques primarily intended towards the navigation of land robots, road vehicles, and multirotors, holding various advantages in terms of accuracy and speed over traditional \hypertt{VO} algorithms. By combining the best characteristics of both approaches while avoiding their weaknesses, it obtains high accuracy and robustness with a limited computational budget. This section provides a short summary of the \hypertt{SVO} pipeline, although the interested reader should refer to \cite{Forster2014, Forster2016} for a more detailed description; the pose optimization phase is however described in depth, as it is the focus of the proposed modifications described in section \ref{sec:IA-SVO}.
\begin{figure}[h]
\centering
\begin{tikzpicture}[auto, node distance=2cm,>=latex']
	\node [coordinate](map) {};
	\node [coordinate, above of=map, node distance=0.30cm] (coord_map) {};
	\node [coordinate, above of=map, node distance=1.35cm] (coord_map_fd) {};
	\node [coordinate, above of=map, node distance=0.45cm] (coord_map_mp) {};
	\node [coordinate, below of=map, node distance=0.45cm] (coord_map_df) {};
	\node [coordinate, below of=map, node distance=1.35cm] (coord_map_fa) {};
	
	\node [coordinate, left of=map,  node distance=4.50cm] (aux) {};
	\node [coordinate, above of=aux, node distance=0.30cm] (coord_aux) {};
	\node [coordinate, above of=aux, node distance=1.35cm] (coord_aux_in) {};
	\node [coordinate, above of=aux, node distance=0.45cm] (coord_aux_hm) {};
	\node [coordinate, below of=aux, node distance=0.45cm] (coord_aux_vs) {};
	
	\node [coordinate, right of=map, node distance=5.40cm] (mot) {};
	\node [coordinate, above of=mot, node distance=0.30cm] (coord_mot) {};
	\node [coordinate, above of=mot, node distance=1.35cm] (coord_mot_sia) {};
	\node [coordinate, above of=mot, node distance=0.45cm] (coord_mot_fa) {};
	\node [coordinate, below of=mot, node distance=0.45cm] (coord_mot_po) {};
	\node [coordinate, below of=mot, node distance=1.35cm] (coord_mot_so) {};

	\pgfdeclarelayer{background};
	\node [rectangle, draw=black!50, fill=black!8,   dashed, rounded corners, left of=coord_aux,  node distance=0.56cm, minimum width=4.7cm, minimum height=4.2cm] (AUXILIARY) {};
	\node [rectangle, draw=black!50, fill=red!30,    dashed, rounded corners, right of=coord_map, node distance=0.0cm, minimum width=4.5cm, minimum height=4.2cm] (MAPPING) {};
	\node [rectangle, draw=black!50, fill=yellow!30, dashed, rounded corners, right of=coord_mot, node distance=0.0cm, minimum width=5.4cm, minimum height=4.2cm] (MOTION) {};
	\pgfdeclarelayer{foreground};
	
	\node [rectangle, draw=black!50, fill=black!24, left of=coord_aux_in, minimum width=2.5cm, node distance=0.92cm, align=center, minimum height=0.7cm] (IN) {1. \texttt{Initialization}};
	\node [rectangle, draw=black!50, fill=black!24, left of=coord_aux_hm, minimum width=2.5cm, node distance=0.54cm, align=center, minimum height=0.7cm] (HM) {2. \texttt{Initial Homography}};
		
	\node [rectangle, draw=black!50, fill=blue!30, left of=coord_map_fd, minimum width=2.5cm, node distance=0.0cm,  align=center, minimum height=0.7cm] (FD) {1. \texttt{Feature Detection}};
	\node [rectangle, draw=black!50, fill=blue!30, left of=coord_map_mp, minimum width=2.0cm, node distance=0.90cm, align=center, minimum height=0.7cm] (MP) {2. \texttt{Mapping}};	
	\node [rectangle, draw=black!50, fill=blue!30, left of=coord_map_df, minimum width=2.5cm, node distance=0.46cm, align=center, minimum height=0.7cm] (DF) {3. \texttt{Depth Filter}};
	\node [rectangle, draw=black!50, fill=blue!30, left of=coord_map_fa, minimum width=2.5cm, node distance=0.0cm,  align=center, minimum height=0.7cm] (FD) {4. \texttt{Feature Alignment}};
		
	\node [rectangle, draw=black!50, fill=green!30, left of=coord_mot_sia, minimum width=2.5cm, node distance=0.0cm, align=center, minimum height=0.7cm] (SIA) {1. \texttt{Sparse Image Alignment}};
	\node [rectangle, draw=black!50, fill=green!30, left of=coord_mot_fa,  minimum width=2.5cm, node distance=0.46cm, align=center, minimum height=0.7cm] (FA) {2. \texttt{Feature Alignment}};
	\node [rectangle, draw=black!50, fill=green!30, left of=coord_mot_po,  minimum width=2.5cm, node distance=0.46cm, align=center, minimum height=0.7cm] (PO) {3. \texttt{Pose Optimization}};
	\node [rectangle, draw=black!50, fill=green!30, left of=coord_mot_so,  minimum width=2.5cm, node distance=0.0cm, align=center, minimum height=0.7cm] (SO)  {4. \texttt{Structure Optimization}};
		
	\node at ($(AUXILIARY.west) +(+1.15cm,1.8cm)$) {\texttt{AUXILIARY}};
	\node at ($(MAPPING.west) +(+1.63cm,1.8cm)$) {\texttt{MAPPING THREAD}};
	\node at ($(MOTION.west) +(+1.55cm,1.8cm)$) {\texttt{MOTION THREAD}};
	
\end{tikzpicture}
\caption{\texttt{SVO} threads and processes}
\label{fig:svo_threads_processes}
\end{figure}
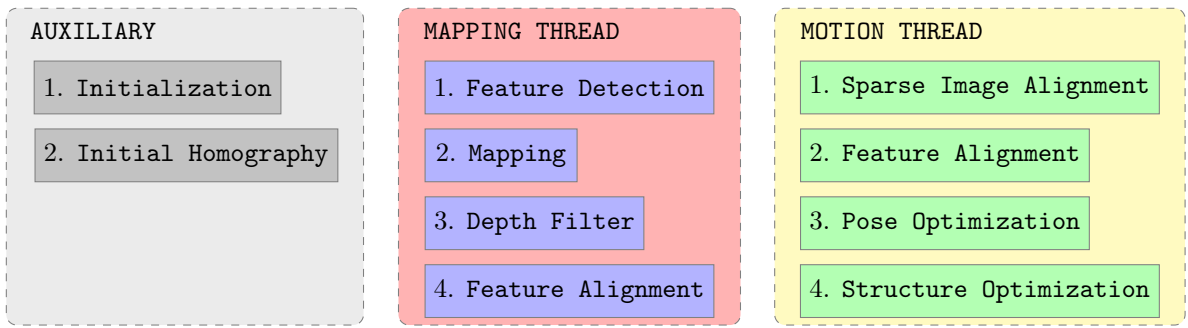

\hypertt{SVO} initializes like a feature-based monocular method, requiring the height over the terrain to provide the scale (initialization), and using feature matching and \hypertt{RANSAC} \cite{Fischler1981} based triangulation (initial homography) to obtain a first estimation of the terrain 3D position of the identified features. After initialization, the \hypertt{SVO} pipeline for each new image can be divided into two different threads: the \emph{mapping thread}, which generates terrain 3D points, and the \emph{motion thread}, which estimates the camera motion.

Once initialized, the expensive feature detection process (mapping thread) that obtains the features does not occur in every frame but only once a sufficiently large motion has occurred since the last feature extraction. When processing each new frame, \hypertt{SVO} initially behaves like a direct method, discarding the feature descriptors and skipping the matching process, and employing the luminosity values of small patches centered around every feature to (i) obtain a rough estimation of the camera pose (sparse image alignment, motion thread), followed by (ii) a relaxation of the epipolar restrictions to achieve a better estimation of the different features sub-pixel location in the new frame (feature alignment, motion thread), which introduces a reprojection residual that is exploited in the next steps. At this point, \hypertt{SVO} once again behaves like a feature-based method, refining (iii) the camera pose (pose optimization, motion thread) and (iv) the terrain coordinates of the 3D points associated to each feature (structure optimization, motion thread) based on nonlinear minimization of the reprojection error.

In this way, \hypertt{SVO} is capable of obtaining the accuracy of direct methods at a very high computational speed, due to only extracting features in selected frames, avoiding (for the most part) robust algorithms when tracking features, and only reconstructing the structure sparsely. The accuracy of \hypertt{SVO} improves if the pixel displacement between consecutive frames is reduced (high frame rate), which is generally possible as the computational expenses associated to each frame are low.

None of the motion thread four nonlinear optimization processes listed above makes use of \hypertt{RANSAC}, and pose optimization is the only one that employs a robust M-estimator \cite{Huber1981, Fox2013} instead of the traditional mean or squared error estimator. This has profound benefits in terms of computational speed but leaves the whole process vulnerable to the presence of outliers in either the features terrain or image positions. To prevent this, once a feature is detected in a given frame (note that the extraction process obtains pixel coordinates, not terrain 3D ones), it is immediately assigned with a depth filter (mapping thread) initialized with a large enough uncertainty around the average depth in the scene; in each subsequent frame, the feature 3D position is estimated by reprojection and the depth filter uncertainty reduced. Once the feature depth filter has converged, the detected feature and its associated 3D point become a map candidate, which it is not yet employed in the motion thread optimizations required to estimate the camera pose. The feature alignment process is however applied in the background to the map candidates, and it is only after several successful reprojections that a candidate is upgraded to a map 3D point and hence allowed to influence the motion result. This two step verification process that requires depth filter convergence and various successful reprojections before a 3D point is employed in the (mostly) non-robust optimizations is key to prevent outliers from contaminating the solution and reducing its accuracy.


\subsection{Pose Optimization}\label{subsec:SVO_pose}

\begin{figure}[h]
\centering
\begin{tikzpicture}[auto, node distance=2cm,>=latex']
	\node [coordinate](midinput) {};
	\node [coordinate, above of=midinput, node distance=0.8cm] (aboveinput) {};
	\node [coordinate, below of=midinput, node distance=0.8cm] (belowinput) {};

	\node [coordinate, above of=midinput, node distance=0.2cm] (refinput) {};
	
	\pgfdeclarelayer{background};
	\node [rectangle, draw=black!50, fill=yellow!30, dashed, rounded corners, right of=refinput, node distance=6.6cm, minimum width=10.2cm, minimum height=3.6cm] (MOTION) {};
	\pgfdeclarelayer{foreground};

	\node [blockgreen, right of=midinput, minimum width=2.0cm, node distance=10.2cm, align=center, minimum height=2.9cm] (PO) {\texttt{POSE} \\ \texttt{OPTIMIZATION} \\ \texttt{Frame} i \\ \nm{\sum_{\ds{j}}}};
	\node [rectangle, draw=black!50, fill=green!30, dashed, right of=aboveinput, minimum width=4.5cm, node distance=4.0cm, align=center, minimum height=1.0cm] (SIA) {\texttt{SPARSE IMAGE ALIGNMENT}};
	\node [rectangle, draw=black!50, fill=green!30, dashed, right of=belowinput, minimum width=4.5cm, node distance=4.0cm, align=center, minimum height=1.0cm] (FA) {\texttt{FEATURE ALIGNMENT}};

	\draw [->] (midinput) -- node[pos=0.05] {\nm{\vec p_j^{\sss E}}} (PO.west);
	\draw [->] ($(PO.east)+(+0.0cm,-0.0cm)$) -- node[pos=0.7] {\nm{\circled{\vec \zeta}_{{\sss EC}i}}} ($(PO.east)+(+1.8cm,-0.0cm)$);	
	\draw [dashed,->] (SIA.east) -- node[pos=0.4] {\nm{\circled{\vec \zeta}_{{\sss EC}i0} = \circled{\vec \zeta}_{{\sss EC}i}^{\star}}} ($(PO.west)+(+0.0cm,+0.8cm)$);
	\draw [->] (FA.east) -- node[pos=0.2] {\nm{\vec p_{ij}^{\sss IMG}}} ($(PO.west)+(+0.0cm,-0.8cm)$);
	
	\node at ($(MOTION.west) +(+2.1cm,1.50cm)$) {\texttt{MOTION THREAD Frame} i};
	
\end{tikzpicture}
\caption{Pose optimization flow diagram}
\label{fig:svo_motion_thread_pose_optim_flow_diagram}
\end{figure}
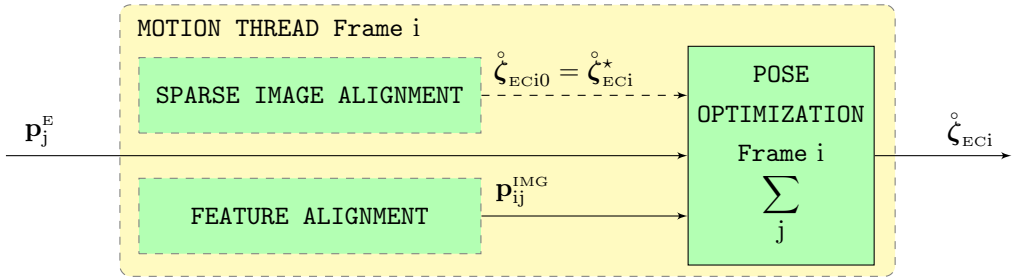

As the focus of the proposed improvements within section \ref{sec:IA-SVO}, the pose optimization phase is the only one described in detail in this article. Graphically depicted in figure \ref{fig:svo_motion_thread_pose_optim_flow_diagram}, pose optimization is executed for every new frame \emph{i} and makes use of the different features and associated 3D points \emph{j} present in the map to generate the visual estimation of the pose between the \hypertt{ECEF} (\nm{\FE}) and camera (\nm{\FC}) frames \nm{\circled{\vec \zeta}_{{\sss EC}i}}. 

The previous feature alignment phase obtains the projection of the feature points into the current image \nm{\vec p_{ij}^{\sss IMG}} by minimizing the photometric error while ignoring their epipolar constraints, which enables fine tuning the camera pose \nm{\circled{\vec \zeta}_{{\sss EC}i}^{\star}} provided by the sparse image alignment phase by minimizing the reprojection error of the different terrain 3D points \nm{\vec p_j^{\sss E}} present in the map, which originate at the structure optimization phase of the previous frame. This process is known as \emph{pose refinement} or \emph{motion-only bundle adjustment} \cite{Forster2014}. 

The \emph {reprojection error} \nm{E_{RP}} is defined in (\ref{eq:svo_pose_optim_problem_general}) as the sum for each 3D point of the norm of the difference between the camera projection \nm{\vec \Pi} of the \hypertt{ECEF} coordinates \nm{\vec p_{j}^{\sss E}} transformed into the camera frame\footnote{\nm{\vec g_{\ds{\vec \zeta_{{\sss AB}}}}\lrp{}} represents the \nm{\mathbb{SE}(3)} transformation of a point from frame B to frame A, as described in \cite{LIE}.} and the image coordinates \nm{\vec p_{j}^{\sss IMG}}. 
\neweq{E_{RP}\lrp{\zetaECivis} = \sum_j \biggr\lVert \vec \Pi\big(\vec g_{\ds{\circled{\vec \zeta}_{{\sss EC}i}}}^{-1}(\vec p_{j}^{\sss E})\big) - \vec p_{ij}^{\sss IMG}\biggr\rVert}{eq:svo_pose_optim_problem_general}

This problem can be solved by means of an iterative Gauss-Newton gradient descent process \cite{LIE,Baker2004}. Given an initial pose estimation \nm{\circled{\vec \zeta}_{{\sss EC}i0}} taken from the sparse image alignment result (\nm{\circled{\vec \zeta}_{{\sss EC}i}^{\star}}), each iteration step \emph{l} minimizes (\ref{eq:svo_pose_optim_problem}) and advances the estimated solution by means of (\ref{eq:svo_pose_optim_iterative}) until the step diminution of the reprojection error falls below a given threshold \nm{\lrp{E_{RP,i,l} - E_{RP,i,l+1} < \delta_{RP}}}:
\begin{eqnarray}
\nm{E_{RPi,l+1}\lrp{\Delta \circled{\vec \tau}_{{\sss EC}il}^{\sss C}}} & = & \nm{\sum_j \biggr\lVert \vec \Pi\big(\vec g_{\ds{\circled{\vec \zeta}_{{\sss EC}il}} \oplus \Delta \circled{\vec \tau}_{{\sss EC}il}^{\sss C}}^{-1}(\vec p_{j}^{\sss E})\big) - \vec p_{ij}^{\sss IMG} \biggr\rVert = \sum_j \Bigr\lVert \vec E_{RPi,l+1,j}\lrp{\Delta \circled{\vec \tau}_{{\sss EC}il}^{\sss C}}  \Bigr\rVert} \label{eq:svo_pose_optim_problem} \\
\nm{\circled{\vec \zeta}_{{\sss EC}i,l+1}} & \nm{\longleftarrow} & \nm{\circled{\vec \zeta}_{{\sss EC}il} \circ Exp\lrp{\Delta \circled{\vec \tau}_{{\sss EC}il}^{\sss C}} = \circled{\vec \zeta}_{{\sss EC}il} \oplus \Delta \circled{\vec \tau}_{{\sss EC}il}^{\sss C}} \label{eq:svo_pose_optim_iterative}
\end{eqnarray}

Each \nm{\Delta \circled{\vec \tau}_{{\sss EC}il}^{\sss C}} represents the update to the camera pose given by the transform vector \cite{LIE,Sola2018} viewed in the local camera frame \nm{F_{{\sss C}il}}, which is obtained by following the process described in \cite{LIE,Baker2004}, and results in (\ref{eq:svo_pose_optim_solution}), where \nm{\vec J_{OF,ilj}} (\ref{eq:svo_pose_optim_jacobian}) represents the optical flow obtained in appendix \ref{sec:OpticalFlow}:
\begin{eqnarray}
\nm{\Delta \circled{\vec \tau}_{{\sss EC}il}^{\sss C}} & = & \nm{- \bigg[\sum_j {\vec J_{OF,ilj}}^T \ \vec J_{OF,ilj}\bigg]^{-1} \ \sum_j {\vec J_{OF,ilj}}^T \ \Big[\vec \Pi\big(\vec g_{\ds{\circled{\vec \zeta}_{{\sss EC}il}}}^{-1}(\vec p_{j}^{\sss E})\big) - \vec p_{ij}^{\sss IMG}\Big] \ \in \mathbb{R}^6} \label{eq:svo_pose_optim_solution} \\
\nm{\vec J_{OF,ilj}} & = & \nm{\vec J_{OF}\Big(\vec \Pi\big(\vec g_{\circled{\vec \zeta}_{{\sss EC}il}}^{-1}\lrp{\vec p_j^{\sss E}\big)}\Big) \ \  \ \in \mathbb{R}^{2x6}} \label{eq:svo_pose_optim_jacobian} 
\end{eqnarray}

In order to protect the resulting pose from the possible presence of outliers in either the 3D points \nm{\vec p_{j}^{\sss E}} or their image projection \nm{\vec p_{ij}^{\sss IMG}}, it is better to replace the above squared error or mean estimator by a more robust M-estimator, such as the bisquare or Tukey estimator \cite{Huber1981, Fox2013}. The error to be minimized in each iteration step is then given by (\ref{eq:svo_pose_optim_problem_tukey}), where the Tukey error function \nm{\varrho_{TUK}\lrp{x} = \rho_{TUK}\lrp{\sqrt{x}}} can be found in \cite{Fox2013}. 
\begin{eqnarray}
\nm{E_{RPi,l+1}\lrp{\Delta \circled{\vec \tau}_{{\sss EC}il}^{\sss C}}} & = & \nm{\sum_j \varrho_{TUK}\lrp{ \Big[\vec \Pi\big(\vec g_{\ds{\circled{\vec \zeta}_{{\sss EC}il}} \oplus \Delta \circled{\vec \tau}_{{\sss EC}il}^{\sss C}}^{-1}(\vec p_{j}^{\sss E})\big) - \vec p_{ij}^{\sss IMG} \Big]^T \Big[\vec \Pi\big(\vec g_{\ds{\circled{\vec \zeta}_{{\sss EC}il}} \oplus \Delta \circled{\vec \tau}_{{\sss EC}il}^{\sss C}}^{-1}(\vec p_{j}^{\sss E})\big) - \vec p_{ij}^{\sss IMG} \Big]}} \nonumber \\
& = & \nm{\sum_j \varrho_{TUK}\lrp{{\vec E_{RPi,l+1,j}}^T \, \vec E_{RPi,l+1,j}}} \label{eq:svo_pose_optim_problem_tukey} 
\end{eqnarray}

A similar process to that employed above leads to the solution (\ref{eq:svo_pose_optim_solution_tukey}), where the Tukey weight function \nm{w_{TUK}\lrp{x}} is provided by \cite{Fox2013}:
\begin{eqnarray}
\nm{\Delta \circled{\vec \tau}_{{\sss EC}il}^{\sss C}} & = & \nm{- \bigg[\sum_j w_{TUK}\lrp{{\vec E_{RP,ilj}}^T \vec E_{RP,ilj}} {\vec J_{OF,ilj}}^T \vec J_{OF,ilj}\bigg]^{-1}} \nonumber \\
 &  & \nm{\ \ \bigg[\sum_j w_{TUK}\lrp{{\vec E_{RP,ilj}}^T \vec E_{RP,ilj}} {\vec J_{OF,ilj}}^T \vec E_{RP,ilj}\bigg] \ \in \mathbb{R}^6} \label{eq:svo_pose_optim_solution_tukey} \\
\nm{\vec E_{RP,ilj}} & = & \nm{ \vec \Pi\big(\vec g_{\ds{\circled{\vec \zeta}_{{\sss EC}il}}}^{-1}(\vec p_{j}^{\sss E})\big) - \vec p_{ij}^{\sss IMG} \ \in \mathbb{R}^2} \label{eq:svo_pose_optim_comp}
\end{eqnarray}


\section{Proposed Inertially Assisted Semi-Direct Visual Odometry}\label{sec:IA-SVO}

Lacking any absolute references, the \hypertt{SVO} based \hypertt{VNS} described in section \ref{sec:SVO} gradually accumulates errors in each of the six dimensions of the vehicle pose \nm{\zetaEBvis}, as shown in section \ref{sec:Results}. If accurate estimations of attitude and altitude can be provided by an \hypertt{INS} such as that described in \cite{INSE}, these can be employed to ensure that the visual estimations for body attitude and vertical position (\nm{\qNBvis} and \nm{\hvis}) do not deviate in excess from their inertial counterparts \nm{\qNBest} and \nm{\hest}, so they too can be considered as bounded up to a certain degree. 
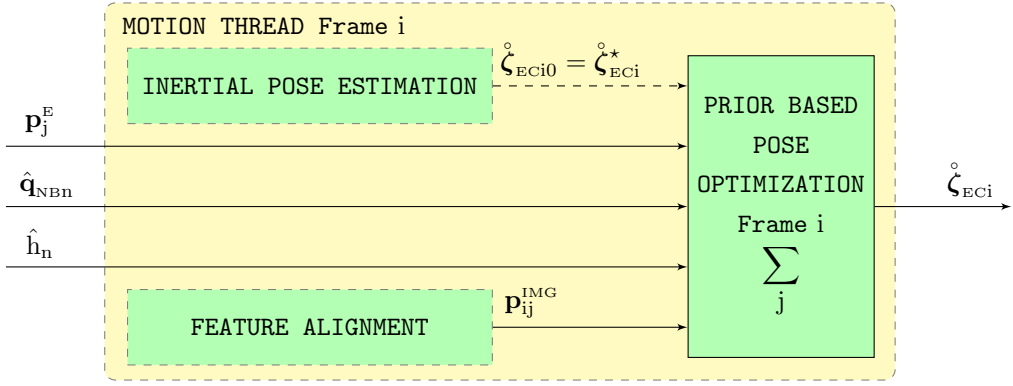
\begin{figure}[h]
\centering
\begin{tikzpicture}[auto, node distance=2cm,>=latex']
	\node [coordinate](midinput) {};
	\node [coordinate, above of=midinput, node distance=1.6cm] (aboveinput_two) {};
	\node [coordinate, above of=midinput, node distance=0.8cm] (aboveinput_one) {};
	\node [coordinate, below of=midinput, node distance=0.8cm] (belowinput_one) {};
	\node [coordinate, below of=midinput, node distance=1.6cm] (belowinput_two) {};
	
	\node [coordinate, above of=midinput, node distance=0.2cm] (refinput) {};
	
	\pgfdeclarelayer{background};
	\node [rectangle, draw=black!50, fill=yellow!30, dashed, rounded corners, right of=refinput, node distance=6.5cm, minimum width=10.4cm, minimum height=5.0cm] (MOTION) {};
	\pgfdeclarelayer{foreground};

	\node [blockgreen, right of=midinput, minimum width=2.0cm, node distance=10.2cm, align=center, minimum height=4.0cm] (PO) {\texttt{PRIOR BASED} \\ \texttt{POSE} \\ \texttt{OPTIMIZATION} \\ \texttt{Frame} i \\ \nm{\sum_{\ds{j}}}};
	\node [rectangle, draw=black!50, fill=green!30, dashed, right of=aboveinput_two, minimum width=4.8cm, node distance=4.0cm, align=center, minimum height=1.0cm] (SIA) {\texttt{INERTIAL POSE ESTIMATION}};
	\node [rectangle, draw=black!50, fill=green!30, dashed, right of=belowinput_two, minimum width=4.8cm, node distance=4.0cm, align=center, minimum height=1.0cm] (FA) {\texttt{FEATURE ALIGNMENT}};

	\draw [->] (aboveinput_one) -- node[pos=0.05] {\nm{\vec p_j^{\sss E}}} ($(PO.west)+(+0.0cm,+0.8cm)$);
	\draw [->] (midinput) -- node[pos=0.06] {\nm{\hat{\vec q}_{{\sss NB}n}}} (PO.west);
	\draw [->] (belowinput_one) -- node[pos=0.05] {\nm{\hat{h}_n}} ($(PO.west)+(+0.0cm,-0.8cm)$);

	\draw [->] ($(PO.east)+(+0.0cm,-0.0cm)$) -- node[pos=0.7] {\nm{\circled{\vec \zeta}_{{\sss EC}i}}} ($(PO.east)+(+1.8cm,-0.0cm)$);	
	\draw [dashed,->] (SIA.east) -- node[pos=0.4] {\nm{\circled{\vec \zeta}_{{\sss EC}i0} = \circled{\vec \zeta}_{{\sss EC}i}^{\star}}} ($(PO.west)+(+0.0cm,+1.6cm)$);
	\draw [->] (FA.east) -- node[pos=0.2] {\nm{\vec p_{ij}^{\sss IMG}}} ($(PO.west)+(+0.0cm,-1.6cm)$);
	
	\node at ($(MOTION.west) +(+2.1cm,2.20cm)$) {\texttt{MOTION THREAD Frame} i};
	
\end{tikzpicture}
\caption{Prior based pose optimization flow diagram}
\label{fig:assist_svo_motion_thread_pose_optim_flow_diagram}
\end{figure}

Note that the inertial estimations should not replace the visual ones within \hypertt{SVO}, as this would destabilize the visual pipeline preventing its convergence, but just act as anchors so the visual estimations oscillate freely as a result of the multiple \hypertt{SVO} optimizations but without drifting from the vicinity of the anchors. This section shows how to modify the cost function within the iterative Gauss-Newton gradient descent pose optimization phase so it can take advantage of the \hypertt{INS} outputs.


\subsection{Rationale for the Introduction of Priors}\label{subsec:IA-SVO_rationale}

The prior based pose optimization process starts by executing exactly the same pose optimization described in section \ref{subsec:SVO_pose}, which seeks to obtain the camera pose \nm{\circled{\vec \zeta}_{{\sss EC}i}} that minimizes the reprojection error \nm{E_{RP}} (\ref{eq:svo_pose_optim_problem_general}). The iterative optimization results in a series of transform vector updates \nm{\Delta \circled{\vec \tau}_{{\sss EC}il}^{\sss C}} (\ref{eq:svo_pose_optim_solution_tukey}), where \emph{l} indicates the iteration step. The camera pose is then advanced per (\ref{eq:svo_pose_optim_iterative}) until the step diminution of the reprojection error falls below a certain threshold. 

The resulting camera pose, \nm{\circled{\vec \zeta}_{{\sss EC}i}}, is marked with the superindex \nm{\star\star} to indicate that it is the reprojection only solution, resulting in \nm{\circled{\vec \zeta}_{{\sss EC}i}^{\star\star}}. Its concatenation with the constant body to camera pose \nm{\vec \zeta_{{\sss BC}}} results in the reprojected body pose\footnote{Note that a single asterisk superindex \nm{< \cdot^{\ast} >} applied to a quaternion refers to its conjugate or inverse.} \nm{\circled{\vec \zeta}_{{\sss EB}i}^{\star\star}}:
\neweq{\circled{\vec \zeta}_{{\sss EB}i}^{\star\star} = \circled{\vec \zeta}_{{\sss EC}i}^{\star\star} \otimes \vec \zeta_{{\sss BC}}^{\ast}}  {eq:nav_vis_assist_pose_optim_zetaEBstarstar}

The reprojected \hypertt{ECEF} body attitude \nm{\circled{\vec q}_{{\sss EB}i}^{\star\star}} and Cartesian coordinates \nm{\circled{\vec T}_{{\sss EB}i}^{{\sss B}\star\star}} can then be readily obtained, which leads on one hand to the reprojected body attitude \nm{\circled{\vec q}_{{\sss NB}i}^{\star\star}}, equivalent to the Euler angles \nm{\circled{\vec \phi}_{{\sss NB}i}^{\star\star} = \lrsb{\circled{\psi}_i^{\star\star}, \circled{\theta}_i^{\star\star}, \circled{\xi}_i^{\star\star}}^T}, and on the other to the geodetic coordinates \nm{\circled{\vec T}_i^{{\sss E,GDT}\star\star}} (including the altitude \nm{\circled{h_i}^{\star\star}}) and \hypertt{ECEF} to \hypertt{NED} rotation \nm{\circled{\vec q}_{{\sss EN}i}^{\star\star}}. 

Let's assume for the time being that the observed body attitude (\nm{\hat{\vec q}_{{\sss NB}n}}) or altitude (\nm{\hat{h}_n}) estimated by the \hypertt{INS} \cite{INSE} enable the \hypertt{IA-VNS} to conclude that it would be preferred if the optimized body attitude were closer to a certain target attitude identified by the superindex \nm{\circ\circ}, \nm{\circled{\vec q}_{{\sss NB}i}^{\circ\circ}}, equivalent to the target Euler angles \nm{\circled{\vec \phi}_{{\sss NB}i}^{\circ\circ} = \lrsb{\circled{\psi}_i^{\circ\circ}, \circled{\theta}_i^{\circ\circ}, \circled{\xi}_i^{\circ\circ}}^T}. Section \ref{subsec:IA-SVO_pi} specifies when this assumption can be considered valid, as well as various alternatives to obtain the target attitude from \nm{\hat{\vec q}_{{\sss NB}n}} and \nm{\hat{h}_n}. The target body attitude is converted into a target camera attitude by means of the constant \nm{\qBC} and the original reprojected \hypertt{ECEF} to \hypertt{NED} rotation \nm{ \circled{\vec q}_{{\sss EN}i}^{\star\star}}, incurring in a negligible error by not considering the attitude change of the \hypertt{NED} frame as the iteration progresses.
\neweq{\circled{\vec q}_{{\sss EC}i}^{\circ\circ} = \circled{\vec q}_{{\sss EN}i}^{\star\star} \otimes \circled{\vec q}_{{\sss NB}i}^{\circ\circ} \otimes \qBC} {eq:nav_vis_assist_pose_optim_qECcirccirc}

Note that the objective is not for the resulting body attitude \nm{\circled{\vec q}_{{\sss NB}i}} to equal the target \nm{\circled{\vec q}_{{\sss NB}i}^{\circ\circ}}, but to balance both objectives (minimization of the reprojection error of the different terrain 3D points and minimization of the attitude differences with the targets) without imposing any hard constraints on the pose (position plus attitude) of the aircraft.


\subsection{Prior Based Pose Optimization}\label{subsec:IA-SVO_prior}

The \emph{attitude adjustment error} \nm{E_{q}} is defined in (\ref{eq:nav_vis_assist_pose_optim_pitch_objective}) as the norm of the Euclidean difference between rotation vectors corresponding to the camera attitude \nm{\circled{\vec q}_{{\sss EC}i}} and the target camera attitude \nm{\circled{\vec q}_{{\sss EC}i}^{\circ\circ}} \cite{LIE,Sola2018}:
\neweq{E_{q}\big(Log\lrp{\circled{\vec q}_{{\sss EC}i}}\big) = E_{q}\lrp{\circled{\vec r}_{{\sss EC}i}} = \Bigr\lVert Log\lrp{\circled{\vec q}_{{\sss EC}i}} - Log\lrp{\circled{\vec q}_{{\sss EC}i}^{\circ\circ}} \Bigr\rVert = \Bigr\lVert \circled{\vec r}_{{\sss EC}i} - \circled{\vec r}_{{\sss EC}i}^{\circ\circ} \Bigr\rVert}{eq:nav_vis_assist_pose_optim_pitch_objective}

Its minimization can be solved by means of an iterative Gauss-Newton gradient descent process \cite{LIE,Baker2004}. Given an initial attitude estimation \nm{\circled{\vec r}_{{\sss EC}i,0} = Log\lrp{\circled{\vec q}_{{\sss EC}i,0}}} taken from \nm{\circled{\vec \zeta}_{{\sss EC}i0} = \circled{\vec \zeta}_{{\sss EC}i}^{\star}}, each iteration step \emph{l} minimizes (\ref{eq:nav_vis_assist_pose_optim_pitch_problem}) and advances the estimated solution by means of (\ref{eq:nav_vis_assist_pose_optim_pitch_iterative}) until the step diminution of the attitude adjustment error falls below a given threshold \nm{\lrp{E_{q,i,l} - E_{q,i,l+1} < \delta_{q}}}:
\begin{eqnarray}
\nm{E_{q,i,l+1}\lrp{\Delta \circled{\vec r}_{{\sss EC}il}^{\sss C}}} & = & \nm{\Bigr\lVert Log\big(\circled{\vec q}_{{\sss EC}il} \oplus \Delta \circled{\vec r}_{{\sss EC}il}^{\sss C}\big) - Log\lrp{\circled{\vec q}_{{\sss EC}i}^{\circ\circ}}\Bigr\rVert} \nonumber \\
& = & \nm{\Bigr\lVert Log\big(Exp\lrp{\circled{\vec r}_{{\sss EC}il}} \oplus \Delta \circled{\vec r}_{{\sss EC}il}^{\sss C}\big) - \circled{\vec r}_{{\sss EC}i}^{\circ\circ}\Bigr\rVert} \label{eq:nav_vis_assist_pose_optim_pitch_problem} \\
\nm{\circled{\vec q}_{{\sss EC}i,l+1}} & \nm{\longleftarrow} & \nm{\circled{\vec q}_{{\sss EC}il} \circ Exp\lrp{\Delta \circled{\vec r}_{{\sss EC}il}^{\sss C}} = \circled{\vec q}_{{\sss EC}il} \oplus \Delta \circled{\vec r}_{{\sss EC}il}^{\sss C}} \label{eq:nav_vis_assist_pose_optim_pitch_iterative}
\end{eqnarray}

Each \nm{\Delta \circled{\vec r}_{{\sss EC}il}^{\sss C}} represents the update to the camera attitude given by the rotation vector viewed in the local camera frame \nm{F_{{\sss C}l}}, which is obtained by following the process described in \cite{LIE,Baker2004}\footnote{Note that in this process the Jacobian coincides with the identity matrix because the map \nm{\vec f\lrp{\circled{\vec r}_{{\sss EC}i}} = \circled{\vec r}_{{\sss EC}i}} coincides with the rotation vector itself.}, and results in (\ref{eq:nav_vis_assist_pose_optim_pitch_solution}), where \nm{\vec J_{Ril}} is the \nm{\mathbb{SO}(3)} right Jacobian \nm{\vec J_R\lrp{\vec r}} provided by \cite{LIE,Sola2018}\footnote{\cite{LIE,Sola2018} also include an expression for the right Jacobian inverse (\nm{\vec J_{Ril}^{-1} = \vec J_R^{-1}\lrp{\vec r}}).}. 
\begin{eqnarray}
\nm{\Delta \circled{\vec r}_{{\sss EC}il}^{\sss C}} & = & \nm{- \Big[\vec J_{Ril}^{-T} \ \vec J_{Ril}^{-1}\Big]^{-1} \ \vec J_{Ril}^{-T} \ \big[\circled{\vec r}_{{\sss EC}il} - \circled{\vec r}_{{\sss EC}i}^{\circ\circ} \big]} \nonumber \\
 & = & \nm{- \Big[\vec J_{Ril}^{-T} \ \vec J_{Ril}^{-1}\Big]^{-1} \ \vec J_{Ril}^{-T} \ \big[Log\big(\circled{\vec q}_{{\sss EC}il}\big) - Log\big(\circled{\vec q}_{{\sss EC}i}^{\circ\circ}\big) \big]} \nonumber \\
 & = & \nm{- \Big[\vec J_{Ril}^{-T} \ \vec J_{Ril}^{-1}\Big]^{-1} \ \vec J_{Ril}^{-T} \ {\vec E}_{q,il} \ \in \mathbb{R}^3} \label{eq:nav_vis_assist_pose_optim_pitch_solution} \\
\nm{\vec J_{Ril}} & = & \nm{\vec J_R\lrp{\circled{\vec r}_{{\sss EC}il}} = \vec J_R\Big(Log\lrp{\circled{\vec q}_{{\sss EC}il}}\Big) \ \ \ \in \mathbb{R}^{3x3}} \label{eq:nav_vis_assist_pose_optim_pitch_jacobian} 
\end{eqnarray}

The prior based pose adjustment algorithm attempts to obtain the \texttt{ECEF} camera pose \nm{\circled{\vec \zeta}_{{\sss EC}i}} that minimizes the reprojection error \nm{E_{RP}} discussed in section \ref{sec:SVO} combined with the weighted attitude adjustment error \nm{E_{q}}. The specific weight \nm{f_q} is discussed in section \ref{subsec:IA-SVO_pi}. Inspired in \cite{Baker2004_part4}, the main goal of the optimization algorithm is to minimize the reprojection error of the different terrain 3D points while simultaneously trying to be close to the attitude and altitude targets derived from the inertial filter.
\neweq{E_{PO}\lrp{\circled{\vec \zeta}_{{\sss EC}i}} = E_{RP}\lrp{\circled{\vec \zeta}_{{\sss EC}i}} + f_q \cdot E_{q}\big(\circled{\vec r}_{{\sss EC}i}\big)} {eq:nav_vis_assist_pose_optim_joint_objective}

Although the rotation vector \nm{\circled{\vec r}_{{\sss EC}i} = Log\lrp{\circled{\vec q}_{{\sss EC}i}}} can be directly obtained from the pose \nm{\circled{\vec \zeta}_{{\sss EC}i}} \cite{LIE,Sola2018}, merging the two algorithms requires a dimension change in the (\ref{eq:nav_vis_assist_pose_optim_pitch_jacobian}) Jacobian, as indicated by (\ref{eq:nav_vis_assist_optim_pitch_jacobian6}). 
\neweq{{\vec J}_{RRil}^{-1} = \Big[\vec O_{3x3} \ \ {\vec J}_{Ril}^{-1}\Big] \ \ \in \mathbb{R}^{3x6}}{eq:nav_vis_assist_optim_pitch_jacobian6}

The application of the iterative process described in \cite{Baker2004_part4} results in the following solution, which combines the contributions from the two different optimization targets:
\begin{eqnarray}
\nm{\vec H_{PO,il}} & = & \nm{\Big[\sum_j w_{TUK}\lrp{{\vec E_{RP,ilj}}^T \vec E_{RP,ilj}} {\vec J_{OF,ilj}}^T \vec J_{OF,ilj}\Big] + \ f_q^2 \cdot \Big[\vec J_{RRil}^{-T} \ \vec J_{RRil}^{-1}\Big] \ \ \in \mathbb{R}^{6x6}} \label{eq:nav_vis_assist_optim_joint_hes} \\
\nm{\Delta \circled{\vec \tau}_{{\sss EC}il}^{\sss C}} & = & \nm{- \vec H_{PO,il}^{-1} \bigg[\Big[\sum_j w_{TUK}\lrp{{\vec E_{RP,ilj}}^T \vec E_{RP,ilj}} {\vec J_{OF,ilj}}^T \vec E_{RP,ilj}\Big] + \ f_q \cdot \vec J_{RRil}^{-T} \ {\vec E}_{q,il}\bigg]} \label{eq:nav_vis_assist_pose_optim_joint_solution} \\
\nm{\circled{\vec \zeta}_{{\sss EC}i,l+1}} & \nm{\longleftarrow} & \nm{\circled{\vec \zeta}_{{\sss EC}il} \circ Exp\lrp{\Delta \circled{\vec \tau}_{{\sss EC}il}^{\sss C}} = \circled{\vec \zeta}_{{\sss EC}il} \oplus \Delta \circled{\vec \tau}_{{\sss EC}il}^{\sss C}} \label{eq:nav_vis_assist_pose_optim_joint_iterative}
\end{eqnarray}


\subsection{PI Control Inspired Pose Adjustment Activation}\label{subsec:IA-SVO_pi}

Sections \ref{subsec:IA-SVO_rationale} and \ref{subsec:IA-SVO_prior} describe the attitude adjustment and its fusion with the default reprojection error minimization pose optimization algorithm, but they do not specify the conditions under which the adjustment is activated, how the \nm{\circled{\vec q}_{{\sss NB}i}^{\circ\circ} \equiv \circled{\vec \phi}_{{\sss NB}i}^{\circ\circ}} target is determined, or the obtainment of its \nm{f_q} relative weight when applying the (\ref{eq:nav_vis_assist_pose_optim_joint_objective}) joint optimization. These parameters are determined below in three different cases: an adjustment in which only pitch is controlled, an adjustment in which both pitch and bank angles are controlled, and a complete attitude adjustment.
\begin{itemize}

\item \textbf{Pitch Adjustment Activation}. The attitude adjustment described in (\ref{eq:nav_vis_assist_pose_optim_pitch_objective}) through (\ref{eq:nav_vis_assist_pose_optim_pitch_jacobian}) can be converted into a pitch only adjustment by forcing the yaw and bank angle targets to coincide in each optimization step \emph{i} with the output of the reprojection only optimization:
\begin{eqnarray}
\nm{\circled{\psi}_i^{\circ\circ}} & = & \nm{\circled{\psi}_i^{\star\star}} \label{eq:nav_vis_assist_priors_activation_pitch_yaw_equal} \\
\nm{\circled{\theta}_i^{\circ\circ}} & = & \nm{\circled{\theta}_i^{\star\star} + \Delta \circled{\theta}_i^{\circ\circ}} \label{eq:nav_vis_assist_priors_activation_pitch_pitch} \\
\nm{\circled{\xi}_i^{\circ\circ}} & = & \nm{\circled{\xi}_i^{\star\star}} \label{eq:nav_vis_assist_priors_activation_pitch_bank_equal}
\end{eqnarray}

When activated as explained below, the new body pose target \nm{\circled{\vec \zeta}_{{\sss EB}i}^{\circ\circ}} only differs in one out of six dimensions (the pitch) from the reprojection only \nm{E_{RP}} optimum pose \nm{\circled{\vec \zeta}_{{\sss EB}i}^{\star\star}}, and the difference is very small as its effects are intended to accumulate over many successive images. This does not mean however that the other five components do not vary, as the joint optimization process described in (\ref{eq:nav_vis_assist_pose_optim_joint_objective}) through (\ref{eq:nav_vis_assist_pose_optim_joint_iterative}) freely optimizes within \nm{\mathbb{SE}\lrp{3}} with six degrees of freedom to minimize the joint cost function \nm{E_{PO}} that not only considers the reprojection error, but also the resulting pitch target.

The pitch adjustment aims for the visual estimations for altitude \nm{\circled{h}_i} and pitch \nm{\circled{\theta}_i} (in this order) not to deviate in excess from their inertially estimated counterparts \nm{\hat{h}_n} and \nm{\hat{\theta}_n}. It is inspired in a \emph{proportional integral} (\hypertt{PI}) control scheme \cite{Ogata2002, Skogestad2005, Stevens2003, Franklin1998} as the geometric altitude \emph{adjustment error}\footnote{In this context, \emph{adjustment error} is understood as the difference between the visual and inertial estimations.} \nm{\Delta h = \circled{h}_i - \hat{h}_n} can be considered as the integral of the pitch adjustment error \nm{\Delta \theta = \circled{\theta}_i - \hat{\theta}_n} in the sense that any difference between adjusted pitch angles (the \hypertt{P} control) slowly accumulate over time generating differences in adjusted altitude (the \hypertt{I} control). In addition, the adjustment also depends on the rate of climb (\hypertt{ROC}) adjustment error\footnote{To avoid noise, this is smoothed over the last 100 images or \nm{10 \ s}.} \nm{\Delta ROC = \circled{ROC}_i - \hat{ROC}_n}, which can be considered a second \hypertt{P} control as \hypertt{ROC} is the time derivative of the pressure altitude.
\begin{center}
\begin{tabular}{lccp{1.0cm}lcc}
	\hline
	Variable & Value & Unit & & Variable & Value & Unit \\
	\hline
	\nm{\Delta h_{LOW}}         & 25.0    & m             & & \nm{\Delta \circled{\theta}_{1,MAX}^{\circ\circ}}   & 0.0005  & \nm{^{\circ}} \\
	\nm{\Delta \theta_{LOW}}    & 0.2     & \nm{^{\circ}} & & \nm{\Delta \circled{\theta}_{2,MAX}^{\circ\circ}}   & 0.0003  & \nm{^{\circ}} \\
  \nm{\Delta ROC_{LOW}}       & 0.01    & m/s           & & \\
  \nm{\Delta \xi_{LOW}}  	    & 0.2     & \nm{^{\circ}} & & \nm{\Delta \circled{\xi}_{1,MAX}^{\circ\circ}} & 0.0003  & \nm{^{\circ}} \\
	\hline
\end{tabular}
\end{center}
\captionof{table}{Pitch and bank adjustment settings} \label{tab:nav_vis_assist_priors_activation_pitch}

Note that the objective is not for the visual estimations to closely track the inertial ones, but only to avoid excessive deviations, so there exist lower thresholds \nm{\Delta h_{LOW}}, \nm{\Delta \theta_{LOW}}, and \nm{\Delta ROC_{LOW}} below which the adjustments are not activated. These thresholds are arbitrary but have been set taking into account the \hypertt{INS} accuracy and its sources of error, as described in \cite{INSE}. If the absolute value of a certain adjustment error (difference between the visual and estimated estates) is above its threshold, the \hypertt{IA-VNS} can conclude with a high degree of confidence that the adjustment procedure can be applied; if below the threshold, the adjustment should not be employed as there is a significant risk that the true visual error (difference between the visual and actual state) may have the opposite sign, in which case the adjustment would be counterproductive. 

As an example, let's consider a case in which the visual altitude \nm{\circled{h}_i} is significantly higher than the observed one \nm{\hat{h}_n}, resulting in \nm{|\Delta h| > \Delta h_{LOW}}; in this case the \hypertt{IA-VNS} concludes that the aircraft is \say{high} and applies a negative pitch adjustment to slowly decrease the body pitch visual estimation \nm{\thetavis} over many images, with these accumulating over time into a lower altitude \nm{\hvis} that what would be the case if no adjustment were applied. On the other hand, if the absolute value of the adjustment error is below the threshold (\nm{|\Delta h| < \Delta h_{LOW}}), the adjustment should not be applied as there exists a significant risk that the aircraft is in fact \say{low} instead of \say{high} (when compared with the true altitude \nm{h_t}, not the the observed one \nm{\hat{h}_n}), and a negative pitch adjustment would only exacerbate the situation. A similar reasoning applies for the adjustment pitch error, in which the \hypertt{IA-VNS} reacts or not to correct perceived \say{nose-up} or \say{nose-down} visual estimations. The applied thresholds are displayed in table \ref{tab:nav_vis_assist_priors_activation_pitch}.

The \nm{\circled{\theta}_i^{\circ\circ}} pitch target to be applied for each image is given by (\ref{eq:nav_vis_assist_priors_activation_pitch_pitch}), where the obtainment of the pitch adjustment \nm{\Delta \circled{\theta}_i^{\circ\circ}} is explained below based on its three components (\ref{eq:nav_vis_assist_priors_activation_pitch_pitch_formula}):
\neweq{\Delta \circled{\theta}_i^{\circ\circ} = \Delta \circled{\theta}_h^{\circ\circ} + \Delta \circled{\theta}_\theta^{\circ\circ} + \Delta \circled{\theta}_{ROC}^{\circ\circ}} {eq:nav_vis_assist_priors_activation_pitch_pitch_formula}

\begin{itemize}
\item The pitch adjustment due to altitude, \nm{\Delta \circled{\theta}_h^{\circ\circ}}, linearly varies between zero when the adjustment error is below the threshold \nm{\Delta h_{LOW}} to \nm{\Delta \circled{\theta}_{1,MAX}^{\circ\circ}} when the error is twice the threshold, as shown in (\ref{eq:nav_vis_assist_priors_activation_pitch_Deltathetah}). The adjustment is bounded at this value to avoid destabilizing \hypertt{SVO} with pose adjustments that differ too much from their reprojection only optimum \nm{\circled{\vec \zeta}_{{\sss EB}i}^{\star\star}} (\ref{eq:nav_vis_assist_pose_optim_zetaEBstarstar}).
\end{itemize}
\neweq{\Delta \circled{\theta}_h^{\circ\circ} = \begin{dcases*}
\nm{0} & when \nm{|\Delta h| < \Delta h_{LOW}} \\
\nm{- \ sign\lrp{\Delta h} \, \Delta \circled{\theta}_{1,MAX}^{\circ\circ} \, \lrp{|\Delta h| - \Delta h_{LOW}} / \Delta h_{LOW}} & \nm{\Delta h_{LOW} \leq |\Delta h| \leq 2 \cdot \Delta h_{LOW}} \\
\nm{- \ sign\lrp{\Delta h} \, \Delta \circled{\theta}_{1,MAX}^{\circ\circ}} & when \nm{|\Delta h| > 2 \cdot \Delta h_{LOW}}  
\end{dcases*}} {eq:nav_vis_assist_priors_activation_pitch_Deltathetah}

\begin{itemize}
\item The pitch adjustment due to pitch, \nm{\Delta \circled{\theta}_\theta^{\circ\circ}}, works similarly but employing \nm{\Delta \theta} instead of \nm{\Delta h} and \nm{\Delta \theta_{LOW}} instead of \nm{\Delta h_{LOW}}, while also relying on the same limit \nm{\Delta \circled{\theta}_{1,MAX}^{\circ\circ}}. In addition, \nm{\Delta \circled{\theta}_\theta^{\circ\circ}} is set to zero if its sign differs from that of \nm{\Delta \circled{\theta}_h^{\circ\circ}}, and reduced so the combined effect of both targets does not exceed the limit (\nm{|\Delta \circled{\theta}_h^{\circ\circ} + \Delta \circled{\theta}_\theta^{\circ\circ}| \leq \Delta \circled{\theta}_{1,MAX}^{\circ\circ}}).

\item The pitch adjustment due to rate of climb, \nm{\Delta \circled{\theta}_{ROC}^{\circ\circ}}, also follows a similar scheme but employing \nm{\Delta ROC} instead of \nm{\Delta h}, \nm{\Delta ROC_{LOW}} instead of \nm{\Delta h_{LOW}}, and \nm{\Delta \circled{\theta}_{2,MAX}^{\circ\circ}} instead of \nm{\Delta \circled{\theta}_{1,MAX}^{\circ\circ}}. Additionally, it is multiplied by the ratio between \nm{\Delta \circled{\theta}_h^{\circ\circ}} and \nm{\Delta \circled{\theta}_{1,MAX}^{\circ\circ}} to limit its effects when the altitude estimated error \nm{\Delta h} is small. This adjustment can act in both directions, imposing bigger pitch adjustments if the altitude error is increasing or lower one if it is already diminishing.
\end{itemize}

If activated, the weight value \nm{f_q} required for the (\ref{eq:nav_vis_assist_pose_optim_joint_objective}) joint optimization is determined by imposing that the weighted attitude error \nm{f_q \cdot E_{q}\big(\circled{\vec r}_{{\sss EC}i0}\big)} coincides with the reprojection error \nm{E_{RP}(\circled{\vec \zeta}_{{\sss EC}i0})} when evaluated before the first iteration, this is, it assigns the same weight to the two active components of the joint \nm{E_{PO}(\circled{\vec \zeta}_{{\sss EC}i})} cost function (\ref{eq:nav_vis_assist_pose_optim_joint_objective}).

\item \textbf{Pitch and Bank Adjustment Activation}. The previous scheme can be modified to also make use of the observed body bank angle \nm{\hat{\xi}_n} within the framework established by the (\ref{eq:nav_vis_assist_pose_optim_pitch_objective}) through (\ref{eq:nav_vis_assist_pose_optim_pitch_jacobian}) attitude adjustment optimization:
\begin{eqnarray}
\nm{\circled{\psi}_i^{\circ\circ}} & = & \nm{\circled{\psi}_i^{\star\star}} \label{eq:nav_vis_assist_priors_activation_pitchbank_yaw_equal} \\
\nm{\circled{\theta}_i^{\circ\circ}} & = & \nm{\circled{\theta}_i^{\star\star} + \Delta \circled{\theta}_i^{\circ\circ}} \label{eq:nav_vis_assist_priors_activation_pitchbank_pitch} \\
\nm{\circled{\xi}_i^{\circ\circ}} & = & \nm{\circled{\xi}_i^{\star\star} + \Delta \circled{\xi}_i^{\circ\circ}} \label{eq:nav_vis_assist_priors_activation_pitch_bankbank_bank}
\end{eqnarray}

Although the new body pose target \nm{\circled{\vec \zeta}_{{\sss EB}i}^{\circ\circ}} only differs in two out of six dimensions (pitch and bank) from the optimum pose \nm{\circled{\vec \zeta}_{{\sss EB}i}^{\star\star}} obtained by minimizing the reprojection error exclusively, all six degrees of freedom are allowed to vary when minimizing the joint cost function.

The determination of the pitch adjustment \nm{\Delta \circled{\theta}_i^{\circ\circ}} does not vary, and that of the bank adjustment \nm{\Delta \circled{\xi}_i^{\circ\circ}} relies on a linear adjustment between two values similar to any of the three components of (\ref{eq:nav_vis_assist_priors_activation_pitch_pitch_formula}), but relying on the bank angle adjustment error \nm{\Delta \xi = \circled{\xi}_i - \hat{\xi}_n}, as well as a \nm{\Delta \xi_{LOW}} threshold and \nm{\Delta \circled{\xi}_{1,MAX}^{\circ\circ}} maximum adjustment whose values are provided in table \ref{tab:nav_vis_assist_priors_activation_pitch}. Note that the value of the \nm{\Delta \xi_{LOW}} threshold coincides with that of \nm{\Delta \theta_{LOW}} as the \hypertt{INS} accuracy for both pitch and roll is similar according to \cite{INSE}. 

It is important to remark that the combined pitch and bank adjustment activation is the one employed to generate the \hypertt{IA-VNS} results in sections \ref{sec:Results} and \ref{sec:Influence_terrain}.

\item \textbf{Attitude Adjustment Activation}. The use of the observed yaw angle \nm{\psiest} is not recommended as the visual estimation \nm{\psivis} provided by the \hypertt{VNS} (without any inertial inputs) is in general more accurate than its inertial counterpart \nm{\psiest}, as shown in section \ref{sec:Results} (table \ref{tab:Res_vis_assist_results_euler}). This can be traced on one side to the bigger influence that a yaw change has on the resulting optical flow when compared with those caused by pitch and bank changes, which makes the body yaw angle easier to track by the \hypertt{VNS}, and on the other to the \hypertt{INS} relying on the gravity pointing down to control pitch and bank adjustments versus the less robust dependence on the Earth magnetic field and associated magnetometer readings used to estimate the aircraft heading \cite{INSE}.

For this reason, the attitude adjustment process described next has not been implemented, although it is included here as a suggestion for other applications in which the objective may be to adjust the vehicle attitude as a whole. The process relies on the observed attitude \nm{\hat{\vec q}_{{\sss NB}n}} provided by the \hypertt{INS} and the initial estimation \nm{\circled{\vec q}_{{\sss NB}i}^{\star\star}} provided by the reprojection only pose optimization process. Its difference is given by \nm{\Delta \circled{\vec r}^{{\sss B}i,\star\star} = \hat{\vec q}_{{\sss NB}n} \ominus \circled{\vec q}_{{\sss NB}i}^{\star\star}}, where the superindex \say{\texttt{Bi}} indicates that it is viewed in the pose optimized body frame. This perturbation can be decoupled into  a rotating direction and an angular displacement \cite{LIE,Sola2018}, resulting in \nm{\Delta \circled{\vec r}^{{\sss B}i,\star\star} = \circled{\vec n}^{{\sss B}i,\star\star} \ \Delta \circled{\phi}^{\star\star}}.

Let's now consider that the \hypertt{IA-VNS} decides to set an attitude target that differs by \nm{\Delta \circled{\phi}^{\circ\circ}} from its reprojection only solution \nm{\circled{\vec q}_{{\sss NB}i}^{\star\star}}, but rotating about the axis that leads towards its inertial estimation \nm{\hat{\vec q}_{{\sss NB}n}}. The target attitude \nm{\circled{\vec q}_{{\sss NB}i}^{\circ\circ}} can then be obtained by \nm{\mathbb{SO}(3)} Spherical Linear Interpolation (\hypertt{SLERP}) \cite{LIE,Sola2017}, where \nm{t = \Delta \circled{\phi}^{\circ\circ} \ / \ \Delta \circled{\phi}^{\star\star}} is the ratio between the target rotation and the attitude error or estimated angular displacement:
\neweq{\circled{\vec q}_{{\sss NB}i}^{\circ\circ} = \circled{\vec q}_{{\sss NB}i}^{\star\star} \otimes \lrp{{\circled{\vec q}_{{\sss NB}i}^{\star\star}}^\ast \otimes \hat{\vec q}_{{\sss NB}n}}^t}{eq:nav_vis_assist_priors_activation_pitch_atttitude}
\end{itemize}


\subsection{Additional Modifications to SVO}\label{subsec:IA-SVO_extra}

In addition to the \hypertt{PI} inspired introduction of priors into the pose optimization phase, the proposed \hypertt{IA-VNS} also includes various other modifications to the original \hypertt{SVO} pipeline described in section \ref{sec:SVO}. These include the addition of the current features to the structure optimization phase (so the pose adjustments introduced by the prior based pose optimization are not reverted), the replacement of the sparse image alignment phase by an inertial estimation of the \nm{\circled{\vec \zeta}_{{\sss EC}i0} = \circled{\vec \zeta}_{{\sss EC}i}^{\star}} input to the pose optimization process, and the use of the \hypertt{GNSS}-Based inertial distance estimations to obtain more accurate height and path angle values for the \hypertt{SVO} initialization.


\section{Testing: High Fidelity Simulation and Scenarios}\label{sec:Simulation}

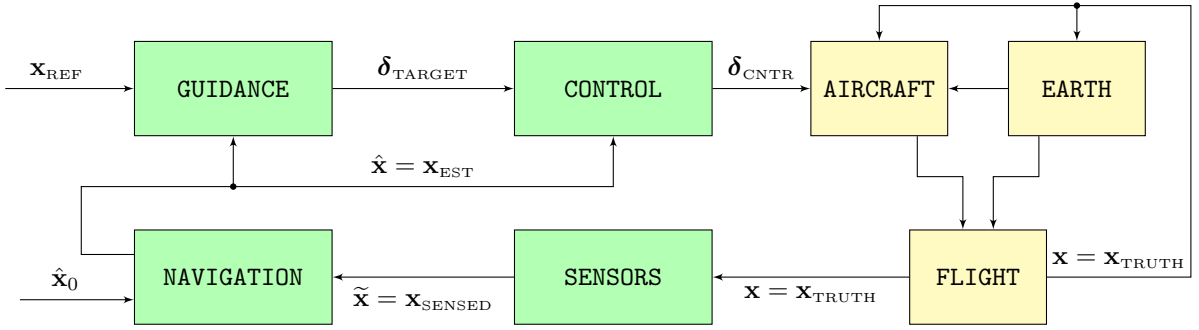
\begin{figure}[h]
\centering
\begin{tikzpicture}[auto, node distance=2cm,>=latex']

	\node [coordinate](xrefinput) {};
	\node [blockgreen, right of=xrefinput, minimum width=2.6cm, node distance=3.0cm, align=center, minimum height=1.25cm] (GUIDANCE) {\texttt{GUIDANCE}};
	\draw [->] (xrefinput) -- node[pos=0.4] {\nm{\xREF}} (GUIDANCE.west);

	\node [blockgreen, right of=GUIDANCE, minimum width=2.6cm, node distance=5.0cm, align=center, minimum height=1.25cm] (CONTROL) {\texttt{CONTROL}};
	\draw [->] (GUIDANCE.east) -- node[pos=0.5] {\nm{\deltaTARGET}} (CONTROL.west);

	\node [blockyellow, right of=CONTROL, minimum width=1.8cm, node distance=3.5cm, align=center, minimum height=1.25cm] (AIRCRAFT) {\texttt{AIRCRAFT}};
	\draw [->] (CONTROL.east) --  node[pos=0.5] {\nm{\deltaCNTR}} (AIRCRAFT.west);

	\node [blockyellow, right of=AIRCRAFT, minimum width=1.8cm, node distance=2.6cm, align=center, minimum height=1.25cm] (EARTH) {\texttt{EARTH}};
	\draw [->] (EARTH.west) -- (AIRCRAFT.east);

	\node [coordinate, right of=AIRCRAFT, node distance=1.3cm] (midpoint){};
	\node [blockyellow, below of=midpoint, minimum width=1.8cm, node distance=2.5cm, align=center, minimum height=1.25cm] (FLIGHT) {\texttt{FLIGHT}};
	\draw [->] ($(EARTH.south)-(0.50cm,0cm)$) |- ($(FLIGHT.north)+(0.20cm,0.70cm)$) -- ($(FLIGHT.north)+(0.20cm,0cm)$);
	\draw [->] ($(AIRCRAFT.south)+(0.50cm,0cm)$) |- ($(FLIGHT.north)+(-0.20cm,0.70cm)$) -- ($(FLIGHT.north)-(0.20cm,0cm)$);

	\node [blockgreen, below of=GUIDANCE, minimum width=2.6cm, node distance=2.5cm, align=center, minimum height=1.25cm] (NAVIGATION) {\texttt{NAVIGATION}};
	\node [coordinate, left of=NAVIGATION, node distance=2.0cm] (pointnav1){};
	\node [coordinate, above of=pointnav1, node distance=1.2cm] (pointnav2){};
	\node [coordinate, below of=GUIDANCE, node distance=1.3cm] (pointnav3){};
	\filldraw [black] (pointnav3) circle [radius=1pt];
	\draw [->] ($(NAVIGATION.west)+(0cm,0.3cm)$) -| (pointnav2) -- (pointnav3) -- (GUIDANCE.south);
	\draw [->] (pointnav3) -| node[pos=0.25] {\nm{\xvecest = \xEST}} (CONTROL.south);
	\draw [->] ($(NAVIGATION.west)+(-1.5cm,-0.3cm)$) -- node[pos=0.4] {\nm{\xvecestzero}} ($(NAVIGATION.west)+(0cm,-0.3cm)$);

	\node [blockgreen, below of=CONTROL, minimum width=2.6cm, node distance=2.5cm, align=center, minimum height=1.25cm] (SENSORS) {\texttt{SENSORS}};
	\draw [->] (SENSORS.west) -- node[pos=0.5] {\nm{\xvectilde = \xSENSED}} (NAVIGATION.east);

	\node [coordinate, right of=EARTH, node distance=1.5cm] (pointflight1){};
	\node [coordinate, above of=EARTH, node distance=1.1cm] (pointflight2){};
	\filldraw [black] (pointflight2) circle [radius=1pt];
	\draw [->] (FLIGHT.west) -- node[pos=0.5] {\nm{\xvec = \xTRUTH}} (SENSORS.east);
	\draw [->] (FLIGHT.east) -| node[pos=0.25] {\nm{\xvec = \xTRUTH}} (pointflight1) |- (pointflight2) -| (AIRCRAFT.north);
	\draw [->] (pointflight2) -- (EARTH.north);
\end{tikzpicture}
\caption{Components of the high fidelity simulation}
\label{fig:flow_diagram}
\end{figure}

To evaluate the performance of the proposed visual navigation algorithms, this article relies on Monte Carlo simulations consisting of one hundred runs each of two different scenarios based on the high fidelity stochastic flight simulator graphically depicted in figure \ref{fig:flow_diagram}. Described in detail in \cite{SIMULATION} and with its open source \nm{\CC} implementation available in \cite{Gallo2020_simulation}, the simulator models the flight in varying weather and turbulent conditions of a low \hypertt{SWaP} fixed wing piston engine autonomous \hypertt{UAV}.

The simulator consists on two distinct processes. The first, represented by the yellow blocks on the right of figure \ref{fig:flow_diagram}, models the physics of flight and the interaction between the aircraft and its surroundings that results in the real aircraft trajectory \nm{\xvec = \xTRUTH}; the second, represented by the green blocks on the left, contains the aircraft systems in charge of ensuring that the resulting trajectory adheres as much as possible to the mission objectives. It includes the different sensors whose output comprise the sensed trajectory \nm{\xvectilde = \xSENSED}, the navigation system in charge of filtering it to obtain the estimated trajectory \nm{\xvecest = \xEST}, the guidance system that converts the reference objectives \nm{\xREF} into the control targets \nm{\deltaTARGET}, and the control system that adjusts the position of the throttle and aerodynamic control surfaces \nm{\deltaCNTR} so the estimated trajectory \nm{\xvecest} is as close as possible to the reference objectives \nm{\xREF}. Table \ref{tab:frequencies} provides the working frequencies employed for the different trajectories shown in figures \ref{fig:flow_diagram_ins}, \ref{fig:flow_diagram_vns}, \ref{fig:flow_diagram_iavns}, and \ref{fig:flow_diagram}.
\begin{center}
\begin{tabular}{lrrcl}
	\hline
	\multicolumn{1}{c}{Discrete Time} & Frequency & Period & Variables & Systems \\
	\hline
	\nm{t_t = t \cdot \DeltatTRUTH}            & \nm{500 \ Hz} & \nm{0.002 \ s} & \nm{\xvec = \xTRUTH}             & Flight physics \\
	\nm{t_s = s \cdot \DeltatSENSED}           & \nm{100 \ Hz} & \nm{0.01 \ s}  & \nm{\xvectilde = \xSENSED}       & Sensors \\ 
	\nm{t_n = n \cdot \DeltatEST}              & \nm{100 \ Hz} & \nm{0.01 \ s}  & \nm{\xvecest = \xEST}            & Inertial navigation \\ 
	\nm{t_c = c \cdot \DeltatCNTR}             & \nm{ 50 \ Hz} & \nm{0.02 \ s}  & \nm{\deltaTARGET, \, \deltaCNTR} & Guidance \& control \\
	\nm{t_i = i \cdot \DeltatIMG}              & \nm{ 10 \ Hz} & \nm{0.1 \ s}   & \nm{\xvecvis = \xIMG}            & Visual navigation \& camera \\
	\hline
\end{tabular}
\end{center}
\captionof{table}{Working frequencies of the different systems and trajectory representations} \label{tab:frequencies}

All components of the flight simulator have been modeled with as few simplifications as possible to increase the realism of the results, as explained in \cite{SIMULATION,SENSORS}. With the exception of the aircraft performances and its control system, which are deterministic, all other simulator components are treated as stochastic and hence vary from one execution to the next, enhancing the significance of the Monte Carlo simulation results. 


\subsection{Camera}\label{subsec:Simulation_camera}

The flight simulator has the capability, when provided with the camera pose\footnote{The camera is positioned facing down and rigidly attached to the aircraft structure.} with respect to the Earth at equally time spaced intervals, of generating images that resemble the view of the Earth surface that the camera would record if located at that particular pose. To do so, it relies on the \texttt{Earth Viewer} library, a modification to \texttt{osgEarth} \cite{osgEarth} (which in turn relies on \texttt{OpenSceneGraph} \cite{OpenSceneGraph}) capable of generating realistic Earth images as long as the camera height over the terrain is significantly higher than the vertical relief present in the image. A more detailed explanation of the image generation process is provided in \cite{SENSORS}.

It is assumed that the shutter speed is sufficiently high that all images are equally sharp, and that the image generation process is instantaneous. In addition, the camera \hypertt{ISO} setting remains constant during the flight, and all generated images are noise free. The simulation also assumes that the visible spectrum radiation reaching all patches of the Earth surface remains constant, and the terrain is considered Lambertian \cite{Soatto2001}, so its appearance at any given time does not vary with the viewing direction. The combined use of these assumptions implies that a given terrain object is represented with the same luminosity in all images, even as its relative pose (position and attitude) with respect to the camera varies. Geometrically, the simulation adopts a perspective projection or pinhole camera model \cite{Soatto2001}, which in addition is perfectly calibrated and hence shows no distortion. The camera has a focal length of \nm{19 \, mm} and a sensor with 768 by 1024 pixels.


\subsection{Scenarios}\label{subsec:Simulation_scenarios}

Most \hypertt{VIO} packages discussed in section \ref{sec:GNSS-Denied} include in their release articles an evaluation when applied to the \texttt{EuRoC} Micro Air Vehicle (\hypertt{MAV}) datasets \cite{Burri2016}, and so do independent articles such as \cite{Delmerico2018}. These datasets contain perfectly synchronized stereo images, \hypertt{IMU} measurements, and ground truth readings obtained with laser, for eleven different indoor trajectories flown with a \hypertt{MAV}, each with a duration in the order of two minutes and a total distance in the order of \nm{100 \, m}. This fact by itself indicates that the target application of exiting \hypertt{VIO} implementations differs significantly from the main focus of this article, which is the long term flight of a fixed wing \hypertt{UAV} in \hypertt{GNSS}-Denied conditions, as there may exist accumulating errors that are completely non discernible after such short periods of time, but that grow non linearly and have the capability of inducing significant pose errors when the aircraft remains aloft for long periods of time.

The algorithms introduced in this article are hence tested through simulation under two different scenarios designed to analyze the consequences of losing the \hypertt{GNSS} signals for long periods of time. Although a short summary is included below, detailed descriptions of the mission, weather, and wind field employed in each scenario can be found in \cite{SIMULATION}. Most parameters comprising the scenario are defined stochastically, resulting in different values for every execution. Note that all results shown in sections \ref{sec:Results} and \ref{sec:Influence_terrain} are based on Monte Carlo simulations comprising one hundred runs of each scenario, testing the sensitivity of the proposed navigation algorithms to a wide variety of values in the parameters.
\begin{itemize}
\item Scenario \#1 has been defined with the objective of adequately representing the challenges faced by an autonomous fixed wing \hypertt{UAV} that suddenly cannot rely on \hypertt{GNSS} and hence changes course to reach a predefined recovery location situated at approximately one hour of flight time. In the process, in addition to executing an altitude and airspeed adjustment, the autonomous aircraft faces significant weather and wind field changes that make its \hypertt{GNSS}-Denied navigation even more challenging. 

With respect to the mission, the stochastic parameters include the initial airspeed, pressure altitude, and bearing (\nm{\vtasINI, \HpINI, \chiINI}), their final values (\nm{\vtasEND, \HpEND, \chiEND}), and the time at which each of the three maneuvers is initiated\footnote{Turns are executed with a bank angle of \nm{\xiTURN = \pm 10 \, ^{\circ}}, altitude changes employ an aerodynamic path angle of \nm{\gammaTASCLIMB = \pm 2 \, ^{\circ}}, and airspeed modifications are automatically executed by the control system as set-point changes.}. The scenario lasts for \nm{\tEND = 3800 \, s}, while the \hypertt{GNSS} signals are lost at \nm{\tGNSS = 100 \, s}.

The wind field is also defined stochastically, as its two parameters (speed and bearing) are constant both at the beginning (\nm{\vwindINI, \chiWINDINI}) and conclusion (\nm{\vwindEND, \chiWINDEND}) of the scenario, with a linear transition in between. The specific times at which the wind change starts and concludes also vary stochastically among the different simulation runs. As described in \cite{SIMULATION}, the turbulence remains strong throughout the whole scenario, but its specific values also vary stochastically from one execution to the next.

A similar linear transition occurs with the temperature and pressure offsets that define the atmospheric properties \cite{INSA}, as they are constant both at the start (\nm{\DeltaTINI, \DeltapINI}) and end (\nm{\DeltaTEND, \DeltapEND}) of the flight. In contrast with the wind field, the specific times at which the two transitions start and conclude are not only stochastic but also different from each other.

\item Scenario \#2 represents the challenges involved in continuing with the original mission upon the loss of the \hypertt{GNSS} signals, executing a series of continuous turn maneuvers over a relatively short period of time with no atmospheric or wind variations. As in scenario \nm{\#1}, the \hypertt{GNSS} signals are lost at \nm{\tGNSS = 100 \, s}, but the scenario duration is shorter (\nm{\tEND = 500 \, s}). The initial airspeed and pressure altitude (\nm{\vtasINI, \HpINI}) are defined stochastically and do not change throughout the whole scenario; the bearing however changes a total of eight times between its initial and final values, with all intermediate bearing values as well as the time for each turn varying stochastically from one execution to the next. Although the same turbulence is employed as in scenario \nm{\#1}, the wind and atmospheric parameters (\nm{\vwindINI, \chiWINDINI, \DeltaTINI, \DeltapINI}) remain constant throughout scenario \nm{\#2}.
\end{itemize}


\section{Visual Navigation System Error in GNSS-Denied Conditions}\label{sec:Results}

This section presents the results obtained with the proposed \hypertt{IA-VNS} when executing Monte Carlo simulations of the two \hypertt{GNSS}-Denied scenarios over the \hypertt{MX} terrain type\footnote{Section \ref{sec:Influence_terrain} defines the various terrain types than can be overflown by the aircraft, and then analyzes their influence on the simulation results.}, each consisting of one hundred executions. They are compared with the results obtained with the stand alone \hypertt{VNS} that relies on the original \hypertt{SVO} algorithm, and with those of the \hypertt{INS} described in \cite{INSE}. The tables below contain the \emph{navigation system error}\footnote{The \hypertt{NSE} is the difference between the real states (\nm{\xvec}) and their estimation (\nm{\xvecest} or \nm{\xvecvis}) by the navigation system.} (\hypertt{NSE}) incurred by the various navigation systems (and accordingly denoted as \hypertt{INSE}, \hypertt{VNSE}, and \hypertt{IA-VNSE}) at the conclusion of the two \hypertt{GNSS}-Denied scenarios, represented by the mean, standard deviation, and maximum value of the estimation errors. In addition, the figures depict the variation with time of the \hypertt{NSE} mean and standard deviation for the one hundred executions. The following remarks are necessary:
\begin{itemize}
\item The results obtained with the \hypertt{INS} under the same two \hypertt{GNSS}-Denied scenarios are described in detail in \cite{INSE}. The attitude \hypertt{INSE} does not drift and is bounded by the quality of the onboard sensors, ensuring the aircraft can remain aloft for as long as there is fuel available. The vertical position and ground velocity \hypertt{INSE}s are also bounded by atmospheric physics and do not drift; their estimation errors depend on the atmospheric pressure offset and wind field changes that occur since the \hypertt{GNSS} signals are lost. On the other hand, the horizontal position \hypertt{INSE} drifts as a consequence of integrating the ground velocity without absolute observations. Of the six \nm{\mathbb{SE}(3)} degrees of freedom (three for attitude, two for horizontal position, one for altitude), the \hypertt{INS} is hence capable of successfully estimating four of them in \hypertt{GNSS}-Denied conditions.

\item Visual navigation systems (either \hypertt{VNS} or \hypertt{IA-VNS}) are only necessary to reduce the estimation error in the two remaining degrees of freedom (the horizontal position). Although both of them estimate the complete six dimensional aircraft pose, their attitude and altitude estimations shall only be understood as a means to provide an accurate horizontal position estimation, which represents their sole objective.

\item As described below, the \hypertt{SVO} based \hypertt{VNS} drifts in all six degrees of freedom. The main focus of this article is on how the addition of \hypertt{INS} based priors enables the \hypertt{IA-VNS} to reduce the drift in all six dimensions, with the resulting horizontal position \hypertt{IA-VNSE} being just a fraction of the \hypertt{INSE}. The attitude and altitude \hypertt{IA-VNSE}s, although improved when compared to the \hypertt{VNSE}s, are qualitatively inferior to the driftless \hypertt{INSE}s, but note that their purpose is just to enable better horizontal position \hypertt{IA-VNS} estimations, not to replace the attitude and altitude \hypertt{INS} outputs.  
\end{itemize}


\subsection{Body Attitude Estimation}\label{subsec:Results_att}

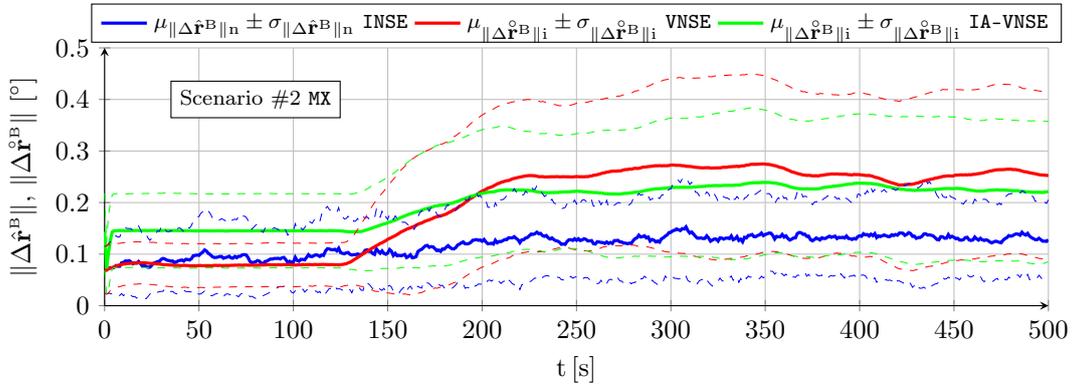
\begin{figure}[h]
\centering
\pgfplotsset{
	every axis legend/.append style={
		at={(0.50,1.02)},
		anchor=south,
	},
}
\begin{tikzpicture}
\begin{axis}[
cycle list={{blue,no markers,very thick},
            {red,no markers,very thick},
            {green,no markers,very thick},
            {blue,dashed,no markers,ultra thin},{blue,dashed,no markers,ultra thin},
            {red,dashed,no markers,ultra thin},{red,dashed,no markers,ultra thin},
            {green,dashed,no markers,ultra thin},{green,dashed,no markers,ultra thin}},
width=14.0cm,
height=5.0cm,
xmin=0, xmax=500, xtick={0,50,...,500},
xlabel={\nm{t \lrsb{s}}},
xmajorgrids,
ymin=0, ymax=0.5, ytick={0,0.1,0.2,0.3,0.4,0.5},
ylabel={\nm{\DeltarBestnorm, \, \DeltarBvisnorm \, \lrsb{^{\circ}}}},
ymajorgrids,
axis lines=left,
axis line style={-stealth},
legend entries={\nm{\mun{\DeltarBestnorm} \pm \sigman{\DeltarBestnorm}} \hypertt{INSE},
                \nm{\mui{\DeltarBvisnorm} \pm \sigmai{\DeltarBvisnorm}} \hypertt{VNSE},
				\nm{\mui{\DeltarBvisnorm} \pm \sigmai{\DeltarBvisnorm}} \hypertt{IA-VNSE}},
legend columns=3,
legend style={font=\footnotesize},
legend cell align=left,
]
\pgfplotstableread{figs/ch18_res_vis_assist/error_vis_att/error_vis_alter_euler_deg_nav.txt}\mytablenav
\pgfplotstableread{figs/ch18_res_vis_assist/error_vis_att/error_vis_alter_euler_deg_vis_relocalization.txt}\mytablevispartial
\pgfplotstableread{figs/ch18_res_vis_assist/error_vis_att/error_vis_alter_euler_deg_vis.txt}\mytablevis
\addplot table [header=false, x index=0,y index=1] {\mytablenav};
\addplot table [header=false, x index=0,y index=1] {\mytablevispartial};
\addplot table [header=false, x index=0,y index=1] {\mytablevis};
\addplot table [header=false, x index=0,y index=2] {\mytablenav};
\addplot table [header=false, x index=0,y index=3] {\mytablenav};
\addplot table [header=false, x index=0,y index=2] {\mytablevispartial};
\addplot table [header=false, x index=0,y index=3] {\mytablevispartial};
\addplot table [header=false, x index=0,y index=2] {\mytablevis};
\addplot table [header=false, x index=0,y index=3] {\mytablevis};
\path node [draw, shape=rectangle, fill=white] at (80,0.40) {\footnotesize Scenario \nm{\#2} \hypertt{MX}};
\end{axis}   
\end{tikzpicture}
\caption{Body attitude \texttt{INSE}, \texttt{VNSE}, and \texttt{IA-VNSE} for scenario \nm{\#2} \texttt{MX} (100 runs)}
\label{fig:Res_vis_assist_results_euler_alter}
\end{figure}

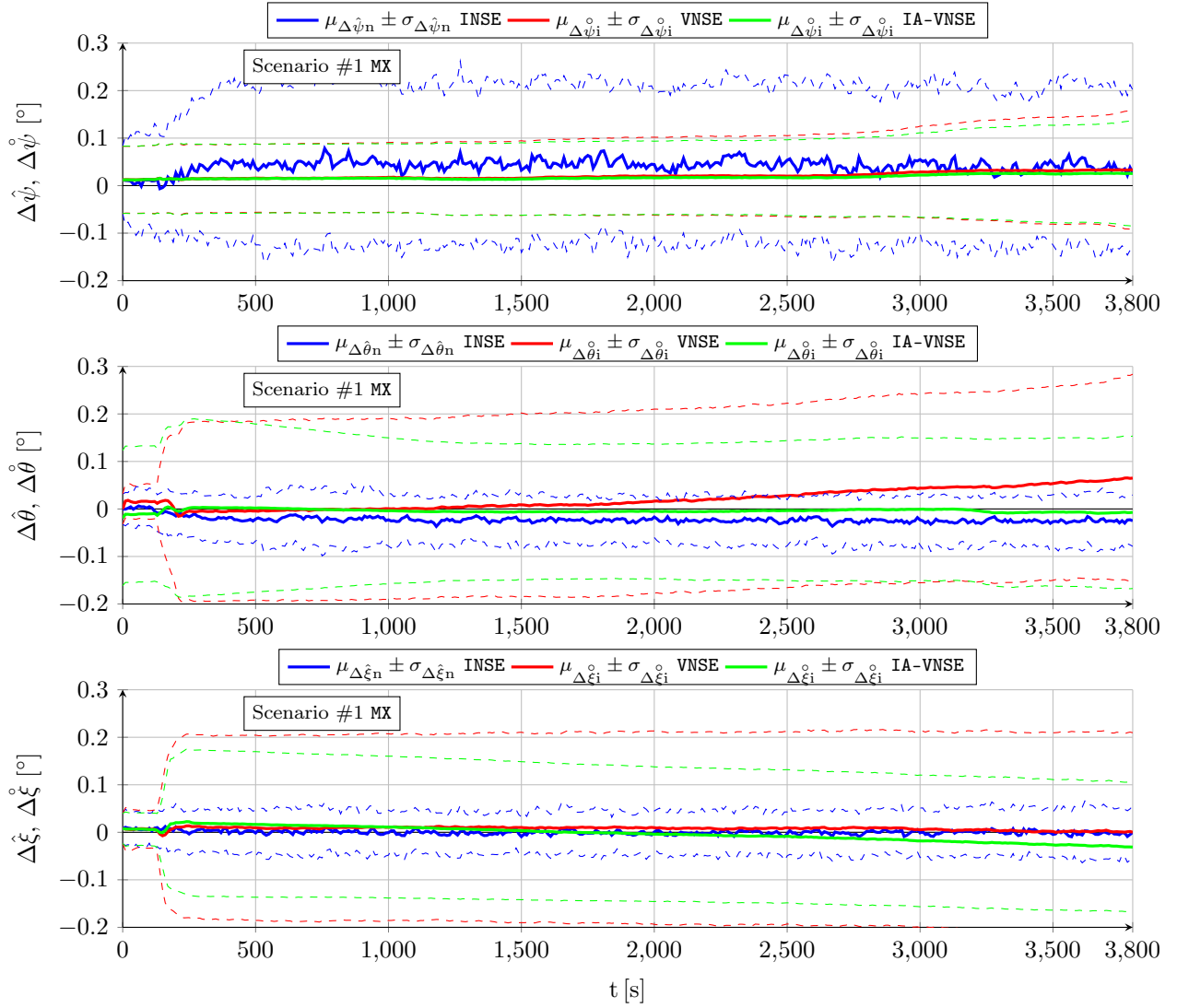
\begin{figure}[h]
\centering
\pgfplotsset{
	every axis legend/.append style={
		at={(0.50,1.02)},
		anchor=south,
	},
}
\begin{tikzpicture}
\begin{axis}[
cycle list={{blue,no markers,very thick},
            {red,no markers,very thick},
            {green,no markers,very thick},
            {blue,dashed,no markers},{blue,dashed,no markers},
            {red,dashed,no markers},{red,dashed,no markers},
            {green,dashed,no markers},{green,dashed,no markers}},
width=16.0cm,
height=5.0cm,
xmin=0, xmax=3800, xtick={0,500,...,3500,3800}, 
xmajorgrids,
ymin=-0.2, ymax=0.3, ytick={-0.3,-0.2,-0.1,0,0.1,0.2,0.3},
ylabel={\nm{\Delta\psiest, \, \Delta\psivis \, \lrsb{^{\circ}}}},
ymajorgrids,
axis lines=left,
axis line style={-stealth},
legend entries={\nm{\mun{\Delta\psiest} \pm \sigman{\Delta\psiest}} \hypertt{INSE},
                \nm{\mui{\Delta\psivis} \pm \sigmai{\Delta\psivis}} \hypertt{VNSE},
				\nm{\mui{\Delta\psivis} \pm \sigmai{\Delta\psivis}} \hypertt{IA-VNSE}},
legend columns=3,
legend style={font=\footnotesize},
legend cell align=left,
]
\pgfplotstableread{figs/ch18_res_vis_assist/error_vis_att/error_vis_base_psi_deg_nav.txt}\mytablenav
\pgfplotstableread{figs/ch18_res_vis_assist/error_vis_att/error_vis_base_psi_deg_vis_relocalization.txt}\mytablevispartial
\pgfplotstableread{figs/ch18_res_vis_assist/error_vis_att/error_vis_base_psi_deg_vis.txt}\mytablevis
\draw [] (0.0,0.0) -- (3800.0,0.0);
\addplot table [header=false, x index=0,y index=1] {\mytablenav};
\addplot table [header=false, x index=0,y index=1] {\mytablevispartial};
\addplot table [header=false, x index=0,y index=1] {\mytablevis};
\addplot table [header=false, x index=0,y index=2] {\mytablenav};
\addplot table [header=false, x index=0,y index=3] {\mytablenav};
\addplot table [header=false, x index=0,y index=2] {\mytablevispartial};
\addplot table [header=false, x index=0,y index=3] {\mytablevispartial};
\addplot table [header=false, x index=0,y index=2] {\mytablevis};
\addplot table [header=false, x index=0,y index=3] {\mytablevis};
\path node [draw, shape=rectangle, fill=white] at (750,0.25) {\footnotesize Scenario \nm{\#1} \hypertt{MX}};
\end{axis}   
\end{tikzpicture}
\pgfplotsset{
	every axis legend/.append style={
		at={(0.50,1.02)},
		anchor=south,
	},
}
\begin{tikzpicture}
\begin{axis}[
cycle list={{blue,no markers,very thick},
            {red,no markers,very thick},
            {green,no markers,very thick},
            {blue,dashed,no markers},{blue,dashed,no markers},
            {red,dashed,no markers},{red,dashed,no markers},
            {green,dashed,no markers},{green,dashed,no markers}},
width=16.0cm,
height=5.0cm,
xmin=0, xmax=3800, xtick={0,500,...,3500,3800},
xmajorgrids,
ymin=-0.2, ymax=0.3, ytick={-0.2,-0.1,0,0.1,0.2,0.3},
ylabel={\nm{\Delta\thetaest, \, \Delta\thetavis \, \lrsb{^{\circ}}}},
ymajorgrids,
axis lines=left,
axis line style={-stealth},
legend entries={\nm{\mun{\Delta\thetaest} \pm \sigman{\Delta\thetaest}} \hypertt{INSE},
                \nm{\mui{\Delta\thetavis} \pm \sigmai{\Delta\thetavis}} \hypertt{VNSE},
				\nm{\mui{\Delta\thetavis} \pm \sigmai{\Delta\thetavis}} \hypertt{IA-VNSE}},
legend columns=3,
legend style={font=\footnotesize},
legend cell align=left,
]
\pgfplotstableread{figs/ch18_res_vis_assist/error_vis_att/error_vis_base_theta_deg_nav.txt}\mytablenav
\pgfplotstableread{figs/ch18_res_vis_assist/error_vis_att/error_vis_base_theta_deg_vis_relocalization.txt}\mytablevispartial
\pgfplotstableread{figs/ch18_res_vis_assist/error_vis_att/error_vis_base_theta_deg_vis.txt}\mytablevis
\draw [] (0.0,0.0) -- (3800.0,0.0);
\addplot table [header=false, x index=0,y index=1] {\mytablenav};
\addplot table [header=false, x index=0,y index=1] {\mytablevispartial};
\addplot table [header=false, x index=0,y index=1] {\mytablevis};
\addplot table [header=false, x index=0,y index=2] {\mytablenav};
\addplot table [header=false, x index=0,y index=3] {\mytablenav};
\addplot table [header=false, x index=0,y index=2] {\mytablevispartial};
\addplot table [header=false, x index=0,y index=3] {\mytablevispartial};
\addplot table [header=false, x index=0,y index=2] {\mytablevis};
\addplot table [header=false, x index=0,y index=3] {\mytablevis};
\path node [draw, shape=rectangle, fill=white] at (750,0.25) {\footnotesize Scenario \nm{\#1} \hypertt{MX}};
\end{axis}   
\end{tikzpicture}
\pgfplotsset{
	every axis legend/.append style={
		at={(0.50,1.02)},
		anchor=south,
	},
}
\begin{tikzpicture}
\begin{axis}[
cycle list={{blue,no markers,very thick},
            {red,no markers,very thick},
            {green,no markers,very thick},
            {blue,dashed,no markers},{blue,dashed,no markers},
            {red,dashed,no markers},{red,dashed,no markers},
            {green,dashed,no markers},{green,dashed,no markers}},
width=16.0cm,
height=5.0cm,
xmin=0, xmax=3800, xtick={0,500,...,3500,3800},
xlabel={\nm{t \lrsb{s}}},
xmajorgrids,
ymin=-0.2, ymax=0.3, ytick={-0.2,-0.1,0,0.1,0.2,0.3},
ylabel={\nm{\Delta\xiest, \, \Delta\xivis \, \lrsb{^{\circ}}}},
ymajorgrids,
axis lines=left,
axis line style={-stealth},
legend entries={\nm{\mun{\Delta\xiest} \pm \sigman{\Delta\xiest}} \hypertt{INSE},
                \nm{\mui{\Delta\xivis} \pm \sigmai{\Delta\xivis}} \hypertt{VNSE},
				\nm{\mui{\Delta\xivis} \pm \sigmai{\Delta\xivis}} \hypertt{IA-VNSE}},
legend columns=3,
legend style={font=\footnotesize},
legend cell align=left,
]
\pgfplotstableread{figs/ch18_res_vis_assist/error_vis_att/error_vis_base_xi_deg_nav.txt}\mytablenav
\pgfplotstableread{figs/ch18_res_vis_assist/error_vis_att/error_vis_base_xi_deg_vis_relocalization.txt}\mytablevispartial
\pgfplotstableread{figs/ch18_res_vis_assist/error_vis_att/error_vis_base_xi_deg_vis.txt}\mytablevis
\draw [] (0.0,0.0) -- (3800.0,0.0);
\addplot table [header=false, x index=0,y index=1] {\mytablenav};
\addplot table [header=false, x index=0,y index=1] {\mytablevispartial};
\addplot table [header=false, x index=0,y index=1] {\mytablevis};
\addplot table [header=false, x index=0,y index=2] {\mytablenav};
\addplot table [header=false, x index=0,y index=3] {\mytablenav};
\addplot table [header=false, x index=0,y index=2] {\mytablevispartial};
\addplot table [header=false, x index=0,y index=3] {\mytablevispartial};
\addplot table [header=false, x index=0,y index=2] {\mytablevis};
\addplot table [header=false, x index=0,y index=3] {\mytablevis};
\path node [draw, shape=rectangle, fill=white] at (750,0.25) {\footnotesize Scenario \nm{\#1} \hypertt{MX}};
\end{axis}   
\end{tikzpicture}
\caption{Body Euler angles \texttt{INSE}, \texttt{VNSE}, and \texttt{IA-VNSE} for scenario \nm{\#1} \texttt{MX} (100 runs)}
\label{fig:Res_vis_assist_results_euler_individual}
\end{figure}

Table \ref{tab:Res_vis_assist_results_euler} shows the \hypertt{NSE} at the conclusion of both scenarios for the three Euler angles representing the body attitude (yaw \nm{\psi}, pitch \nm{\theta}, roll \nm{\xi})\footnote{The yaw angle estimation errors respond to \nm{\Delta\psiest = \psiest - \psi} and \nm{\Delta\psivis = \psivis - \psi}, respectively. Those for the body pitch and roll angles are defined accordingly.}, as well as the norm of the rotation vector between the real body attitude \nm{\qNB} and its estimations, \nm{\qNBest} by the \hypertt{INS} and \nm{\qNBvis} by the \hypertt{VNS} or \hypertt{IA-VNS}. The errors hence can be formally written as \nm{\DeltarNBBestnorm = \|\qNBest \ominus \qNB\|} or \nm{\DeltarNBBvisnorm = \|\qNBvis \ominus \qNB\|} \cite{LIE}. In addition, figures \ref{fig:Res_vis_assist_results_euler} and \ref{fig:Res_vis_assist_results_euler_alter} depict the variation with time of the body attitude \hypertt{NSE} for both scenarios, while figure \ref{fig:Res_vis_assist_results_euler_individual} shows those of each individual Euler angle for scenario \nm{\#1} exclusively. 

\begin{itemize}
\item After a short transition period following the introduction of \hypertt{GNSS}-Denied conditions at \nm{\tGNSS = 100 \, s}, the body attitude \hypertt{INSE} (blue lines) does not experience any drift with time in either scenario, and is bounded by the quality of the onboard sensors and the inertial navigation algorithms \cite{INSE}.

\item With respect to the \hypertt{VNSE} (red lines), most of the scenario \nm{\#1} error is incurred during the turn maneuver at the beginning of the scenario (refer to \nm{\tTURN} within \cite{SIMULATION}), with only a slow accumulation during the rest of the trajectory, composed by a long straight flight with punctual changes in altitude and speed. Additional error growth would certainly accumulate if more turns were to occur, although this is not tested in the simulation. This statement seems to contradict the results obtained with scenario \nm{\#2}, in which the error grows with the initial turns but then stabilizes during the rest of the scenario, even though the aircraft is executing continuous turn maneuvers. This lack of error growth occurs because the scenario \nm{\#2} trajectories are so twisted (refer to \cite{SIMULATION}) that terrain zones previously mapped reappear in the camera field of view during the consecutive turns, and are hence employed by the pose optimization phase as absolute references, resulting in a much better attitude estimation that what would occur under more spaced turns. A more detailed analysis (not shown in the figures) shows that the estimation error does not occur during the whole duration of the turns, but only during the roll-in and final roll-out maneuvers, where the optical flow is highest and hence more difficult to track by \hypertt{SVO}\footnote{For the two evaluated scenarios, the optical flow during the roll-in and roll-out maneuvers is significantly higher than that induced by straight flight, pull-up and push-down maneuvers, and even the turning maneuvers themselves (once the bank angle is no longer changing).}.
\end{itemize}
\begin{center}
\begin{tabular}{lrrrrrrrrrrrr}
\hline
 & \multicolumn{4}{c}{\hypertt{INSE}} & \multicolumn{4}{c}{\hypertt{VNSE}} & \multicolumn{4}{c}{\hypertt{IA-VNSE}} \\
\nm{\lrsb{^{\circ}}} & \nm{\Delta\psiest} & \nm{\Delta\thetaest} & \nm{\Delta\xiest} & \nm{\DeltarNBBestnorm} & \nm{\Delta\psivis} & \nm{\Delta\thetavis} & \nm{\Delta\xivis} & \nm{\DeltarNBBvisnorm} & \nm{\Delta\psivis} & \nm{\Delta\thetavis} & \nm{\Delta\xivis} & \nm{\DeltarNBBvisnorm} \\
\hline
\multicolumn{4}{l}{Scenario \#1 \hypertt{MX} \nm{\lrp{\tEND}}} \\
mean & +0.03 & -0.03 & -0.00 & \textbf{0.158} & +0.03 & +0.08 & +0.00 & \textbf{0.296} & +0.03 & -0.01 & -0.03 & \textbf{0.218} \\
std  &  0.18 &  0.05 &  0.06 &         0.114  &  0.13 &  0.23 &  0.21 &         0.158  &  0.11 &  0.16 &  0.14 &         0.103  \\
max  & -0.61 & -0.27 & -0.23 &         0.611  & +0.63 & +0.74 & +0.78 &         0.791  & +0.55 & -0.37 & -0.51 &         0.606  \\     
\hline
\multicolumn{4}{l}{Scenario \#2 \hypertt{MX} \nm{\lrp{\tEND}}} \\
mean & -0.02 & +0.01 & +0.00 & \textbf{0.128} & +0.02 & -0.02 & +0.00 & \textbf{0.253} & +0.02 & -0.00 & +0.01 & \textbf{0.221} \\
std  &  0.13 &  0.05 &  0.05 &         0.078  &  0.08 &  0.21 &  0.20 &         0.161  &  0.08 &  0.16 &  0.19 &         0.137  \\
max  & +0.33 & -0.15 & +0.15 &         0.369  & +0.22 & -0.65 & -0.73 &         0.730  & +0.24 & +0.62 & +0.74 &         0.788  \\
\hline
\end{tabular}
\end{center}
\captionof{table}{Aggregated \texttt{MX} final body attitude \texttt{INSE}, \texttt{VNSE}, and \texttt{IA-VNSE} (100 runs)} \label{tab:Res_vis_assist_results_euler}

\begin{itemize}
\item The \hypertt{IA-VNSE} results (green lines) show that the introduction of priors in section \ref{sec:IA-SVO} works as intended and there exists a clear benefit for the use of an \hypertt{IA-VNS} when compared to the stand alone \hypertt{VNS} described in section \ref{sec:SVO}. In spite of \hypertt{IA-VNSE} values at the beginning of both scenarios that are nearly double those of the \hypertt{VNSE} (refer to figures \ref{fig:Res_vis_assist_results_euler} and \ref{fig:Res_vis_assist_results_euler_alter}), caused by the initial pitch adjustment required to improve the fit between the homography output and the inertial estimations (section \ref{subsec:IA-SVO_extra}), the balance between both errors quickly flips as soon as the aircraft starts maneuvering, resulting in body attitude \hypertt{IA-VNSE} values significantly lower than those of the \hypertt{VNSE} for the remaining part of both scenarios. This improvement is more significant in the case of scenario \nm{\#1}, as the prior based pose optimization is by design a slow adjustment that requires significant time to slowly correct attitude and altitude deviations between the visual and inertial estimations.
\end{itemize}
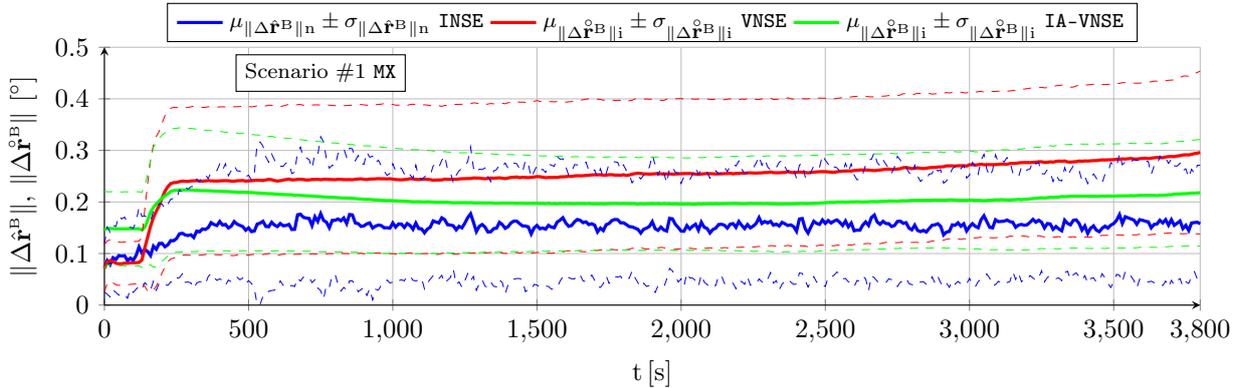
\begin{figure}[h]
\centering
\pgfplotsset{
	every axis legend/.append style={
		at={(0.50,1.02)},
		anchor=south,
	},
}
\begin{tikzpicture}
\begin{axis}[
cycle list={
            {blue,no markers,very thick},
            {red,no markers,very thick},
            {green,no markers,very thick},
            {blue,dashed,no markers,ultra thin},{blue,dashed,no markers,ultra thin},
            {red,dashed,no markers,ultra thin},{red,dashed,no markers,ultra thin},
            {green,dashed,no markers,ultra thin},{green,dashed,no markers,ultra thin}},
width=16.0cm,
height=5.0cm,
xmin=0, xmax=3800, xtick={0,500,...,3500,3800},
xlabel={\nm{t \lrsb{s}}},
xmajorgrids,
ymin=0, ymax=0.5, ytick={0,0.1,0.2,0.3,0.4,0.5},
ylabel={\nm{\DeltarBestnorm, \, \DeltarBvisnorm \, \lrsb{^{\circ}}}},
ymajorgrids,
axis lines=left,
axis line style={-stealth},
legend entries={
				\nm{\mun{\DeltarBestnorm} \pm \sigman{\DeltarBestnorm}} \hypertt{INSE},
                \nm{\mui{\DeltarBvisnorm} \pm \sigmai{\DeltarBvisnorm}} \hypertt{VNSE},
				\nm{\mui{\DeltarBvisnorm} \pm \sigmai{\DeltarBvisnorm}} \hypertt{IA-VNSE}},
legend columns=3,
legend style={font=\footnotesize},
legend cell align=left,
]
\pgfplotstableread{figs/ch18_res_vis_assist/error_vis_att/error_vis_base_euler_deg_nav.txt}\mytablenav
\pgfplotstableread{figs/ch18_res_vis_assist/error_vis_att/error_vis_base_euler_deg_vis_relocalization.txt}\mytablevispartial
\pgfplotstableread{figs/ch18_res_vis_assist/error_vis_att/error_vis_base_euler_deg_vis.txt}\mytablevis
\addplot table [header=false, x index=0,y index=1] {\mytablenav};
\addplot table [header=false, x index=0,y index=1] {\mytablevispartial};
\addplot table [header=false, x index=0,y index=1] {\mytablevis};
\addplot table [header=false, x index=0,y index=2] {\mytablenav};
\addplot table [header=false, x index=0,y index=3] {\mytablenav};
\addplot table [header=false, x index=0,y index=2] {\mytablevispartial};
\addplot table [header=false, x index=0,y index=3] {\mytablevispartial};
\addplot table [header=false, x index=0,y index=2] {\mytablevis};
\addplot table [header=false, x index=0,y index=3] {\mytablevis};
\path node [draw, shape=rectangle, fill=white] at (750,0.45) {\footnotesize Scenario \nm{\#1} \hypertt{MX}};
\end{axis}   
\end{tikzpicture}
\caption{Body attitude \texttt{INSE}, \texttt{VNSE}, and \texttt{IA-VNSE} for scenario \nm{\#1} \texttt{MX} (100 runs)}
\label{fig:Res_vis_assist_results_euler}
\end{figure}

Qualitatively, the biggest difference between the three estimations resides in the nature of the errors. While the attitude \hypertt{INSE} is bounded, drift is present in both the \hypertt{VNS} and \hypertt{IA-VNS} estimations. The drift resulting from the Monte Carlo simulations may be small, and so is the attitude estimation error \nm{\DeltarNBBvisnorm}, but more challenging conditions with more drastic maneuvers and a less idealized image generation process than that described in section \ref{sec:Simulation} may generate additional drift.

Focusing now on the quantitative results shown in table \ref{tab:Res_vis_assist_results_euler}, aggregated errors for each individual Euler angle are always unbiased and zero mean for each of the three estimations (\hypertt{INS}, \hypertt{VNS}, \hypertt{IA-VNS}), as the means tend to zero as the number of runs grows, and are much smaller than both the standard deviations and the maximum values. With respect to the attitude error \nm{\DeltarNBBestnorm} and \nm{\DeltarNBBvisnorm}, their aggregated means are not zero (they are norms), but are nevertheless quite repetitive in all three cases, as the mean is always significantly higher than the standard deviation, while the maximum values only represent a small multiple of the means. It is interesting to point out that while in the case of the \hypertt{INSE} the contribution of the yaw error is significantly higher than that of the pitch and roll errors, the opposite occurs for both the \hypertt{VNSE} and the \hypertt{IA-VNSE}. This makes sense as the the gravity direction is employed by the \hypertt{INS} as a reference from where the estimated pitch and roll angles can not deviate, but slow changes in yaw generate larger optical flow variations than those caused by pitch and roll variations.

These results prove that the \hypertt{IA-VNS} succeeds when employing the inertially observed pitch and bank angles (\nm{\thetaest}, \nm{\xiest}), whose errors are bounded, to limit the drift of their visual counterparts (\nm{\thetavis}, \nm{\xivis}), as \nm{\sigmaEND{\thetavis}} and \nm{\sigmaEND{\xivis}}\footnote{As the individual Euler angle metrics are unbiased or zero mean, the benefits of the proposed approach are reflected in the variation of the remaining metrics, this is, the standard deviation and the maximum value.} are significantly lower for the \hypertt{IA-VNS} than for the \hypertt{VNS}. Remarkably, this is achieved with no degradation in the body yaw angle, as \nm{\sigmaEND{\psivis}} remains stable. Note that adjusting the output of certain variables in a minimization algorithm (such as pose optimization) usually results in a degradation in the accuracy of the remaining variables as the solution moves away from the true optimum. In this case, however, the improved fit between the adjusted aircraft pose and the terrain displayed in the images, results in the \hypertt{SVO} pipeline also slightly improving its body yaw estimation \nm{\psivis}. Section \ref{subsec:Results_hor} shows how the benefits of an improved fit between the displayed terrain and the adjusted pose also improve the horizontal position estimation.

In the case of a real life scenario based on a more realistic image generation process than that described in section \ref{sec:Simulation}, the \hypertt{VNS} would likely incur in additional body attitude drift than in the simulations. If this were to occur, the \hypertt{IA-VNS} pose adjustment algorithms described in section \ref{sec:IA-SVO} would react more aggressively to counteract the higher pitch and bank deviations, eliminating most of the extra drift, although it is possible that higher pose adjustment parameters than those listed in table \ref{tab:nav_vis_assist_priors_activation_pitch} would be required. The \hypertt{IA-VNS} is hence more resilient against high drift values than the \hypertt{VNS}.


\subsection{Vertical Position Estimation}\label{subsec:Results_h}

Table \ref{tab:Res_vis_assist_results_h} contains the vertical position \hypertt{NSE} (\nm{\Delta\hest = \hest - h}, \nm{\Delta\hvis = \hvis - h}) at the conclusion of both scenarios, which can be considered unbiased or zero mean in all six cases (two scenarios and three estimation methods) as the mean \nm{\muEND{h}} is always significantly lower than both the standard deviation \nm{\sigmaEND{h}} or the maximum value {\nm{\maxEND{h}}. The \hypertt{NSE} evolution with time is depicted in figures \ref{fig:Res_vis_assist_results_h} and \ref{fig:Res_vis_assist_results_h_alter}, which also include (magenta lines) those Monte Carlo executions that result in the highest \hypertt{IA-VNSE}s. 
\begin{center}
\begin{tabular}{llrcrcr}
\hline
\multicolumn{2}{l}{Scenario \hypertt{MX} \nm{\lrp{\tEND}}} & \multicolumn{1}{c}{\hypertt{INSE}} & & \multicolumn{1}{c}{\hypertt{VNSE}} & & \multicolumn{1}{c}{\hypertt{IA-VNSE}} \\
& \nm{\lrsb{m}} & \multicolumn{1}{c}{\nm{\Delta\hest}} & & \multicolumn{1}{c}{\nm{\Delta\hvis}} & & \multicolumn{1}{c}{\nm{\Delta\hvis}} \\
\hline
\multirow{3}{*}{\#1} & mean &          -4.18  & &          +82.91  & &          +22.86  \\
                     & std  & \textbf{ 25.78} & & \textbf{ 287.58} & & \textbf{  49.17} \\
                     & max  &         -70.49  & &         +838.32  & &         +175.76  \\
\hline
\multirow{3}{*}{\#2} & mean &          +0.76  & &          +3.45  & &          +3.59  \\
                     & std  & \textbf{  7.55} & & \textbf{ 20.56} & & \textbf{ 13.01} \\
                     & max  &         -19.86  & &         +72.69  & &         +71.64  \\
\hline
\end{tabular}
\end{center}

\captionof{table}{Aggregated \texttt{MX} final vertical position \texttt{INSE}, \texttt{VNSE}, and \texttt{IA-VNSE} (100 runs)} \label{tab:Res_vis_assist_results_h}

\begin{itemize}
\item The geometric altitude \hypertt{INSE} (blue lines) is bounded by the change in atmospheric pressure offset since the time the \hypertt{GNSS} signals are lost. Refer to \cite{INSE} for additional information.

\item The \hypertt{VNS} estimation of the geometric altitude (red lines) is worse than that by the \hypertt{INS} both qualitatively and quantitatively, even with the results being optimistic because of the ideal image generation process employed in the simulation. A continuous drift or error growth with time is present, and results in final errors much higher than those obtained with the \hypertt{GNSS}-Denied inertial filter. These errors are logically bigger for scenario \nm{\#1} because of its much longer duration.
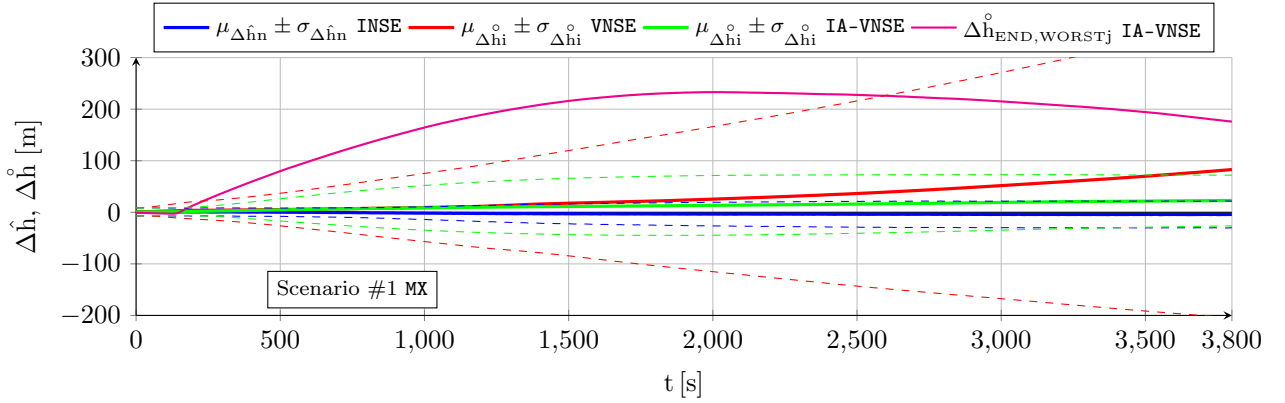
\begin{figure}[h]
\centering
\pgfplotsset{
	every axis legend/.append style={
		at={(0.50,1.02)},
		anchor=south,
	},
}
\begin{tikzpicture}
\begin{axis}[
cycle list={
            {blue,no markers,very thick},
            {red,no markers,very thick},
            {green,no markers,very thick},
            {magenta,no markers,thick},
            {blue,dashed,no markers},{blue,dashed,no markers},
            {red,dashed,no markers},{red,dashed,no markers},
            {green,dashed,no markers},{green,dashed,no markers}},
width=16.0cm,
height=5.0cm,
xmin=0, xmax=3800, xtick={0,500,...,3500,3800},
xlabel={\nm{t \lrsb{s}}},
xmajorgrids,
ymin=-200, ymax=300, ytick={-200,-100,...,300},
ylabel={\nm{\Delta\hest, \, \Delta\hvis \, \lrsb{m}}},
ymajorgrids,
axis lines=left,
axis line style={-stealth},
legend entries={
				\nm{\mun{\Delta\hest} \pm \sigman{\Delta\hest}} \hypertt{INSE},
                \nm{\mui{\Delta\hvis} \pm \sigmai{\Delta\hvis}} \hypertt{VNSE},
				\nm{\mui{\Delta\hvis} \pm \sigmai{\Delta\hvis}} \hypertt{IA-VNSE},
                \nm{\XENDjworst{\Delta\hvis}} \hypertt{IA-VNSE}},
legend columns=4,
legend style={font=\footnotesize},
legend cell align=left,
]
\draw [] (0.0,0.0) -- (3800.0,0.0);
\pgfplotstableread{figs/ch18_res_vis_assist/error_vis_h/error_vis_base_h_m_nav.txt}\mytablenav
\pgfplotstableread{figs/ch18_res_vis_assist/error_vis_h/error_vis_base_h_m_vis_relocalization.txt}\mytablevispartial
\pgfplotstableread{figs/ch18_res_vis_assist/error_vis_h/error_vis_base_h_m_vis.txt}\mytablevis
\addplot table [header=false, x index=0,y index=1] {\mytablenav};
\addplot table [header=false, x index=0,y index=1] {\mytablevispartial};
\addplot table [header=false, x index=0,y index=1] {\mytablevis};
\addplot table [header=false, x index=0,y index=5] {\mytablevis};
\addplot table [header=false, x index=0,y index=2] {\mytablenav};
\addplot table [header=false, x index=0,y index=3] {\mytablenav};
\addplot table [header=false, x index=0,y index=2] {\mytablevispartial};
\addplot table [header=false, x index=0,y index=3] {\mytablevispartial};
\addplot table [header=false, x index=0,y index=2] {\mytablevis};
\addplot table [header=false, x index=0,y index=3] {\mytablevis};
\path node [draw, shape=rectangle, fill=white] at (750,-150) {\footnotesize Scenario \nm{\#1} \hypertt{MX}};
\end{axis}   
\end{tikzpicture}
\caption{Vertical position \texttt{INSE}, \texttt{VNSE}, and \texttt{IA-VNSE} for scenario \nm{\#1} \texttt{MX} (100 runs)}
\label{fig:Res_vis_assist_results_h}
\end{figure}

A small percentage of this drift can be attributed to the slow accumulation of error inherent to the \hypertt{SVO} motion thread algorithms introduced in section \ref{sec:SVO}, but most of it results from adding the estimated relative pose between two consecutive images to a pose (that of the previous image) with an attitude that already possesses a small pitch error (refer to the attitude estimation analysis in section \ref{subsec:Results_att}). Note that even a fraction of a degree deviation in pitch can result in hundreds of meters in vertical error when applied to the total distance flown in scenario \nm{\#1}, as \hypertt{SVO} can be very precise when estimating pose changes between consecutive images, but lacks any absolute reference to avoid slowly accumulating these errors over time. This fact is precisely the reason why the vertical position \hypertt{VNSE} grows more slowly in the second half of scenario \nm{\#2}, as shown in figure \ref{fig:Res_vis_assist_results_h_alter}. As explained in section \ref{subsec:Results_att} above, continuous turn maneuvers cause previously mapped terrain points to reappear in the camera field of view, stopping the growth in the attitude error (pitch included), which indirectly has the effect of slowing the growth in altitude estimation error.

\item The benefits of introducing priors to limit the differences between the visual and inertial altitude estimations are reflected in the \hypertt{IA-VNSE} (green lines). The error reduction is drastic in the case of the scenario \nm{\#1}, where its extended duration allows the pose optimization small pitch adjustments to accumulate into significant altitude corrections over time, and less pronounced but nevertheless significant for scenario \nm{\#2}, where the \hypertt{VNSE} (an hence also the \hypertt{IA-VNSE}) results already benefit from previously mapped terrain points reappearing in the aircraft field of view as a result of the continuous maneuvers. It is necessary to remark the amount of the improvement, as the final standard deviation \nm{\sigmaEND{h}} diminishes from \nm{287.58} to \nm{49.17 \, m} for scenario \nm{\#1}, and from \nm{20.56} to \nm{13.01 \, m} for scenario \nm{\#2}.
\end{itemize}

The benefits of the prior based pose optimization algorithm can be clearly observed in the case of the scenario \nm{\#1} execution with the worst final altitude estimation error, whose error variation with time is depicted in figure \ref{fig:Res_vis_assist_results_h} (magenta line). After a rapid growth in the first third of the scenario following a particularly negative estimation during the initial turn, the altitude error reaches a maximum of \nm{+233.05 \, m} at \nm{2007.5 \, s}. Attitude adjustment has become active long before, lowering the estimated pitch angle to first diminish the growth of the altitude error and then being able to reduce the error itself, reaching a final value of \nm{+175.76 \, m} at \nm{\tEND}. As soon as the differences between the visual pitch, bank, or altitude estimations (\nm{\thetavis, \, \xivis, \, \hvis}) and their inertial counterparts (\nm{\thetaest, \, \xiest, \, \hest}) exceed certain limits (section \ref{sec:IA-SVO}), the attitude adjustment comes into play and slowly adjusts the aircraft pitch to prevent the visual altitude from deviating in excess from the inertial one. This behavior not only improves the \hypertt{IA-VNS} altitude estimation accuracy when compared to that of the \hypertt{VNS}, but also its resilience, as the system actively opposes elevated altitude errors. 
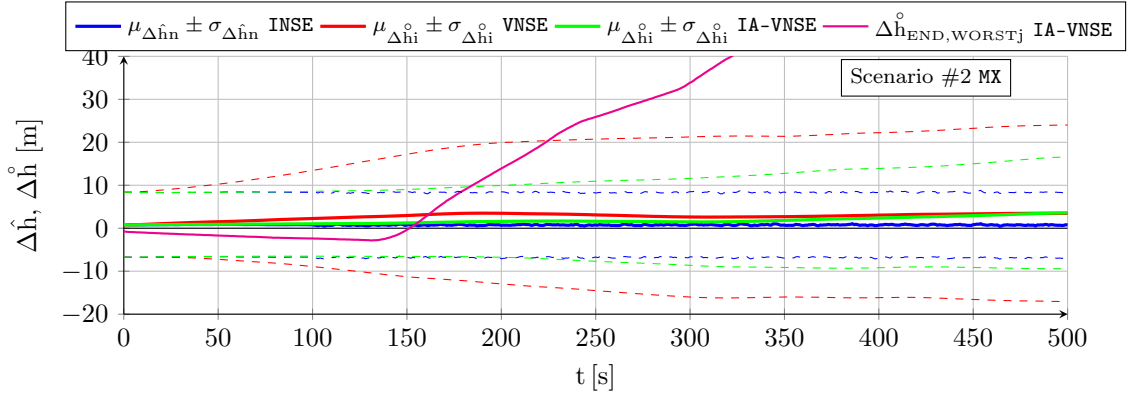
\begin{figure}[h]
\centering
\pgfplotsset{
	every axis legend/.append style={
		at={(0.50,1.02)},
		anchor=south,
	},
}
\begin{tikzpicture}
\begin{axis}[
cycle list={
            {blue,no markers,very thick},
            {red,no markers,very thick},
            {green,no markers,very thick},
            {magenta,no markers,thick},
            {blue,dashed,no markers},{blue,dashed,no markers},
            {red,dashed,no markers},{red,dashed,no markers},
            {green,dashed,no markers},{green,dashed,no markers}},
width=14.0cm,
height=5.0cm,
xmin=0, xmax=500, xtick={0,50,...,500},
xlabel={\nm{t \lrsb{s}}},
xmajorgrids,
ymin=-20, ymax=40, ytick={-20,-10,...,40},
ylabel={\nm{\Delta\hest, \, \Delta\hvis \, \lrsb{m}}},
ymajorgrids,
axis lines=left,
axis line style={-stealth},
legend entries={
				\nm{\mun{\Delta\hest} \pm \sigman{\Delta\hest}} \hypertt{INSE},
                \nm{\mui{\Delta\hvis} \pm \sigmai{\Delta\hvis}} \hypertt{VNSE},
				\nm{\mui{\Delta\hvis} \pm \sigmai{\Delta\hvis}} \hypertt{IA-VNSE},
                \nm{\XENDjworst{\Delta\hvis}} \hypertt{IA-VNSE}},
legend columns=4,
legend style={font=\footnotesize},
legend cell align=left,
]
\draw [] (0.0,0.0) -- (500.0,0.0);
\pgfplotstableread{figs/ch18_res_vis_assist/error_vis_h/error_vis_alter_h_m_nav.txt}\mytablenav
\pgfplotstableread{figs/ch18_res_vis_assist/error_vis_h/error_vis_alter_h_m_vis_relocalization.txt}\mytablevispartial
\pgfplotstableread{figs/ch18_res_vis_assist/error_vis_h/error_vis_alter_h_m_vis.txt}\mytablevis
\addplot table [header=false, x index=0,y index=1] {\mytablenav};
\addplot table [header=false, x index=0,y index=1] {\mytablevispartial};
\addplot table [header=false, x index=0,y index=1] {\mytablevis};
\addplot table [header=false, x index=0,y index=5] {\mytablevis};
\addplot table [header=false, x index=0,y index=2] {\mytablenav};
\addplot table [header=false, x index=0,y index=3] {\mytablenav};
\addplot table [header=false, x index=0,y index=2] {\mytablevispartial};
\addplot table [header=false, x index=0,y index=3] {\mytablevispartial};
\addplot table [header=false, x index=0,y index=2] {\mytablevis};
\addplot table [header=false, x index=0,y index=3] {\mytablevis};
\path node [draw, shape=rectangle, fill=white] at (425,35) {\footnotesize Scenario \nm{\#2} \hypertt{MX}};
\end{axis}   
\end{tikzpicture}
\caption{Vertical position \texttt{INSE}, \texttt{VNSE}, and \texttt{IA-VNSE} for scenario \nm{\#2} \texttt{MX} (100 runs)}
\label{fig:Res_vis_assist_results_h_alter}
\end{figure}

Significantly better altitude estimation errors (closer to the inertial ones) could be obtained if more aggressive settings were employed for \nm{\Delta \circled{\theta}_{1,MAX}^{\circ\circ}} and \nm{\Delta \circled{\theta}_{2,MAX}^{\circ\circ}} within table \ref{tab:nav_vis_assist_priors_activation_pitch}, as the selected values are far from the level at which the pose optimization convergence is compromised. This would result in more aggressive adjustments and important accuracy improvements for those cases in which altitude error growth is highest. The settings employed in this article are modest, as the final objective is not to obtain the smallest possible attitude or vertical position \hypertt{IA-VNSE} (as they are always bigger than their \hypertt{INSE} counterparts), but to limit them to acceptable levels so \hypertt{SVO} can build a more accurate terrain map, improving the fit between the multiple terrain 3D points displayed in the images and the estimated aircraft pose. To do so it is mandatory to balance the pitch and bank angle adjustments with the need to stick to solutions close to those that minimize the reprojection error, as explained in section \ref{sec:IA-SVO}. Higher \nm{\Delta \circled{\theta}_{1,MAX}^{\circ\circ}} and \nm{\Delta \circled{\theta}_{2,MAX}^{\circ\circ}} accelerate the adjustments but may decrease the quality of the map. It is expected that a better rendition of the real 3D position of the features detected in the keyframes as they are tracked along successive images will lower the incremental horizontal displacement errors, and hence result in a lower horizontal position \hypertt{IA-VNSE}, which is the real objective for the introduction of the priors.

The \hypertt{IA-VNS} altitude estimation improvements over those of the \hypertt{VNS} are not only quantitative. Figure \ref{fig:Res_vis_assist_results_h} shows no increment in \nm{\sigman{\hvis}} (green lines) in the second half of scenario \nm{\#1} (once on average the deviation has activated the attitude adjustment feature). The altitude estimation by the \hypertt{IA-VNS} can hence also be described as bounded and driftless, which represents a qualitative and not only quantitative improvement over that of the \hypertt{VNS}. The bounds are obviously bigger for the \hypertt{IA-VNS} than for the \hypertt{INS}. In the case of scenario \nm{\#2}, figure \ref{fig:Res_vis_assist_results_h_alter} shows a slow but steady \nm{\sigman{\hvis}} growth with time, but this is only because the error amount on average is not yet significant enough to activate the attitude adjustment feature within pose optimization. 


\subsection{Horizontal Position Estimation}\label{subsec:Results_hor}

The horizontal position estimation capabilities of the \hypertt{INS}, \hypertt{VNS}, and \hypertt{IA-VNS} share the fact that all of them exhibit an unrestrained drift or growth with time, as shown in figures \ref{fig:Res_vis_assist_results_hor} and \ref{fig:Res_vis_assist_results_hor_alter}. The errors obtained at the end of both scenarios are shown in table \ref{tab:Res_vis_assist_results_hor}, following the same scheme as in previous sections. While the approximately linear \hypertt{INS} drift appears when integrating the bounded ground velocity errors \cite{INSE}, the visual drifts (both \hypertt{VNS} and \hypertt{IA-VNS}) originate in the slow accumulation of errors caused by the concatenation of the relative poses between consecutive images without absolute references, but also show a direct relationship with the scale error committed when estimating the aircraft height over the terrain during the initial homography. 
\begin{center}
\begin{tabular}{llrrrcrrcrr}
\hline
\multicolumn{3}{l}{Scenario \hypertt{MX} \nm{\lrp{\tEND}}} &  \multicolumn{2}{c}{\hypertt{INSE}} & \ \ \ & \multicolumn{2}{c}{\hypertt{VNSE}} & & \multicolumn{2}{c}{\hypertt{IA-VNSE}} \\
& & \multicolumn{1}{c}{Distance} & \multicolumn{2}{c}{\textbf{\nm{\Delta \hat{x}_{\sss HOR}}}} & & \multicolumn{2}{c}{\textbf{\nm{\Delta \circled{x}_{\sss HOR}}}} & & \multicolumn{2}{c}{\textbf{\nm{\Delta \circled{x}_{\sss HOR}}}} \\
& & \multicolumn{1}{c}{\nm{\lrsb{m}}} & \multicolumn{1}{c}{\nm{\lrsb{m}}} & \multicolumn{1}{c}{\nm{\lrsb{\%}}} & & \multicolumn{1}{c}{\nm{\lrsb{m}}} & \multicolumn{1}{c}{\nm{\lrsb{\%}}} & & \multicolumn{1}{c}{\nm{\lrsb{m}}} & \multicolumn{1}{c}{\nm{\lrsb{\%}}} \\
\hline
\multirow{3}{*}{\nm{\#1}} & mean & 107873 &  7276 & \textbf{7.10} & &  4179 &  \textbf{3.82} & &  488 & \textbf{0.46} \\
                          & std  &  19756 &  4880 &  5.69 & &  3308 &  2.73 & &  350 & 0.31 \\
                          & max  & 172842 & 25288 & 32.38 & & 21924 & 14.22 & & 1957 & 1.48 \\
\hline
\multirow{3}{*}{\nm{\#2}} & mean & 14198 & 216 & \textbf{1.52} & & 251 & \textbf{1.77} & &  33 & \textbf{0.23} \\
                          & std  &  1176 & 119 & 0.86 & & 210 & 1.48 & &  26 & 0.18 \\
                          & max  & 18253 & 586 & 4.38 & & 954 & 7.08 & & 130 & 0.98 \\
\hline
\end{tabular}
\end{center}

\captionof{table}{Aggregated \texttt{MX} final horizontal position \texttt{INSE}, \texttt{VNSE}, and \texttt{IA-VNSE} (100 runs)} \label{tab:Res_vis_assist_results_hor}

In the case of the \hypertt{VNS} (red lines), its scenario \nm{\#1} horizontal position estimations appear to be significantly more accurate than those of the \hypertt{INS} (blue lines). Note however that the ideal image generation process discussed in section \ref{sec:Simulation} implies that the simulation results should be treated as a best case only, and that the results obtained in real world conditions would likely imply a higher horizontal position drift. The drift experienced by the  \hypertt{VNS} in figure \ref{fig:Res_vis_assist_results_hor_alter} (scenario \nm{\#2}) also shows the same diminution in its slope in the second half of the scenario discussed in previous sections, which is attributed to previously mapped terrain points reappearing in the camera field of view as a consequence of the continuous turns present in scenario \nm{\#2}. Additionally, notice how the \hypertt{VNSE} starts growing at the beginning of the scenario, while the \hypertt{INSE} only starts doing so after the \hypertt{GNSS} signals are lost at \nm{\tGNSS = 100 \, s} \cite{INSE}.
\begin{figure}[h]
\centering
\pgfplotsset{
	every axis legend/.append style={
		at={(0.19,0.55)},
		anchor=south,
	},
}
\begin{tikzpicture}
\begin{axis}[
cycle list={
			{blue,no markers,very thick},
            {red,no markers,very thick},
            {green,no markers,very thick},
            {magenta,no markers,thick},
            {blue,dashed,no markers,ultra thin},{blue,dashed,no markers,ultra thin},
            {red,dashed,no markers,ultra thin},{red,dashed,no markers,ultra thin},
            {green,dashed,no markers,ultra thin},{green,dashed,no markers,ultra thin}},
width=16.0cm,
height=5.5cm,
xmin=0, xmax=3800, xtick={0,500,...,3500,3800},
xlabel={\nm{t \lrsb{sec}}},
xmajorgrids,
ymin=0, ymax=8000, ytick={0,2000,4000,6000,8000},
ylabel={\nm{\Deltaxhorest, \, \Deltaxhorvis \, \lrsb{m}}},
ymajorgrids,
axis lines=left,
axis line style={-stealth},
legend entries={
				\nm{\mun{\Deltaxhorest} \pm \sigman{\Deltaxhorest}} \hypertt{INSE},
				\nm{\mui{\Deltaxhorvis} \pm \sigmai{\Deltaxhorvis}} \hypertt{VNSE},
				\nm{\mui{\Deltaxhorvis} \pm \sigmai{\Deltaxhorvis}} \hypertt{IA-VNSE},
				\nm{\XENDjworst{\Deltaxhorvis}} \hypertt{IA-VNSE}},
legend columns=1,
legend style={font=\footnotesize},
legend cell align=left,
]
\pgfplotstableread{figs/ch18_res_vis_assist/error_vis_hor/error_vis_base_hor_m_pc_nav.txt}\mytablenav
\pgfplotstableread{figs/ch18_res_vis_assist/error_vis_hor/error_vis_base_hor_m_pc_vis_relocalization.txt}\mytablevispartial
\pgfplotstableread{figs/ch18_res_vis_assist/error_vis_hor/error_vis_base_hor_m_pc_vis.txt}\mytablevis
\addplot table [header=false, x index=0,y index=1] {\mytablenav};
\addplot table [header=false, x index=0,y index=1] {\mytablevispartial};
\addplot table [header=false, x index=0,y index=1] {\mytablevis};
\addplot table [header=false, x index=0,y index=5] {\mytablevis};
\addplot table [header=false, x index=0,y index=2] {\mytablenav};
\addplot table [header=false, x index=0,y index=3] {\mytablenav};
\addplot table [header=false, x index=0,y index=2] {\mytablevispartial};
\addplot table [header=false, x index=0,y index=3] {\mytablevispartial};
\addplot table [header=false, x index=0,y index=2] {\mytablevis};
\addplot table [header=false, x index=0,y index=3] {\mytablevis};
\path node [draw, shape=rectangle, fill=white] at (650,3200) {\footnotesize Scenario \nm{\#1} \hypertt{MX}};
\end{axis}   
\end{tikzpicture}
\caption{Horizontal position \texttt{INSE}, \texttt{VNSE}, and \texttt{IA-VNSE} for scenario \nm{\#1} \texttt{MX} (100 runs)}
\label{fig:Res_vis_assist_results_hor}
\end{figure}
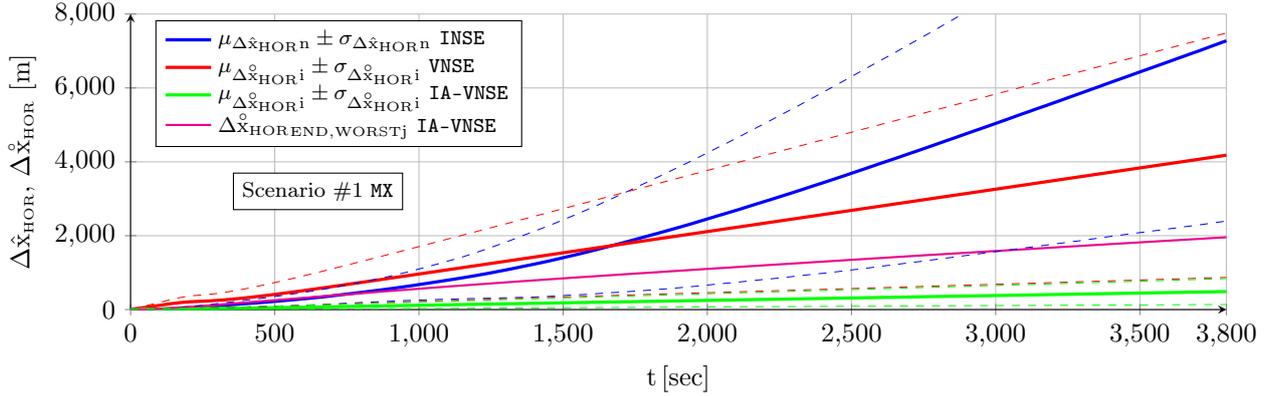

The \hypertt{IA-VNS} (green lines) results in major horizontal position estimation improvements over the \hypertt{VNS}. The final horizontal position error mean \nm{\muEND{\Deltaxhorvis}} diminishes from \nm{3.82} to \nm{0.46 \, \%} for scenario \nm{\#1}, and from \nm{1.77} to \nm{0.23 \, \%} for scenario \nm{\#2}. The repeatability of the results also improves, as the final standard deviation \nm{\sigmaEND{\Deltaxhorvis}} falls from \nm{2.73} to \nm{0.31 \, \%} and from \nm{1.48} to \nm{0.18 \, \%} for both scenarios. Note that although these results may be slightly optimistic due to the optimized image generation process, they are much more accurate than those obtained with the \hypertt{INS}, for which the error mean and standard deviation amount to \nm{7.10} and \nm{5.69 \, \%} for scenario \nm{\#1}, and \nm{1.52} and \nm{0.86 \, \%} in case of scenario \nm{\#2}.
\begin{figure}[h]
\centering
\pgfplotsset{
	every axis legend/.append style={
		at={(0.50,1.05)},
		anchor=south,
	},
}
\begin{tikzpicture}
\begin{axis}[
cycle list={
			{blue,no markers,very thick},
            {red,no markers,very thick},
            {green,no markers,very thick},
            {magenta,no markers,thick},
            {blue,dashed,no markers,ultra thin},{blue,dashed,no markers,ultra thin},
            {red,dashed,no markers,ultra thin},{red,dashed,no markers,ultra thin},
            {green,dashed,no markers,ultra thin},{green,dashed,no markers,ultra thin}},
width=14.0cm,
height=5.5cm,
xmin=0, xmax=500, xtick={0,50,...,500},
xlabel={\nm{t \lrsb{sec}}},
xmajorgrids,
ymin=0, ymax=400, ytick={0,100,...,400},
ylabel={\nm{\Deltaxhorest, \, \Deltaxhorvis \, \lrsb{m}}},
ymajorgrids,
axis lines=left,
axis line style={-stealth},
legend entries={
				\nm{\mun{\Deltaxhorest} \pm \sigman{\Deltaxhorest}} \hypertt{INSE},
				\nm{\mui{\Deltaxhorvis} \pm \sigmai{\Deltaxhorvis}} \hypertt{VNSE},
				\nm{\mui{\Deltaxhorvis} \pm \sigmai{\Deltaxhorvis}} \hypertt{IA-VNSE},
				\nm{\XENDjworst{\Deltaxhorvis}} \hypertt{IA-VNSE}},
legend columns=2,
legend style={font=\footnotesize},
legend cell align=left,
]
\pgfplotstableread{figs/ch18_res_vis_assist/error_vis_hor/error_vis_alter_hor_m_pc_nav.txt}\mytablenav
\pgfplotstableread{figs/ch18_res_vis_assist/error_vis_hor/error_vis_alter_hor_m_pc_vis_relocalization.txt}\mytablevispartial
\pgfplotstableread{figs/ch18_res_vis_assist/error_vis_hor/error_vis_alter_hor_m_pc_vis.txt}\mytablevis
\addplot table [header=false, x index=0,y index=1] {\mytablenav};
\addplot table [header=false, x index=0,y index=1] {\mytablevispartial};
\addplot table [header=false, x index=0,y index=1] {\mytablevis};
\addplot table [header=false, x index=0,y index=5] {\mytablevis};
\addplot table [header=false, x index=0,y index=2] {\mytablenav};
\addplot table [header=false, x index=0,y index=3] {\mytablenav};
\addplot table [header=false, x index=0,y index=2] {\mytablevispartial};
\addplot table [header=false, x index=0,y index=3] {\mytablevispartial};
\addplot table [header=false, x index=0,y index=2] {\mytablevis};
\addplot table [header=false, x index=0,y index=3] {\mytablevis};
\path node [draw, shape=rectangle, fill=white] at (70,360) {\footnotesize Scenario \nm{\#2} \hypertt{MX}};
\end{axis}   
\end{tikzpicture}
\caption{Horizontal position \texttt{INSE}, \texttt{VNSE}, and \texttt{IA-VNSE} for scenario \nm{\#2} \texttt{MX} (100 runs)}
\label{fig:Res_vis_assist_results_hor_alter}
\end{figure}
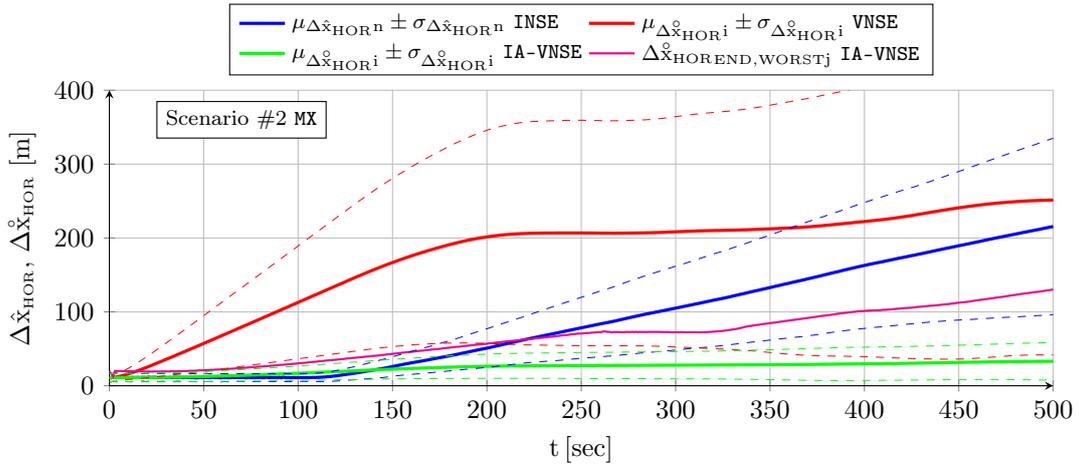

It is interesting to remark how the prior based pose optimization described in section \ref{sec:IA-SVO}, an algorithm that adjusts the aircraft pitch and bank angles based on deviations between the visually estimated pitch angle, bank angle, and geometric altitude, and their inertially estimated counterparts, is capable of not only improving the visual estimations of those three variables, but doing so with a minor improvement in the body yaw estimation and an extreme reduction in the horizontal position error. When the cost function within an optimization algorithm is modified to adjust certain target components, the expected result is that this can be achieved only at the expense of the accuracy in the remaining target components, not in addition to it. The reason why in this case all target components improve lies in that the adjustment creates a better fit between the ground terrain and associated 3D points depicted in the images on one side, and the estimated aircraft pose indicating the position and attitude from where the images are taken on the other.


\section{Influence of Terrain Type}\label{sec:Influence_terrain}

The type of terrain overflown by the aircraft has a significant influence on the performance of the visual navigation algorithms described in sections \ref{sec:SVO} and \ref{sec:IA-SVO}, which can not operate unless the feature detector is capable of periodically locating features in the various keyframes, and which also requires the depth filter to correctly estimate the 3D terrain coordinates of each feature. The terrain texture (or lack of) and its elevation relief are hence the two most important characteristics in this regard. To evaluate its influence, each of the scenario \#1 one hundred Monte Carlo runs is executed flying above four different zones or types of terrain\footnote{All section \ref{sec:Results} results have been obtained with the \say{mix} (\hypertt{MX}) zone.}, intended to represent a wide array of conditions; images representative of each zone as viewed by the onboard camera are included below. The use of terrains that differ in both their texture and vertical relief is intended to provide a more complete validation of the proposed algorithms. Note that the only variation among the different simulations is the terrain type, as all other parameters defining each scenario (mission, aircraft, sensors, weather, wind, turbulence, geophysics, initial estimations) are exactly the same for all simulation runs.
 
\begin{itemize}
\item The \say{desert} (\hypertt{DS}) zone is located in the Sonoran desert of southern Arizona (\hypertt{USA}) and northern Mexico. It is characterized by a combination of bajadas (broad slopes of debris) and isolated very steep mountain ranges. There is virtually no human infrastructure or flat terrain, as the bajadas have sustained slopes of up to \nm{7^{\circ}}. The altitude of the bajadas ranges from \nm{300} to \nm{800 \, m} above \hypertt{MSL}, and the mountains reach up to \nm{800 \, m} above the surrounding terrain. Texture is abundant because of the cacti and the vegetation along the dry creeks.

\item The \say{farm} (\hypertt{FM}) zone is located in the fertile farmland of southeastern Illinois and southwestern Indiana (\hypertt{USA}). A significant percentage of the terrain is made of regular plots of farmland, but there also exists some woodland, farm houses, rivers, lots of little towns, and roads. It is mostly flat with an altitude above \hypertt{MSL} between \nm{100} and \nm{200 \, m}, and altitude changes are mostly restricted to the few forested areas. Texture is nonexistent in the farmlands, where extracting features is often impossible.
\begin{figure}[h]
\centering
\includegraphics[width=6.5cm]{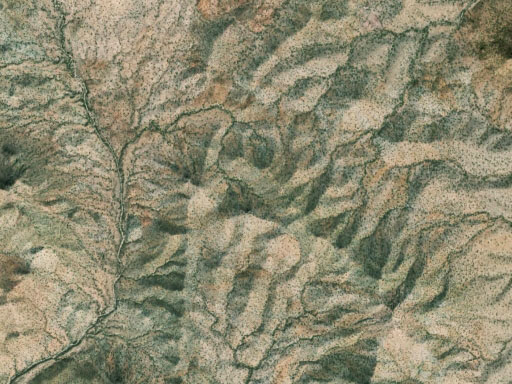}
\hskip 10pt
\includegraphics[width=6.5cm]{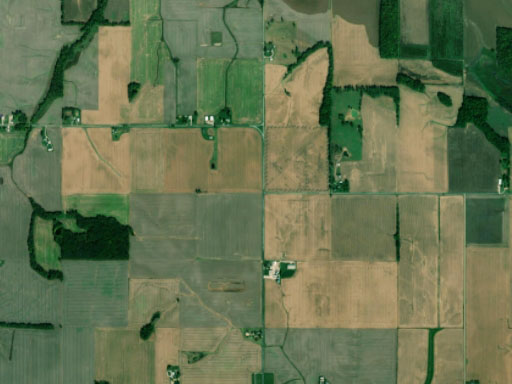}
\caption{Typical ``desert'' (\texttt{DS}) and ``farm'' (\texttt{FM}) terrain views}
\label{fig:Sim_Scenario_Terrain_desert}
\end{figure}

\item The \say{forest} (\hypertt{FR}) zone is located in the deciduous forestlands of Vermont and New Hampshire (\hypertt{USA}). The terrain is made up of forests and woodland, with some clearcuts, small towns, and roads. There are virtually no flat areas, as the land is made up by hills and small to medium size mountains that are never very steep. The valleys range from \nm{100} to \nm{300 \, m} above \hypertt{MSL}, while the tops of the mountains reach \nm{500} to \nm{900 \, m}. Features are plentiful in the woodlands.

\item The \say{mix} (\hypertt{MX}) zone is located in northern Mississippi and extreme southwestern Tennessee (\hypertt{USA}). Approximately half of the land consists of woodland in the hills, and the other half is made up by farmland in the valleys, with a few small towns and roads. Altitude changes are always presents and the terrain is never flat, but they are smaller than in the \hypertt{DS} and \hypertt{FR} zones, with the altitude oscillating between \nm{100} and \nm{200 \, m} above \hypertt{MSL}. 
\begin{figure}[h]
\centering
\includegraphics[width=6.5cm]{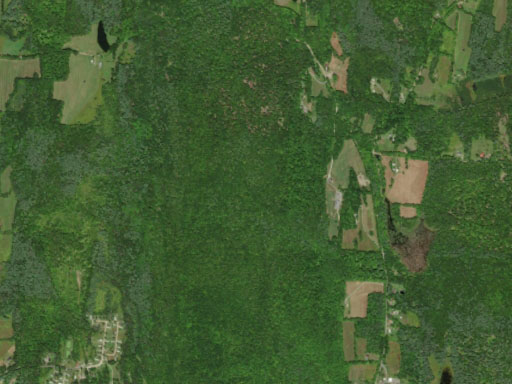}
\hskip 10pt
\includegraphics[width=6.5cm]{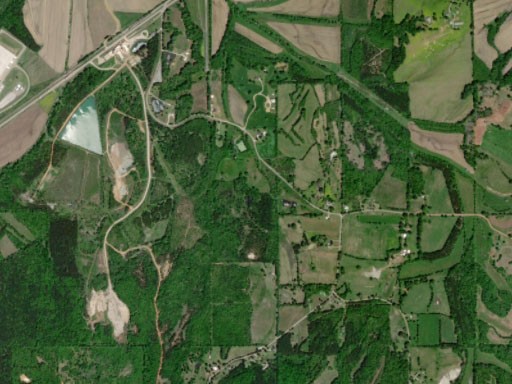}
\caption{Typical ``forest'' (\texttt{FR}) and ``mix'' (\texttt{MX}) terrain views}
\label{fig:Sim_Scenario_Terrain_forest}
\end{figure}
\end{itemize}

The short duration and continuous maneuvering of scenario \#2 enables the use of two additional terrain types. These two zones are not employed in scenario \#1 because the authors could not locate wide enough areas with a prevalence of this type of terrain (note that scenario \#1 trajectories can conclude up to \nm{125 \, km} in any direction from its initial coordinates, but only \nm{12 \, km} for scenario \#2).
\begin{itemize}
\item The \say{prairie} (\hypertt{PR}) zone is located in the Everglades floodlands of southern Florida (\hypertt{USA}). It consists of flat grasslands, swamps, and tree islands located a few meters above \hypertt{MSL}, with the only human infrastructure being a few dirt roads and landing strips, but no settlements. Features may be difficult to obtain in some areas due to the lack of texture.

\item The \say{urban} (\hypertt{UR}) zone is located in the Los Angeles metropolitan area (California, \hypertt{USA}). It is composed by a combination of single family houses and commercial buildings separated by freeways and streets. There is some vegetation but no natural landscapes, and the terrain is flat and close to \hypertt{MSL}.
\begin{figure}[h]
\centering
\includegraphics[width=6.5cm]{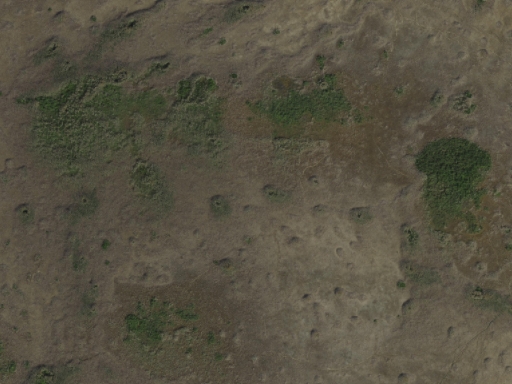}
\hskip 10pt
\includegraphics[width=6.5cm]{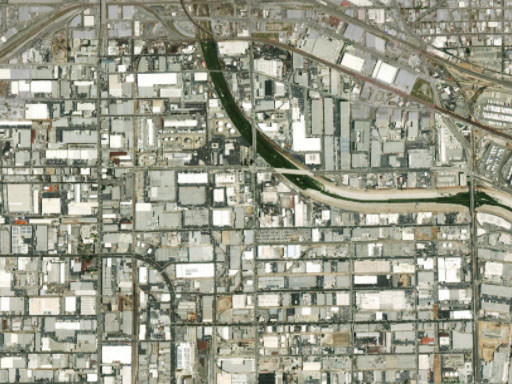}
\caption{Typical ``prairie'' (\texttt{PR}) and ``urban'' (\texttt{UR}) terrain views}
\label{fig:Sim_Scenario_Terrain_prairie}
\end{figure}
\end{itemize}

The \hypertt{MX} terrain zone is considered the most generic and hence employed to evaluate the visual algorithms in section \ref{sec:Results}. Although scenario \#2 also makes use of the four terrain types listed for scenario \#1 (\hypertt{DS}, \hypertt{FM}, \hypertt{FR}, and \hypertt{MX}), it is worth noting that the variability of the terrain is significantly higher for scenario \#1 because of the bigger land extension covered. The altitude relief, abundance or scarcity of features, land use diversity, and presence of rivers and mountains is hence more varied when executing a given run of scenario \#1 over a certain type of terrain, than when executing the same run for scenario \#2. From the point of view of the influence of the terrain on the visual navigation algorithms, scenario \#1 should theoretically be more challenging than \#2.
\begin{center}
\begin{tabular}{llrrrrrrrr}
\hline
\multicolumn{2}{l}{Scenario \nm{\#1} Zone} & \multicolumn{2}{c}{\hypertt{MX}} & \multicolumn{2}{c}{\hypertt{FR}} & \multicolumn{2}{c}{\hypertt{FM}} & \multicolumn{2}{c}{\hypertt{DS}} \\
\multicolumn{2}{l}{\nm{\Deltaxhorvis\lrp{\tEND}}} & \multicolumn{1}{c}{\nm{\lrsb{m}}} & \multicolumn{1}{c}{\nm{\lrsb{\%}}} & \multicolumn{1}{c}{\nm{\lrsb{m}}} & \multicolumn{1}{c}{\nm{\lrsb{\%}}} & \multicolumn{1}{c}{\nm{\lrsb{m}}} & \multicolumn{1}{c}{\nm{\lrsb{\%}}} & \multicolumn{1}{c}{\nm{\lrsb{m}}} & \multicolumn{1}{c}{\nm{\lrsb{\%}}} \\
\hline
\multirow{3}{*}{\hypertt{IA-VNSE}} & mean &  488 & \textbf{0.46} &  566 & \textbf{0.53} &  489 & \textbf{0.45} &  514 & \textbf{0.48} \\
                                   & std  &  350 &         0.31  &  406 &         0.38  &  322 &         0.28  &  352 &         0.31 \\
								   & max  & 1957 &         1.48  & 2058 &         1.71  & 1783 &         1.34  & 1667 &         1.37 \\
\hline
\end{tabular}
\end{center}

\captionof{table}{Influence of terrain type on final horizontal position \texttt{IA-VNSE} for scenario \nm{\#1} (100 runs)} \label{tab:Ana_vis_assist_zone_results_hor_base}

Table \ref{tab:Ana_vis_assist_zone_results_hor_base} and figure \ref{fig:Ana_vis_assist_zone_results_hor_base} show the horizontal position \hypertt{IA-NVSE} for scenario \nm{\#1} and all terrain types. Table \ref{tab:Ana_vis_assist_zone_results_hor_alter} and figure \ref{fig:Ana_vis_assist_zone_results_hor_alter} do the same for scenario \nm{\#2}. 
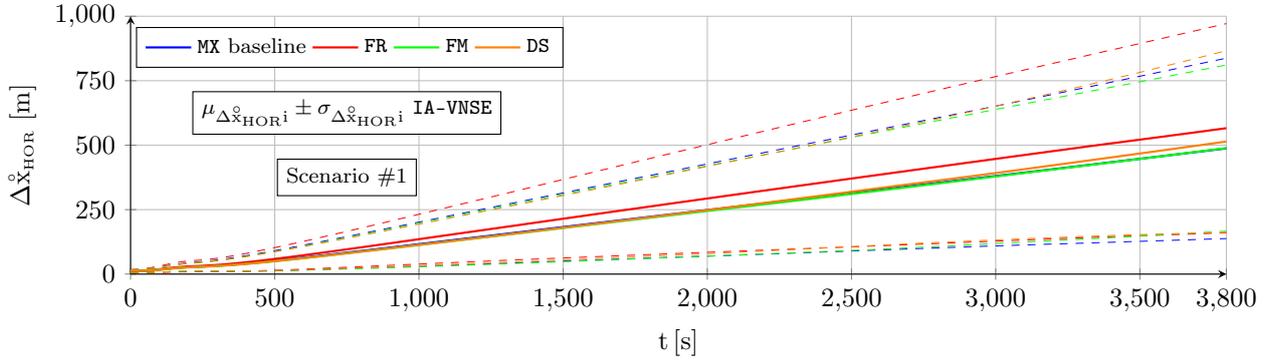
\begin{figure}[h]
\centering
\pgfplotsset{
	every axis legend/.append style={
		at={(0.20,0.80)},
		anchor=south,
	},
}
\begin{tikzpicture}
\begin{axis}[
cycle list={{blue,no markers,thick},
            {red,no markers,thick},
            {green,no markers,thick},
			{orange,no markers,thick},
            {blue,dashed,no markers,ultra thin},{blue,dashed,no markers,ultra thin},
            {red,dashed,no markers,ultra thin},{red,dashed,no markers,ultra thin},
            {green,dashed,no markers,ultra thin},{green,dashed,no markers,ultra thin},
            {orange,dashed,no markers,ultra thin},{orange,dashed,no markers,ultra thin}},
width=16.0cm,
height=5.0cm,
xmin=0, xmax=3800, xtick={0,500,...,3500,3800},
xlabel={\nm{t \lrsb{s}}},
xmajorgrids,
ymin=0, ymax=1000, ytick={0,250,...,1000},
ylabel={\nm{\Deltaxhorvis \, \lrsb{m}}},
ymajorgrids,
axis lines=left,
axis line style={-stealth},
legend entries={\hypertt{MX} baseline,
				\hypertt{FR},
				\hypertt{FM},
				\hypertt{DS}},
legend columns=4,
legend style={font=\footnotesize},
legend cell align=left,
]
\pgfplotstableread{figs/ch19_ana_vis_assist/zone_xhor/error_base_pos_hor_m_pc_vis.txt}\mytablevis
\addplot table [header=false, x index=0,y index=1] {\mytablevis};
\addplot table [header=false, x index=0,y index=4] {\mytablevis};
\addplot table [header=false, x index=0,y index=7] {\mytablevis};
\addplot table [header=false, x index=0,y index=10] {\mytablevis};
\addplot table [header=false, x index=0,y index=2] {\mytablevis};
\addplot table [header=false, x index=0,y index=3] {\mytablevis};
\addplot table [header=false, x index=0,y index=5] {\mytablevis};
\addplot table [header=false, x index=0,y index=6] {\mytablevis};
\addplot table [header=false, x index=0,y index=8] {\mytablevis};
\addplot table [header=false, x index=0,y index=9] {\mytablevis};
\addplot table [header=false, x index=0,y index=11] {\mytablevis};
\addplot table [header=false, x index=0,y index=12] {\mytablevis};
\path node [draw, shape=rectangle, fill=white] at (750,375) {\footnotesize Scenario \nm{\#1}};
\path node [draw, shape=rectangle, fill=white] at (750,625) {\footnotesize \nm{\mui{\Deltaxhorvis} \pm \sigmai{\Deltaxhorvis}} \hypertt{IA-VNSE}};
\end{axis}   
\end{tikzpicture}
\caption{Influence of terrain type on horizontal position \texttt{IA-VNSE} for scenario \nm{\#1} (100 runs)} 
\label{fig:Ana_vis_assist_zone_results_hor_base}
\end{figure}

The influence of the terrain type on the horizontal position \hypertt{IA-VNSE} is very small, with slim differences among the various evaluated terrains. The only terrain type that clearly deviates from the others is \hypertt{FR}, with slight but consistently worse horizontal position estimations for both scenarios. This behavior stands out as the abundant texture and continuous smooth vertical relief of the \hypertt{FR} terrain is a priori beneficial for the visual estimations.
\begin{center}
\begin{tabular}{llrrrrrrrrrrrr}
\hline
\multicolumn{2}{l}{Scenario \nm{\#2} Zone} & \multicolumn{2}{c}{\hypertt{MX}} & \multicolumn{2}{c}{\hypertt{FR}} & \multicolumn{2}{c}{\hypertt{FM}} & \multicolumn{2}{c}{\hypertt{DS}} & \multicolumn{2}{c}{\hypertt{UR}} & \multicolumn{2}{c}{\hypertt{PR}} \\
\multicolumn{2}{l}{\nm{\Deltaxhorvis\lrp{\tEND}}} & \multicolumn{1}{c}{\nm{\lrsb{m}}} & \multicolumn{1}{c}{\nm{\lrsb{\%}}} & \multicolumn{1}{c}{\nm{\lrsb{m}}} & \multicolumn{1}{c}{\nm{\lrsb{\%}}} & \multicolumn{1}{c}{\nm{\lrsb{m}}} & \multicolumn{1}{c}{\nm{\lrsb{\%}}} & \multicolumn{1}{c}{\nm{\lrsb{m}}} & \multicolumn{1}{c}{\nm{\lrsb{\%}}} & \multicolumn{1}{c}{\nm{\lrsb{m}}} & \multicolumn{1}{c}{\nm{\lrsb{\%}}} & \multicolumn{1}{c}{\nm{\lrsb{m}}} & \multicolumn{1}{c}{\nm{\lrsb{\%}}} \\ 
\hline
\multirow{3}{*}{\hypertt{IA-VNSE}} & mean &  33 & \textbf{0.23} &  40 & \textbf{0.28} &  33 & \textbf{0.23} &  31 & \textbf{0.22} &  32 & \textbf{0.23} &  31 & \textbf{0.22} \\
                                   & std  &  26 &         0.18  &  35 &         0.24  &  24 &         0.17  &  24 &         0.17  &  25 &         0.18  &  25 &         0.17 \\
								   & max  & 130 &         0.98  & 188 &         1.29  & 117 &         0.85  & 114 &         0.86  & 128 &         0.96  & 119 &         0.90 \\
\hline
\end{tabular}
\end{center}

\captionof{table}{Influence of terrain type on final horizontal position \texttt{IA-VNSE} for scenario \nm{\#2} (100 runs)} \label{tab:Ana_vis_assist_zone_results_hor_alter}

Although beneficial for the \hypertt{SVO} pipeline, the more pronounced vertical relief of the \hypertt{FR} terrain type breaches the flat terrain assumption of the initial homography, hampering its accuracy, and hence results in less precise initial estimations, including that of the scale. The \hypertt{IA-VNS} has no means to compensate the initial scale errors, which remain approximately equal (percentage wise) for the full duration of both scenarios.
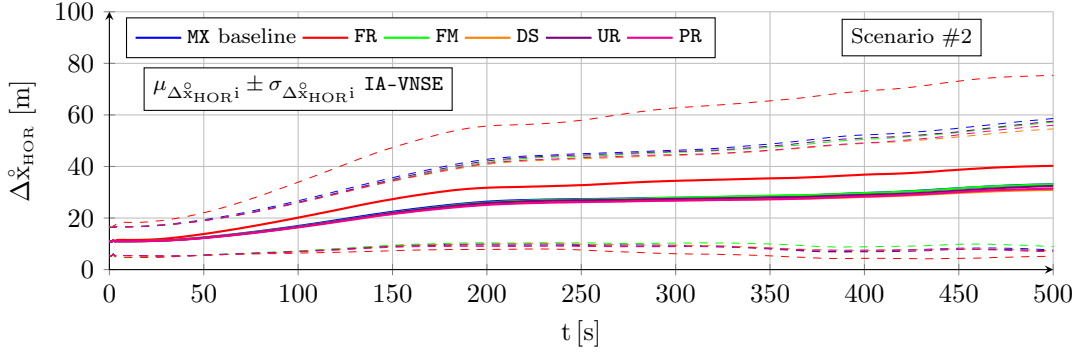
\begin{figure}[h]
\centering
\pgfplotsset{
	every axis legend/.append style={
		at={(0.33,0.82)},
		anchor=south,
	},
}
\begin{tikzpicture}
\begin{axis}[
cycle list={{blue,no markers,thick},
            {red,no markers,thick},
            {green,no markers,thick},
			{orange,no markers,thick},
            {violet,no markers,thick},
            {magenta,no markers,thick},
            {blue,dashed,no markers,ultra thin},{blue,dashed,no markers,ultra thin},
            {red,dashed,no markers,ultra thin},{red,dashed,no markers,ultra thin},
            {green,dashed,no markers,ultra thin},{green,dashed,no markers,ultra thin},
            {orange,dashed,no markers,ultra thin},{orange,dashed,no markers,ultra thin},
            {violet,dashed,no markers,ultra thin},{violet,dashed,no markers,ultra thin},
            {magenta,dashed,no markers,ultra thin},{magenta,dashed,no markers,ultra thin}},
width=14.0cm,
height=5.0cm,
xmin=0, xmax=500, xtick={0,50,...,500},
xlabel={\nm{t \lrsb{s}}},
xmajorgrids,
ymin=0, ymax=100, ytick={0,20,...,100},
ylabel={\nm{\Deltaxhorvis \, \lrsb{m}}},
ymajorgrids,
axis lines=left,
axis line style={-stealth},
legend entries={\hypertt{MX} baseline,
				\hypertt{FR},
				\hypertt{FM},
				\hypertt{DS},
				\hypertt{UR},
				\hypertt{PR}},
legend columns=6,
legend style={font=\footnotesize},
legend cell align=left,
]
\pgfplotstableread{figs/ch19_ana_vis_assist/zone_xhor/error_alter_pos_hor_m_pc_vis.txt}\mytablevis
\addplot table [header=false, x index=0,y index=1] {\mytablevis};
\addplot table [header=false, x index=0,y index=4] {\mytablevis};
\addplot table [header=false, x index=0,y index=7] {\mytablevis};
\addplot table [header=false, x index=0,y index=10] {\mytablevis};
\addplot table [header=false, x index=0,y index=13] {\mytablevis};
\addplot table [header=false, x index=0,y index=16] {\mytablevis};
\addplot table [header=false, x index=0,y index=2] {\mytablevis};
\addplot table [header=false, x index=0,y index=3] {\mytablevis};
\addplot table [header=false, x index=0,y index=5] {\mytablevis};
\addplot table [header=false, x index=0,y index=6] {\mytablevis};
\addplot table [header=false, x index=0,y index=8] {\mytablevis};
\addplot table [header=false, x index=0,y index=9] {\mytablevis};
\addplot table [header=false, x index=0,y index=11] {\mytablevis};
\addplot table [header=false, x index=0,y index=12] {\mytablevis};
\addplot table [header=false, x index=0,y index=14] {\mytablevis};
\addplot table [header=false, x index=0,y index=15] {\mytablevis};
\addplot table [header=false, x index=0,y index=17] {\mytablevis};
\addplot table [header=false, x index=0,y index=18] {\mytablevis};
\path node [draw, shape=rectangle, fill=white] at (425,90) {\footnotesize Scenario \nm{\#2}};
\path node [draw, shape=rectangle, fill=white] at (100,70) {\footnotesize \nm{\mui{\Deltaxhorvis} \pm \sigmai{\Deltaxhorvis}} \hypertt{IA-VNSE}};
\end{axis}   
\end{tikzpicture}
\caption{Influence of terrain type on horizontal position \texttt{IA-VNSE} for scenario \nm{\#2} (100 runs)} 
\label{fig:Ana_vis_assist_zone_results_hor_alter}
\end{figure}

A similar but opposite reasoning is applicable to the \hypertt{FM} type and in a lesser degree to the \hypertt{UR} and \hypertt{PR} types. Although a flat terrain in which all terrain features are located at a similar altitude is detrimental to the overall accuracy of \hypertt{SVO}, and results in slightly worse body attitude and vertical position estimations, it is beneficial for the homography initialization and the scale determination, resulting in consistently more accurate horizontal position estimations. 


\section{Summary of Results} \label{sec:Summary}

This article proposes an \hypertt{SVO} based \hypertt{IA-VNS} installed onboard a fixed wing autonomous \hypertt{UAV} that takes advantage of the \hypertt{GNSS}-Denied estimations provided by an \hypertt{INS} to assist the visual pose optimization algorithms. The method is inspired in a \hypertt{PI} control loop, in which the inertial attitude and altitude outputs act as targets to ensure that the visual estimations do not deviate in excess from their inertial counterparts, resulting in major improvements when estimating the aircraft horizontal position without the use of \hypertt{GNSS} signals. The results obtained when applying the proposed algorithms to high fidelity Monte Carlo simulations of two scenarios representative of the challenges of \hypertt{GNSS}-Denied navigation indicate the following:
\begin{itemize}
\item The \textbf{body attitude} estimation shows significant quantitative improvements over a stand-alone \hypertt{VNS} in both pitch and bank angle estimations, with no negative influence on the yaw angle estimations. A small amount of drift with time is present, and can not be fully eliminated. Body pitch and bank angle estimations do not deviate in excess from their \hypertt{INS} counterparts, while the body yaw angle visual estimation is significantly more accurate than that obtained by the \hypertt{INS}. 

\item The \textbf{vertical position} estimation shows major improvements over that of a stand-alone \hypertt{VNS}, not only quantitatively but also qualitatively, as drift is fully eliminated. The visual estimation does not deviate in excess from the inertial one, which is bounded by atmospheric physics.

\item The \textbf{horizontal position} estimation, whose improvement is the main objective of the proposed algorithm, shows major gains when compared to either the stand-alone \hypertt{VNS} or the \hypertt{INS}, although drift is still present. 
\end{itemize}

In addition, although the \textbf{terrain} texture (or lack of) and its elevation relief are key factors for the visual odometry algorithms, their influence on the aircraft pose estimation results are slim, and the accuracy of the \hypertt{IA-VNS} does not vary significantly among the various evaluated terrain types.


\section{Conclusions} \label{sec:Conclusion}

The proposed inertially assisted \hypertt{VNS} (\hypertt{IA-VNS}), which in addition to the images taken by an onboard camera also relies on the outputs of an \hypertt{INS} specifically designed for the challenges faced by autonomous fixed wing aircraft that encounter \hypertt{GNSS}-Denied conditions, possesses significant advantages in both accuracy and resilience when compared with a stand-alone \hypertt{VNS}, the most important of which is a major reduction in its horizontal position drift independently of the terrain type overflown by the aircraft. The proposed \hypertt{IA-VNS} can significantly increase the possibilities of the aircraft safely reaching the vicinity of the intended recovery location upon the loss of \hypertt{GNSS} signals, from where it can be landed by remote control.

\appendix 

\section{Optical Flow} \label{sec:OpticalFlow}

Consider a pinhole camera \cite{Hartley2003} (one that adopts an ideal perspective projection) such as that depicted in figure \ref{fig:Vis_camera_pinhole}. The \emph{image frame} \nm{\FIMG} is a two-dimensional Cartesian reference frame \nm{\FIMG = \{\OIMG ,\, \iIMGi, \, \iIMGii\}} whose axes are parallel to those of the \nm{\FC} camera frame (\nm{\iIMGi \parallel \iCi, \, \iIMGii \parallel \iCii}), and whose origin \nm{\OIMG} is located on the focal plane displaced a distance \nm{\cIMG} from the principal point so the \nm{\FIMG} coordinates \nm{\pIMGi} and \nm{\pIMGii} of any point in the image domain \nm{\Omega} are always positive. The perspective projection map \nm{\pIMG = \vec{\Pi}\lrp{\pC}} that converts points viewed in \nm{\FC} into \nm{\FIMG} is hence the following:
\begin{eqnarray}
\nm{\pIMGi}  & = & \nm{\dfrac{f}{\sPX} \ \dfrac{\pCi}{\pCiii} + \cIMGi} \label{eq:Vis_camera_pIMGi} \\
\nm{\pIMGii} & = & \nm{\dfrac{f}{\sPX} \ \dfrac{\pCii}{\pCiii} + \cIMGii} \label{eq:Vis_camera_pIMGii} 
\end{eqnarray}
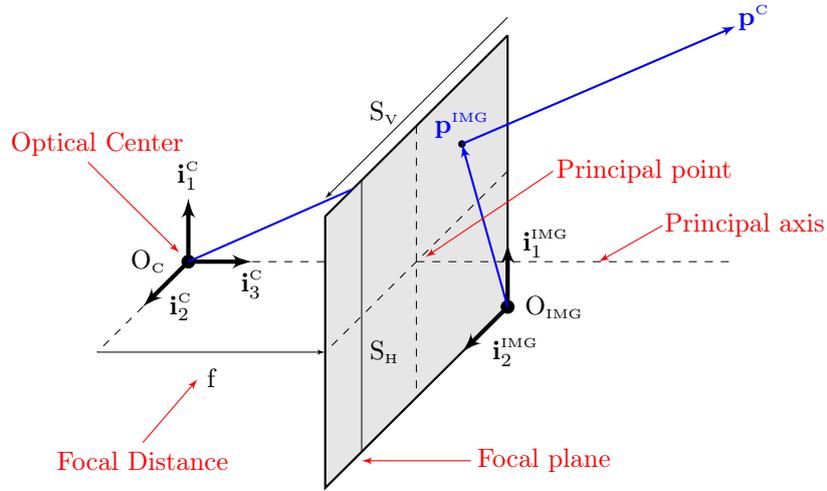
\begin{figure}[h]
\centering
\begin{tikzpicture}[auto,>=latex',scale=1.2]
	\coordinate (OC)    at (+0.0,+0.0);
	\coordinate (iCi)   at  ($(OC) + (+0.0,+0.7)$);
	\coordinate (iCii)  at  ($(OC) + (-0.5,-0.5)$);
	\coordinate (iCiii) at  ($(OC) + (+0.7,+0.0)$);
	\node [above of=iCi,   node distance=0.3cm] () {\nm{\iCi}};
	\node [right of=iCii,   node distance=0.5cm] () {\nm{\iCii}};
	\node [below of=iCiii, node distance=0.3cm] () {\nm{\iCiii}};
	\draw [ultra thick] [->] (OC) -- (iCi) {};
	\draw [ultra thick] [->] (OC) -- (iCii) {};
	\draw [ultra thick] [->] (OC) -- (iCiii) {};		
	\filldraw [black] (OC) circle [radius=2pt] node [left=5pt] {\nm{\OC}};	
	
	\coordinate (PP)    at ($(OC) + (+2.5,+0.0)$);	
	\coordinate (ul)    at ($(PP) + (+1.0,+2.5)$);
	\coordinate (ur)    at ($(PP) + (-1.0,+0.5)$);
	\coordinate (ll)    at ($(PP) + (+1.0,-0.5)$);
	\coordinate (lr)    at ($(PP) + (-1.0,-2.5)$);
	\fill [color=gray!20] (ul) -- (ur) -- (lr) -- (ll) -- cycle;
		
	\coordinate (OIMG)  at ($(PP) + (+1.0,-0.5)$);	
	\coordinate (iIMGi)   at ($(OIMG) + (+0.0,+0.7)$);
	\coordinate (iIMGii)  at ($(OIMG) + (-0.5,-0.5)$);
	\node [right of=iIMGi,   node distance=0.5cm] () {\nm{\iIMGi}};
	\node [right of=iIMGii,   node distance=0.7cm] () {\nm{\iIMGii}};
	\draw [ultra thick] [->] (OIMG) -- (iIMGi) {};
	\draw [ultra thick] [->] (OIMG) -- (iIMGii) {};
	\filldraw [black] (OIMG) circle [radius=2pt] node [right=3pt] {\nm{\OIMG}};

	\draw [thick] [-] (ul) -- (ur) -- (lr) -- (ll) -- (ul);
	\draw [ultra thin, dashed] [-] (PP) -- ($(PP) + (+3.5,+0.0)$);
	\draw [ultra thin, dashed] [-] (OC) -- ($(OC) + (+1.5,+0.0)$);
	\draw [ultra thin, dashed] [-] (OC) -- ($(OC) + (-1.0,-1.0)$);
	\draw [ultra thin, dashed] [-] ($(PP) + (+1.0,+1.0)$) -- ($(PP) + (-1.0,-1.0)$);
	\draw [ultra thin, dashed] [-] ($(PP) + (+0.0,+1.5)$) -- ($(PP) + (-0.0,-1.5)$);
	\draw [ultra thin] [->] ($(OC) + (-1.0,-1.0)$) -- node[below=1mm] {f} ($(PP) + (-1.0,-1.0)$);
		
	\draw [ultra thin] ($(lr) + (+0.4,+0.4)$) -- ($(ur) + (+0.4,+0.4)$);		
	\node at ($(PP) + (-0.35,-1.0)$) {\nm{\Sh}};		
	
	\draw [ultra thin] [->] ($(ul) + (+0.0,+0.2)$) -- ($(ur) + (+0.0,+0.2)$);		
	\node at ($(PP) + (-0.35,+1.65)$) {\nm{\Sv}};		
	
	\coordinate (PIMG)    at ($(OC) + (+3.0,+1.3)$);
	\filldraw [black] (PIMG) circle [radius=1pt] node [left=5pt] {};	
	\draw [thick, blue] [->] (PIMG) -- ($(PIMG) + (+3.0,+1.3)$);
	\draw [thick, blue] [-] (OC) -- ($(OC) + (+1.8,+0.8)$);
	\draw [thick, blue] [->](OIMG) -- (PIMG);

	\node [blue] at ($(PIMG) + (+90:0.2cm)$) {\nm{\pIMG}};		
	\node [blue] at ($(PIMG) + (+3.2,+1.4)$) {\nm{\pC}};		
		
	\draw [red] [->] ($(OC) + (+3.1,-2.2)$) -- ($(OC) + (+1.9,-2.2)$);		
	\node [red] at ($(OC) + (+3.9,-2.2)$) {Focal plane};		

	\draw [red] [->] ($(PP) + (+1.5,+1.0)$) -- ($(PP) + (+0.1,+0.05)$);		
	\node [red] at ($(PP) + (+2.5,1.0)$) {Principal point};		
		
	\draw [red] [->] ($(PP) + (+2.7,+0.4)$) -- ($(PP) + (+2.0,+0.0)$);		
	\node [red] at ($(PP) + (+3.6,0.4)$) {Principal axis};				
		
	\draw [red] [->] ($(OC) + (-1.1,+1.1)$) -- ($(OC) + (-0.1,+0.1)$);		
	\node [red] at ($(OC) + (-1.0,1.3)$) {Optical Center};			
				
	\draw [red] [->] ($(OC) + (-0.5,-2.0)$) -- ($(OC) + (+0.1,-1.3)$);		
	\node [red] at ($(OC) + (-0.5,-2.2)$) {Focal Distance};		
	
\end{tikzpicture}
\caption{Frontal pinhole camera model}
\label{fig:Vis_camera_pinhole}
\end{figure}

Consider also that the camera is moving with respect to the Earth while maintaining within its field of view a given point \nm{\vec p} fixed to the Earth surface. The composition of positions and its time derivation, considering \hypertt{ECEF} as \nm{\FE}, the camera frame as \nm{\FC}, and a frame \nm{\FP} with its origin in the terrain point \nm{\vec p} that does not move with respect to \nm{\FE}, results in the following expression when viewed in \nm{\FC}:
\begin{eqnarray}
\nm{\TEPE} & = & \nm{\TCPE + \TECE = \REC \; \TCPC + \TECE} \label{eq:MOT_Comp_Pos1} \\
\nm{\TEPEdot} & = & \nm{\RECdot \; \TCPC +\REC \; \TCPCdot + \TECEdot = \REC \; \wECCskew \; \TCPC +\REC \; \TCPCdot + \TECEdot} \label{eq:MOT_Comp_LinearVel2} \\
\nm{\vEPE} & = & \nm{\REC \; \vCPC + \vECE +\REC \; \wECCskew \; \TCPC = \vCPE + \vECE + \wECEskew \; \TCPE} \label{eq:MOT_Comp_LinearVel4} \\
\nm{\vEPC} & = & \nm{\RCE \; \vEPE = \vECC + \vCPC + \wECCskew \, \TCPC = \vec 0} \label {eq:Vis_Optical_Flow_1}
\end{eqnarray}

Note that (\ref{eq:Vis_Optical_Flow_1}) connects the point coordinates as viewed from the camera \nm{\TCPC = \pC} and their time derivative \nm{\vCPC = \pCdot} with the twist \nm{\xiECC} of the motion of the camera with respect to the Earth viewed in the \nm{\FC} or local frame, which is composed by its linear and angular velocities \nm{\vECC} and \nm{\wECC} \cite{Sola2018}.
\neweq{\vCPC = \pCdot = - \vECC - \wECCskew \, \TCPC = - \vECC - \wECCskew \, \pC}{eq:Vis_Optical_Flow_2}

The homogeneous camera coordinates \nm{\pCbar} are defined as the ratio between the camera coordinates \nm{\pC} and its third coordinate or depth \nm{\pCiii}, and represent an alternative view to \nm{\pIMG = \vec{\Pi}\lrp{\pC}} of how the point is projected in the image. Its time derivative is hence:
\neweq{\pCbar = \dfrac{\pC}{\pCiii} \ \longrightarrow \ \pCbardot = \lrsb{\pCbari, \, \pCbarii, \, 1}^T = \dfrac{\pCiii \ \pCdot - \pCiiidot \ \pC}{{p_{\sss3}^{{\sss C} \ 2}}}} {eq:Vis_Optical_Flow_3} 

Substituting both \nm{\pC} and \nm{\pCiii} within (\ref{eq:Vis_Optical_Flow_2}) into (\ref{eq:Vis_Optical_Flow_3}), rearranging terms, and considering the (\ref{eq:Vis_camera_pIMGi}, \ref{eq:Vis_camera_pIMGii}) relationship between the image and the homogeneous camera coordinates, leads to the following expression for the \emph{optical flow} \cite{Heeger1998} or variation of the point image coordinates:
\neweq{\pIMGdot = \vec J_{OF}\big(\vec \Pi\lrp{\pC}\big) \ \xiECC = f \ \begin{bmatrix} \nm{- \dfrac{1}{\pCiii}} & 0 & \nm{\dfrac{\pCbari}{\pCiii}} & \nm{\pCbari \, \pCbarii} & \nm{- 1 - {\bar p}_{\sss1}^{{\sss C} \ 2}} & \nm{\pCbarii} \\ 0 & \nm{- \dfrac{1}{\pCiii}} & \nm{\dfrac{\pCbarii}{\pCiii}} & \nm{1 + {\bar p}_{\sss2}^{{\sss C} \ 2}} & \nm{- \pCbari \, \pCbarii} & \nm{- \pCbari} \end{bmatrix} \ \begin{bmatrix} \nm{\vECC} \\ \nm{\wECC} \end{bmatrix}}{eq:Vis_Optical_FlowR}

Considering that the twist \nm{\vec \xi} is the time derivative of the transform vector \nm{\vec \tau} \cite{Sola2018}, the \emph{optical flow} \nm{\vec J_{OF}} is defined as the derivative of the local frame ideal perspective projection of a point fixed to the spatial frame with respect to the \nm{\mathbb{SE}(3)} element \nm{\mathcal M} caused by a perturbation \nm{\Delta \vec \tau} in its local tangent space:
\begin{eqnarray}
\nm{\vec J_{OF}\big(\vec \Pi\lrp{\vec g_{\mathcal M}(\vec p)}\big)} & = & \nm{\lim_{\Delta \vec \tau\to \vec 0} \dfrac{\vec \Pi \big(\vec g_{\mathcal M \oplus \Delta \vec \tau}\lrp{\vec p}\big) - \vec \Pi \big(\vec g_{\mathcal M}\lrp{\vec p}\big)}{\Delta \vec \tau} \ \in \mathbb{R}^{2x6}} \label{eq:Vis_Optical_Flow_defR} \\
\nm{\vec \Pi\big(\vec g_{\mathcal M \oplus \Delta \vec \tau}\lrp{\vec p}\big)} & \nm{\approx} & \nm{\vec \Pi\big(\vec g_{\mathcal M}\lrp{\vec p}\big) + \lrsb{\vec J_{OF}\big(\vec \Pi\lrp{\vec g_{\mathcal M}(\vec p)}\big) \ \Delta \vec \tau} \ \in \mathbb{R}^2} \label{eq:Vis_Optical_Flow_taylorR}
\end{eqnarray}

Less formally, the optical flow Jacobian represents how the projection of a fixed point moves within the image as the camera pose varies. Note that the Jacobian only depends on the point camera (local) coordinates and the camera focal length, and that as all terms multiplying the linear twist component are divided by the image depth \nm{\pCiii}, the effect on the image of a bigger linear velocity can not be distinguished from that of a smaller depth.


\bibliographystyle{ieeetr}   
\bibliography{visual_navigation}

\end{document}